\documentclass[singlecolumn,10pt,twoside,final]{asme2ej_Draft}
\usepackage{amsfonts}
\usepackage{graphicx}
\usepackage{amsfonts}
\usepackage{graphicx}
\usepackage{amsmath, amssymb, bm}
\usepackage[fixamsmath,disallowspaces]{mathtools}
\usepackage{xcolor}
\usepackage{fixmath}

\newtheorem{remark}{Remark}

\begin{document}

\parindent0pt \parskip2pt \setcounter{topnumber}{9} %
\setcounter{bottomnumber}{9} \renewcommand{\textfraction}{0.00001}

\renewcommand {\floatpagefraction}{0.999} \renewcommand{\textfraction}{0.01} %
\renewcommand{\topfraction}{0.999} \renewcommand{\bottomfraction}{0.99} %
\renewcommand{\floatpagefraction}{0.99} \setcounter{totalnumber}{9} %
\setcounter{topnumber}{3} \setcounter{bottomnumber}{2} \renewcommand{%
\textfraction}{0.00001}

\title{A Constraint Embedding Approach for Dynamics Modeling of Parallel Kinematic Manipulators\\
with Hybrid Limbs
\vspace{-8mm}
} 
\author{
	{Andreas M\"uller}
	\affiliation{Johannes Kepler University, Linz, Austria\\
	a.mueller@jku.at}
} 

\maketitle 

\begin{abstract}
Parallel kinematic manipulators (PKM) are characterized by closed kinematic
loops, due to the parallel arrangement of limbs but also due to the
existence of kinematic loops within the limbs. Moreover, many PKM are built
with limbs constructed by serially combining kinematic loops. Such limbs are
called \emph{hybrid}, which form a particular class of complex limbs. Design
and model-based control requires accurate dynamic PKM models desirably
without model simplifications. Dynamics modeling then necessitates kinematic
relations of all members of the PKM, in contrast to the standard kinematics
modeling of PKM, where only the forward and inverse kinematics solution for
the manipulator (relating input and output motions) are computed. This
becomes more involved for PKM with hybrid limbs. In this paper a modular
modeling approach is employed, where limbs are treated separately, and the
individual dynamic equations of motions (EOM) are subsequently assembled to
the overall model. Key to the kinematic modeling is the constraint
resolution for the individual loops within the limbs. This local constraint
resolution is a special case of the general \emph{constraint embedding}
technique. The proposed method finally allows for a systematic modeling of
general PKM. The method is demonstrated for the IRSBot-2, where each limb
comprises two independent loops.

\end{abstract}

\thispagestyle{empty} \pagestyle{empty}

\setcounter{topnumber}{3} \setcounter{bottomnumber}{2} \renewcommand{%
\textfraction}{0.00001}

\section{Introduction}

Control and design of parallel kinematic manipulators (PKM) for highly
dynamic applications necessitate dynamic models with high fidelity as well
as computational efficiency. Computationally efficient modeling approaches
exist for serial manipulators \cite{Rodriguez1987,WittenburgBook,AngelesBook}
including recursive $O\left( n\right) $-formulations \cite%
{ThomasTesar1982,Stelzle1995,Featherstone2008,YamaneNakamura2009}. PKM are
multibody systems (MBS) featuring multiple kinematic loops. The dynamics
modeling of such MBS has been advanced in the last four decades, and lead to
several established modeling approaches. As far as rigid body MBS (including
discrete elastic elements), the latter include formulations in absolute
coordinates \cite{ShabanaBook}, relative coordinates \cite%
{WittenburgBook,JainBook}, and natural coordinates \cite%
{deJalon1986,deJalonBook1994}, which are applicable to general MBS. PKM, on
the other hand, possess a particular kinematic topology, which can be
exploited for deriving dedicated dynamics models. The majority of PKM are
fully parallel, i.e. the moving platform is connected to the ground (fixed
platform) by several serial limbs each comprising one actuator. The limbs
can topologically be classified as simple and complex. The prevailing class
of PKM consists of fully parallel manipulators with \emph{simple limbs},
i.e. each limb is a serial kinematic chain. For such PKM, tailored modeling
approaches were presented in a series of publications \cite%
{LeeShah1988,NakamuraGhodoussi1989,MataProvenzanoCuadradoValero2002,Abedloo2014,TRORedPKM,Dasgupta1998,WangGosselin1998,ZhaoTRO2014,EOMRedCoord1_NLD2012,EOMRedCoord2_RAS2012,KhalilIbrahim2007}%
. The key concept common to these approaches is to model each limb as a
kinematic chain with the platform attached, and to use the inverse
kinematics solution of the individual limbs to express the overall
kinematics and dynamics equations of motions (EOM) in terms of task space
coordinates.

PKM with complex limbs form another practically relevant class, for which
the Delta robot is a good example. Also gravity compensated PKM often
comprise multiple loops \cite{Gosselin2008}. Most complex limbs are built by
a serial arrangement of kinematic loops, which are referred to as \emph{%
hybrid limbs} (occasionally called \emph{serial-parallel limbs}). Moreover,
PKM with hybrid limbs constitute another prevailing class of PKM. The
systematic modeling of such PKM was reported in \cite{MMT2022} as
continuation of formulations presented in \cite%
{KhalilGuegan2004,KhalilIbrahim2007,BriotKhalilBook,AbdellatifHeimann_MMT2009}%
. The crucial step of this method is the solution of the closure constraints
for the kinematic loops within a limb, which are called \emph{intra-limb
constraints}. Incorporating this solution, the kinematics and dynamics
modeling approach for PKM with simple limbs can be adopted. The loops within
a hybrid limb are topologically independent, and the intra-limb constraints
can thus be solved independently. This method of incorporating solutions of
loop constraints is referred to as \emph{constraint embedding} \cite%
{Jain2009,JainNODY2012,Jain2012}. The embedding technique is well-known in
MBS dynamics \cite{WehageHaug1982,ShabanaBook,AmiroucheBOOK2006}, where it
usually refers to incorporating the solution of velocity and acceleration
constraints so to obtain a model in terms of independent generalized
velocities and accelerations, which goes back at least to Voronets \cite%
{Voronets1901}. A proper constraint embedding further involves the solution
of geometric constraints. Solving constraints in closed form, i.e. an
'explicit' constraint embedding, is in general not possible, however. In
this paper, a numerical constraint embedding technique for hybrid limbs is
presented, complementing the formulation in \cite{MMT2022}.

The constraint embedding does not involve simplifications, and the model
accounts for the dynamics of all bodies. Simplifications were used in
various publications, where the loops within limbs are treated as so-called
compound joints. To this end, the loop is replaced by an equivalent
kinematic transformation. For example, parallelogram loops, which are
frequently used to construct complex limbs \cite%
{Klimchik2018,Wen-TRO2020,Gauthier_JMR2008,TaghvaeipourAngelesLessard2013,Altuzarra2009}%
, are often modeled by so-called 1-DOF $\Pi $-joints \cite{AngelesBook}.
While this is kinematically equivalent (but it should be noticed that
possible internal singularities are hidden), it does not account for the
dynamics of all members of the substituted loop. The 3-DOF Delta robot is a
prominent example where each limb is frequently regarded as a hinged
parallelogram $\Pi $-joint. The dynamic effect of the parallelogram is
either disregarded, or is represented by a lumped mass, as this is formed by
slender rods. This does not allow to represent non-symmetric mass
distributions or compute joint reactions, however.

\begin{figure}[h]
\centerline{a)\includegraphics[height=5.6cm]{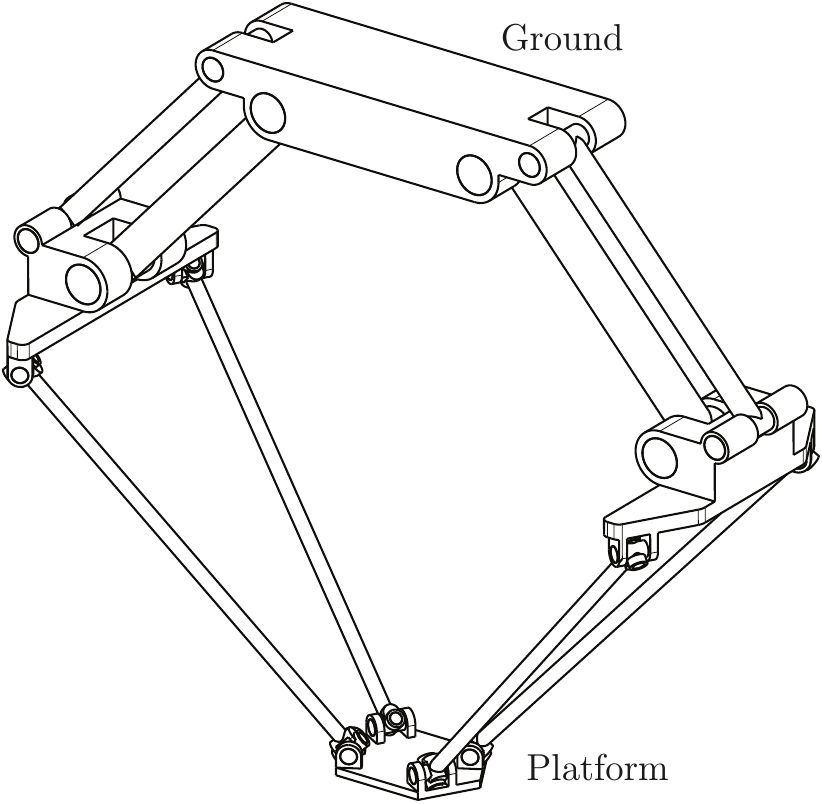}\hfil
b)\includegraphics[height=5.6cm]{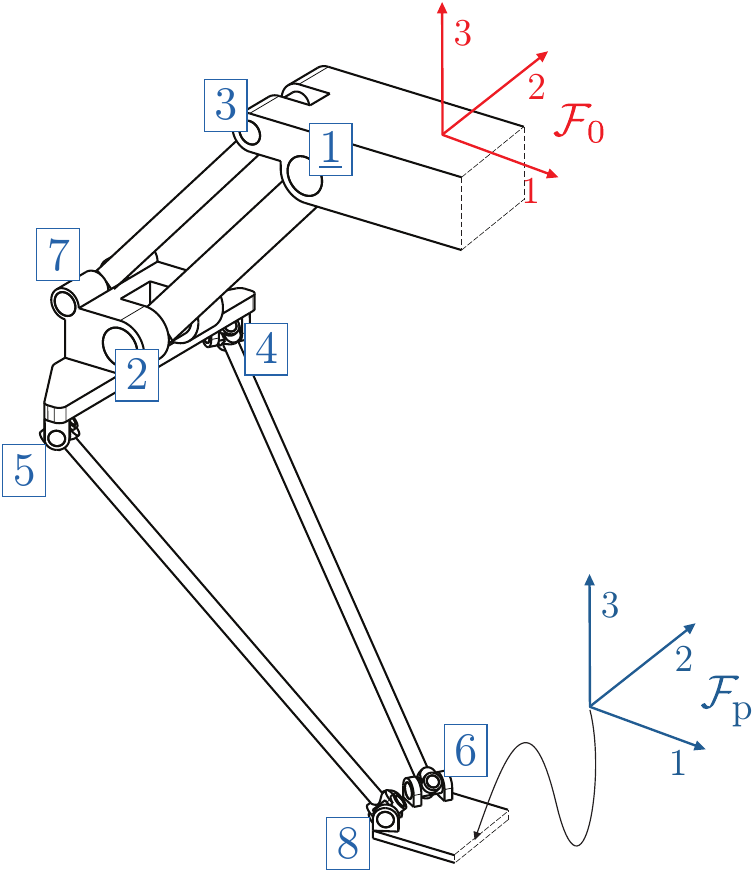}}
\caption{a) Drawing of the 2-DOF IRSBot-2, whose EE performs pure
translations in the 1-3-plane of $\mathcal{F}_{0}$. b) Separated limb
comprising two loops with platform frame $\mathcal{F}_{\mathrm{P}}$
assigned. }
\label{figIRSBot}
\end{figure}

The contribution of this paper can be summarized by means of the example in
Fig. \ref{figIRSBot}, which shows the 2-DOF IRSBot-2 with two limbs \cite%
{Germain2013}. The kinematics of either limb is to be described in terms of
the platform position. This involves two steps: 1) the local solution of the
intra-limb loop constraints, and 2) the separate solution of the inverse
kinematics problem of the individual limbs. Each limb comprises two loops,
and each of the corresponding intra-limb loop constraints can be solved
numerically in terms of independent coordinates. These solutions render the
limb a serial kinematic chain described by some independent coordinates. The
inverse kinematics of the limb, represented as serial chain in terms of the
independent coordinates, can now be solved numerically. Compared to the
standard MBS approach, this method has a reduced complexity and increased
robustness w.r.t. redundant constraints. Standard MBS formulations, on the
other hand, do not take into account the specific topology of PKM, and
operate on the overall system of constraints. As a consequence, a large
system of constraints is to be handled, and often the MBS model involves
redundant constraints (e.g. for lower mobility PKM), although the intra-limb
constraints are non-redundant, which drastically increases the computational
effort. The proposed constraint embedding method treats the intra-limb
constraints separately. It thus reduces the system size but also allows
handling redundant intra-limb constraints. The proposed local constraint
embedding approach implements the two above mentioned steps. 

The paper is organized as follows. In sec. \ref{secTopology}, the
representation of the kinematic topology of a PKM by means of a linear graph
is recalled. Sec. \ref{secLimbKin} addresses the PKM kinematics, which
involves the forward and inverse kinematics of the limbs. A model for the
limb kinematics is presented, which involves solutions of the intra-limb
loop constraints. The constraint embedding approach is then introduced in
sec. \ref{secConsEmbedding} as a numerical algorithm for evaluating the
constraint solutions. An algorithm for solving the constraints along with
the limb inverse kinematics is introduced. The dynamics EOM are presented in
sec. \ref{secEOMPKM}. The proposed task space formulation builds upon the
limb inverse kinematics. Application of the method is presented in detail in
sec. \ref{secIRSBot} for the IRSBot-2 example. The paper closes with a short
summary and conclusion in sec. \ref{secConclusion}. As the cut-joint
constraints for technical joints are crucial for PKM modeling, their
formulation is summarized in detail in appendix \ref{secCutConstr}. This
shall serve as a reference for modeling general robots with kinematic loops.
For better readability, a list of symbols is presented in appendix \ref%
{secSymbols}.

A note on the formalism used for kinematics modeling seems in order:
Throughout the paper, the kinematics is described using the Lie group /
screw theory formalism, which is a compact and 'user friendly' approach to
robot modeling \cite{ModernRobotics,Murray,SeligBook2005}. However, all
relations can be formulated by means of any other modeling approach the
reader may prefer.

\section{Graph Representation of PKM Topology%
\label{secTopology}%
}

\begin{figure}[h]
\centerline{
a)\includegraphics[height=6.3cm]{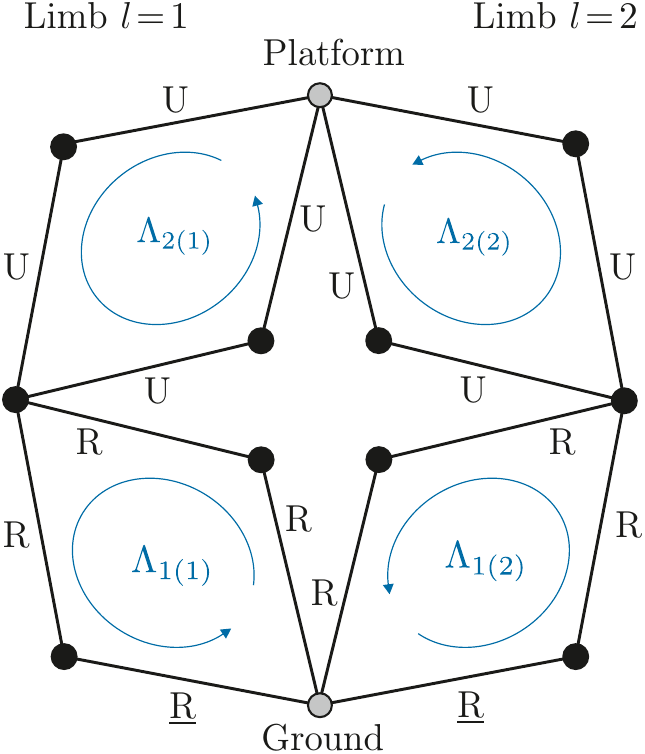}\hfil
b)\includegraphics[height=5.6cm]{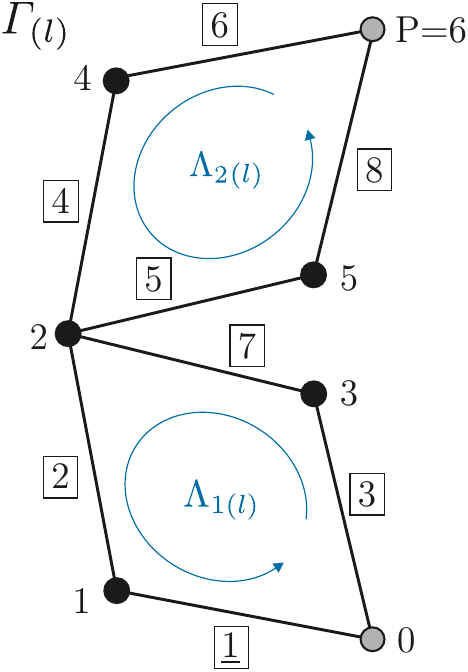}
\hfil
c)\includegraphics[height=5.6cm]{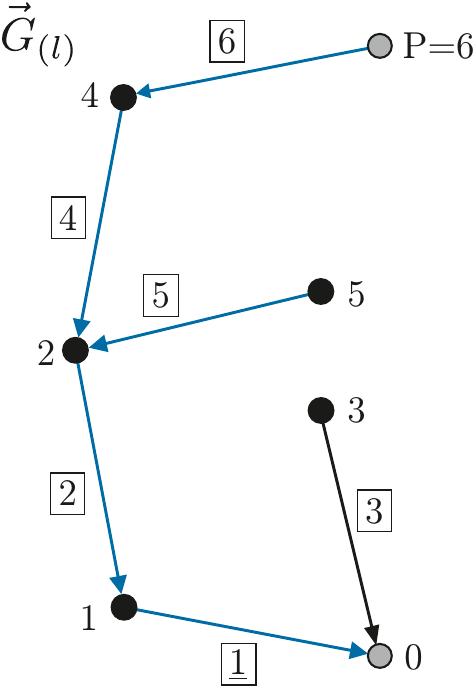}
}
\caption{a) Topological graph $\Gamma $ of the IRSBot-2. Indicated are the
FCs $\Lambda _{1\left( l\right) }$ and $\Lambda _{2\left( l\right) }$ of the
two limbs $l=1,2$. b) Subgraph $\Gamma _{\left( l\right) }$ of $\Gamma $
representing limb $l=1,2$ of the IRSBot-2. c) A ground-directed spanning
tree $\vec{G}_{\left( l\right) }$ on $\Gamma _{\left( l\right) }$.}
\label{figIRSBotGraph}
\end{figure}

The kinematic topology of a PKM is represented by the \emph{topological graph%
}, a labeled graph denoted $\Gamma $ \cite%
{TopologyRobotica,JainMUBO2011_1,WittenburgBook}. Bodies are represented by
the vertices, and joints by the edges of $\Gamma $. The topologically
independent loops of $\Gamma $ are called fundamental cycles (FC). The
number of FC is denoted with $\gamma $. Fig. \ref{figIRSBotGraph}a) shows
the topological graph of a the IRSBot-2 in fig. \ref{figIRSBot}.

Each limb of the PKM connects the base (ground, fixed platform) with the
moving platform. Otherwise the limbs are mutually disconnected, and each
limb is topologically represented by a subgraph. The subgraph associated to
limb $l$ is denoted with $\Gamma _{\left( l\right) }$. Vertices (bodies) of $%
\Gamma _{\left( l\right) }$ are numbered with $i=0,1,2,\ldots \mathfrak{n}%
_{l}$, where the ground is indexed with 0. It will be helpful to label the
platform with P. Edges (joints) are indexed with $i=1,\ldots ,\mathfrak{N}%
_{l}$. Fig. \ref{figIRSBotGraph}b) shows the subgraph for a limb of the
IRSBot-2. Clearly, all limbs are structurally identical, and thus all $%
\Gamma _{\left( l\right) }$ are congruent.

The peculiarity of complex limbs is that the subgraphs $\Gamma _{\left(
l\right) }$ possess closed loops, and a set of $\gamma _{l}$ FCs can be
introduced on $\Gamma _{\left( l\right) }$. These FCs are denoted with $%
\Lambda _{\lambda \left( l\right) },\lambda =1,\ldots ,\gamma _{l}$.

In almost all PKM with complex limbs, the FCs $\Lambda _{\left( l\right) }$
within a limb have at most one body in common ($\Lambda _{\left( l\right) }$
are edge-disjoint). Such limbs are called \emph{hybrid}, as they can be
regarded as serial connection of the kinematic loops. Throughout the paper,
the PKM are assumed to possess hybrid limbs only.

A spanning tree on $\Gamma _{\left( l\right) }$, denoted $G_{\left( l\right)
}$, is obtained by removing one edge (called \emph{cut-edge}) of each FC.
This defines a tree-topology system comprising $\mathfrak{n}_{l}$ moving
bodies and $\mathfrak{n}_{l}$ joints (tree-edges). From any vertex (body) to
the root (ground with index 0) there is a unique path in $G_{\left( l\right)
}$. A \emph{ground-directed spanning tree} $\vec{G}_{\left( l\right) }$ is
then introduced by directing all edges of $G_{\left( l\right) }$ so to point
toward the ground within this path (Fig. \ref{figIRSBotGraph}c). In the so
constructed $\vec{G}_{\left( l\right) }$, there is a unique directed path
from any vertex (moving body) to the ground. In particular, there is a path
from platform to ground, and the platform motion is determined by the motion
of the corresponding kinematic chain. The latter must indeed respect the
loop constraints imposed by the kinematic loops. This induces an ordering,
where $j\prec _{l}i$ denotes that body $j$ is contained in the path from
body $i$ to the ground. Edges of $\Gamma _{\left( l\right) }$ represent
joints with general DOF, but are often used to represent 1-DOF joints, which
are used to model multi-DOF joints. Edges of $G_{\left( l\right) }$ are
called tree-joints. The total number of tree-joint variables of limb $l$ is
denoted with $n_{l}\geq \mathfrak{n}_{l}$. In the following, a canonical
numbering is assumed \cite{JainMUBO2011_1,JainBook}, i.e. $j\prec _{l}i$
implies $j<i$. 
\newpage%

\section{Kinematics of a Limb%
\label{secLimbKin}%
}

The kinematics of the PKM is completely described by separately describing
the kinematics of each limb with the platform attached. In the following,
the limbs are regarded separated, i.e. the platform is only connected to one
limb, as in fig. \ref{figIRSBotGraph}b) and c).

\subsection{Kinematics of Tree-Topology Limbs}

The tree-joints determine the configurations of all bodies, including the
platform (they are indeed subjected to the closure constraints). The
tree-system comprises $\mathfrak{n}_{l}$ bodies and joints. The motion of
the $\mathfrak{n}_{l}$ joints is parameterized by $n_{l}$ joint variables,
which are summarized in the vector $%
\mathbold{\vartheta}%
_{\left( l\right) }:=\left( \vartheta _{1\left( l\right) },\ldots ,\vartheta
_{n_{l}\left( l\right) }\right) ^{T}\in {\mathbb{V}}^{n_{l}}$ (subscript $%
\left( l\right) $ indicates limb $l$). If all joints have 1 DOF, then $n_{l}=%
\mathfrak{n}_{l}$, but generally, $n_{l}\geq \mathfrak{n}_{l}$. For
instance, with the directed tree in fig. \ref{figIRSBotGraph}b), the motion
of body 5 of the IRSBot-2 is determined by the joints 1,2, and 5; the
platform motion is determined by joints 1,2,4, and 6. The vector of $n_{l}=9$
tree-joint variables of limb $l$ is $%
\mathbold{\vartheta}%
_{\left( l\right) }:=\left( \vartheta _{1\left( l\right) },\ldots ,\vartheta
_{9\left( l\right) }\right) ^{T}=(\varphi _{1},\varphi _{2},\varphi
_{3},\varphi _{4,1},\varphi _{4,2},\varphi _{5,1},\varphi _{5,2},\varphi
_{6,1},\varphi _{6,2})^{T}$ where, $\varphi _{1},\varphi _{2},\varphi _{3}$
are the three revolute joint angles, and $\varphi _{i,1},\varphi
_{i,2},i=4,5,6$ are the angles assigned to the three universal joints.

The pose of body $k$ of limb $l$, i.e. of a body-fixed frame $\mathcal{F}%
_{k\left( l\right) }$, is represented by the homogenous transformation
matrix $\mathbf{C}_{k\left( l\right) }\in SE\left( 3\right) $ transforming
from $\mathcal{F}_{k\left( l\right) }$ to the inertial frame (IFR) $\mathcal{%
F}_{0}$. Denote with $\mathbf{Y}_{i},i=1,\ldots n_{l}$ the screw coordinate
vector associated to the joint variable $i$, represented in the IFR. Then
the pose of body $k$ is determined by%
\begin{equation}
\mathbf{C}_{k\left( l\right) }(%
\mathbold{\vartheta}%
_{\left( l\right) })=f_{k\left( l\right) }(%
\mathbold{\vartheta}%
_{\left( l\right) })\mathbf{A}_{k\left( l\right) }  \label{Ckl}
\end{equation}%
with the product of exponentials \cite%
{Brockett1984,ModernRobotics,MUBOScrew2} (omitting subscript $\left(
l\right) $)%
\begin{equation}
f_{k}(%
\mathbold{\vartheta}%
)=\exp \left( \vartheta _{\underline{k}}\mathbf{Y}_{\underline{k}}\right)
\cdot \ldots \cdot \exp \left( \vartheta _{k-1}\mathbf{Y}_{k-1}\right) \exp
\left( \vartheta _{k}\mathbf{Y}_{k}\right)  \label{fk}
\end{equation}%
where $\mathbf{A}_{k\left( l\right) }=\mathbf{C}_{k\left( l\right) }(\mathbf{%
0})\in SE\left( 3\right) $ is the zero reference configuration of body $k$
in limb $l$, and $\underline{k}$ is the index of the last joint in the chain
from body $k$ to the ground (with the canonical ordering, body $\underline{k}
$ is the predecessor of $k$ with the lowest index). The product of
exponentials can also be expressed as 
\begin{equation}
f_{k}(%
\mathbold{\vartheta}%
)=\mathbf{B}_{\underline{k}}\exp \left( \vartheta _{\underline{k}}\mathbf{X}%
_{\underline{k}}\right) \cdot \ldots \cdot \mathbf{B}_{k-1}\exp \left(
\vartheta _{k-1}\mathbf{X}_{k-1}\right) \mathbf{B}_{k}\exp \left( \vartheta
_{k}\mathbf{X}_{k}\right)  \label{fkBody}
\end{equation}%
where $\mathbf{B}_{k\left( l\right) }\in SE\left( 3\right) $ is the
reference configuration of body $k$ relative to its predecessor within limb $%
l$, and $\mathbf{X}_{k}$ is the screw coordinate vector of joint $k$
represented in body-fixed frame at body $k$. The two forms of the POE are
related by $\mathbf{Y}_{k}=\mathbf{Ad}_{\mathbf{C}_{k}}\mathbf{X}_{k}$.

The pose of the platform (index P) is thus%
\begin{equation}
\mathbf{C}_{\mathrm{p}\left( l\right) }(%
\mathbold{\vartheta}%
_{\left( l\right) })=\psi _{\mathrm{p}\left( l\right) }(%
\mathbold{\vartheta}%
_{\left( l\right) })  \label{Cp}
\end{equation}%
with $\psi _{\mathrm{p}\left( l\right) }(%
\mathbold{\vartheta}%
_{\left( l\right) }):=f_{\mathrm{p}\left( l\right) }(%
\mathbold{\vartheta}%
_{\left( l\right) })\mathbf{A}_{\mathrm{p}}$. Note that the reference
configuration of the platform is the same for all limbs.

The body-fixed twist $\mathbf{V}_{k\left( l\right) }=\left( {^{k}}%
\bm{\omega}%
_{k\left( l\right) },{^{k}}\mathbf{v}_{k\left( l\right) }\right) ^{T}$ of
body $k$ in limb $l$ is determined by the geometric Jacobian as%
\begin{equation}
\mathbf{V}_{k\left( l\right) }=\mathbf{J}_{k\left( l\right) }\dot{%
\mathbold{\vartheta}%
}_{\left( l\right) }.  \label{Vk}
\end{equation}%
The geometric Jacobian $\mathbf{J}_{k}$ can be expressed explicitly in terms
of the screw coordinate vectors $\mathbf{Y}_{i}$ or $\mathbf{X}_{i}$ \cite%
{MMT2022,ModernRobotics,MUBOScrew1,Murray}. It is already available as the
block row of the system Jacobian $\mathsf{A}_{\left( l\right) }$, as
described next.

Denote with $\mathsf{V}_{\left( l\right) }:=\left( \mathbf{V}_{1},\cdots ,%
\mathbf{V}_{n_{l}}\right) \in {\mathbb{R}}^{6n_{l}}$ the system twist vector
of limb $l$. In terms of the tree-joint rates $\dot{%
\mathbold{\vartheta}%
}_{\left( l\right) }$, this is

\begin{equation}
\mathsf{V}_{\left( l\right) }=\mathsf{J}_{\left( l\right) }\dot{%
\mathbold{\vartheta}%
}_{\left( l\right) },\ \ \ \mathrm{with\ \ }\mathsf{J}_{\left( l\right)
}=\left( 
\begin{array}{c}
\mathbf{J}_{1\left( l\right) } \\ 
\vdots \\ 
\mathbf{J}_{n_{l}\left( l\right) }%
\end{array}%
\right)  \label{VSys}
\end{equation}%
with the \emph{geometric system Jacobian }$\mathsf{J}_{\left( l\right) }$.
The latter factorizes as%
\begin{equation}
\mathsf{J}_{\left( l\right) }=\mathsf{A_{\left( l\right) }X}_{\left(
l\right) }.  \label{JbSys}
\end{equation}%
Assuming a canonically directed spanning tree $\vec{G}_{\left( l\right) }$,
matrix $\mathsf{A}_{\left( l\right) }\left( 
\mathbold{\vartheta}%
\right) $ is the block-triangular, and $\mathsf{X}_{\left( l\right) }$ is
the block-diagonal%
\begin{equation}
\mathsf{A}_{\left( l\right) }=\left( 
\begin{array}{ccccc}
\mathbf{I} & \mathbf{0} & \mathbf{0} & \cdots & \mathbf{0} \\ 
& \mathbf{I} & \mathbf{0} & \cdots & \mathbf{0} \\ 
&  &  & \ddots & \vdots \\ 
& \mathbf{Ad}_{\mathbf{C}_{i,j}} &  &  & \mathbf{0} \\ 
&  &  &  & \mathbf{I}%
\end{array}%
\right) _{%
\hspace{-1ex}%
\left( l\right) }%
\hspace{-1ex}%
,\ \mathsf{X}_{\left( l\right) }=\left( 
\begin{array}{ccccc}
{^{1}\mathbf{X}}_{1} & \mathbf{0} & \mathbf{0} &  & \mathbf{0} \\ 
\mathbf{0} & {^{2}\mathbf{X}}_{2} & \mathbf{0} & \cdots & \mathbf{0} \\ 
\mathbf{0} & \mathbf{0} & {^{3}\mathbf{X}}_{3} &  & \mathbf{0} \\ 
\vdots & \vdots & \ddots & \ddots &  \\ 
\mathbf{0} & \mathbf{0} & \cdots & \mathbf{0} & {^{n_{l}}}\mathbf{X}_{n_{l}}%
\end{array}%
\right) _{%
\hspace{-1ex}%
\left( l\right) }  \label{Ab}
\end{equation}%
where $\mathbf{Ad}_{\mathbf{C}_{i,j}}=\mathbf{0}$ if $j\npreceq i$ (body $j$
is not a predecessor of body $i$).

The time derivative of the Jacobian can be expressed in closed form \cite%
{MMT2022,MUBOScrew1,MUBO2021} as%
\begin{equation}
\dot{\mathsf{J}}(%
\mathbold{\vartheta}%
,\dot{%
\mathbold{\vartheta}%
})=-\mathsf{A}\left( 
\mathbold{\vartheta}%
\right) \mathsf{a}\left( \dot{%
\mathbold{\vartheta}%
}\right) \mathsf{J}\left( 
\mathbold{\vartheta}%
\right)  \label{JSysDot}
\end{equation}%
with $\mathsf{a}(\dot{%
\mathbold{\vartheta}%
}):=\mathrm{diag\,}(\dot{\vartheta}_{1}\mathbf{ad}_{{{^{1}}\mathbf{X}_{1}}%
},\ldots ,\dot{\vartheta}_{n_{l}}\mathbf{ad}_{{{^{n}}\mathbf{X}_{n}}})$,
where $\mathbf{ad}_{{\mathbf{X}}}\mathbf{Y}$ is the adjoint operator matrix
expressing the Lie bracket of two screws. The system acceleration of limb $l$
is thus 
\begin{equation}
\dot{\mathsf{V}}_{\left( l\right) }=\mathsf{J}_{\left( l\right) }\ddot{%
\mathbold{\vartheta}%
}_{\left( l\right) }+\dot{\mathsf{J}}_{\left( l\right) }\dot{%
\mathbold{\vartheta}%
}_{\left( l\right) }.  \label{AccSys}
\end{equation}

\begin{remark}
The Lie group formulation has several advantages. Firstly it is important to
notice that the kinematic topology enters in the description via the matrix $%
\mathsf{A}_{\left( l\right) }$ only. This is common feature of all matrix
formulations, which is also represented in the more general setting of
spatial operator algebra \cite{JainNODY2012,JainMUBO2011_1,JainBook}. Its
main advantage, however, is that it combines the classical zero-reference
formulation \cite{Gupta1986} with the coordinate invariant geometric
modeling in terms of joint screws \cite{PloenPark1999}. This coordinate
invariance admits using the screw coordinates in the reference
configuration, likewise represented in the body-fixed or the spatial
inertial frame. The kinematic relations is then computationally efficient
expressed by the POE (\ref{fk}) or (\ref{fkBody}) as the exponential on $%
SE\left( 3\right) $ admits efficient implementations \cite{RSPA2021}.
\end{remark}

\subsection{Task Space Velocity}

The constraints imposed by the $L$ limbs determine the mobility of the PKM
platform. The platform DOF is denoted $\delta _{\mathrm{p}}\leq 6$. For
describing the tasks, $\delta _{\mathrm{p}}$ components of the platform
twist $\mathbf{V}_{\mathrm{p}}$ are used. These components form the \emph{%
task space velocity} $\mathbf{V}_{\mathrm{t}}\in {\mathbb{R}}^{\delta _{%
\mathrm{p}}}$. The platform twist is usually represented in the platform
frame $\mathcal{F}_{\mathrm{p}}$ (collocated with the end-effector), and is
aligned so that $\mathbf{V}_{\mathrm{t}}$ consists of exactly $\delta _{%
\mathrm{p}}$ components of the vector $\mathbf{V}_{\mathrm{p}}$. This is
expressed as%
\begin{equation}
\mathbf{V}_{\mathrm{p}}=\mathbf{P}_{\mathrm{p}}\mathbf{V}_{\mathrm{t}}
\label{VpVt}
\end{equation}%
with a unimodular $6\times \delta _{\mathrm{p}}$ velocity distribution
matrix $\mathbf{P}_{\mathrm{p}}$, assigning the $\delta _{\mathrm{p}}$
components of the task space velocity to the components of the platform
twist. Typical choices for often used task spaces can be found in \cite%
{MMT2022}. If the task velocity allocation changes with the task motion,
matrix $\mathbf{P}_{\mathrm{p}}$ will be configuration dependent.

\subsection{Loop Constraints within Hybrid Limbs}

When limbs separated, only the kinematic loops within the individual limbs
remain. Limb $l$ possesses $\gamma _{l}$ FCs, denoted $\Lambda _{\lambda
\left( l\right) },\lambda =1,\ldots ,\gamma _{l}$, and for each FC, a system
of loop closure constraints is introduced, referred to as \emph{intra-limb
constraints}. The number of constraints imposed by FC $\Lambda _{\lambda
\left( l\right) }$ is denoted with $m_{\lambda ,l}$. It is assumed in the
following that these $m_{\lambda ,l}$ constraints are independent. Limbs
where this is not satisfied are overconstrained, and must be dealt with
separately (see remark \ref{remRedConstr} and \ref{remRedConstr2}).

The tree $G_{\left( l\right) }$ is obtained by removing one edge from each
FC. That is, each FC is cut open, and the joint corresponding to the
eliminated edge is called the \emph{cut-joint} of the FC \cite%
{NikraveshBook1988}. Loop closure then imposes constraints on the tree-joint
variables $%
\mathbold{\vartheta}%
_{\left( l\right) }$ of limb $l$. More precisely, for hybrid limbs the FC
are edge-disjoint, so that loop closure of $\Lambda _{\lambda \left(
l\right) }$ only constrains the $n_{\lambda ,l}$ tree-joint variables
belonging to this FC. The latter are denoted with $%
\mathbold{\vartheta}%
_{\left( \lambda ,l\right) }\in {\mathbb{V}}^{n_{\lambda ,l}}$. The
cut-joint of FC $\Lambda _{\lambda \left( l\right) }$ in limb $l$ gives rise
to $m_{\lambda ,l}$ geometric, velocity, and acceleration constraints of the
form%
\begin{eqnarray}
\mathbf{g}_{\left( \lambda ,l\right) }(%
\mathbold{\vartheta}%
_{\left( \lambda ,l\right) }) &=&\mathbf{0}  \label{GeomConsLoopCutJoint} \\
\mathbf{G}_{\left( \lambda ,l\right) }\dot{%
\mathbold{\vartheta}%
}_{\left( \lambda ,l\right) } &=&\mathbf{0}  \label{VelConsLoopCutJoint} \\
\mathbf{G}_{\left( \lambda ,l\right) }\ddot{%
\mathbold{\vartheta}%
}_{\left( \lambda ,l\right) }+\dot{\mathbf{G}}_{\left( \lambda ,l\right) }%
\dot{%
\mathbold{\vartheta}%
}_{\left( \lambda ,l\right) } &=&\mathbf{0}.  \label{AccConsLoopCutJoint}
\end{eqnarray}%
The solution variety of the geometric constraints of $\Lambda _{\lambda
\left( l\right) }$ is $V_{\left( \lambda ,l\right) }:=\{%
\mathbold{\vartheta}%
_{\left( \lambda ,l\right) }\in {\mathbb{V}}^{n_{\lambda ,l}}|\mathbf{g}%
_{\left( \lambda ,l\right) }(%
\mathbold{\vartheta}%
_{\left( \lambda ,l\right) })=\mathbf{0}\}$, which has dimension $\delta
_{\lambda ,l}=n_{\lambda ,l}-m_{\lambda ,l}$. The configuration space of the
separated hybrid limb $l$ is $V_{l}:=\{%
\mathbold{\vartheta}%
_{\left( l\right) }\in {\mathbb{V}}^{n_{l}}|\mathbf{g}_{\left( \lambda
,l\right) }(%
\mathbold{\vartheta}%
_{\left( \lambda ,l\right) })=\mathbf{0},\lambda =1,\ldots ,\gamma
_{l}\}\subset {\mathbb{V}}^{n_{l}}$. Its dimension is denoted with $\delta
_{l}$, which is the DOF of the limb when separated from the PKM.

\subsection{Constraint Resolution}

The constraints for $\Lambda _{\lambda \left( l\right) }$ involve the $%
n_{\lambda ,l}$ coordinates $%
\mathbold{\vartheta}%
_{\left( \lambda ,l\right) }$ within this FC only, i.e. $g_{\left( \lambda
,l\right) }$ depends on $%
\mathbold{\vartheta}%
_{\left( \lambda ,l\right) }$ only. Moreover, when the hybrid limb $l$ is
connected to the platform only, the constraints can be solved for the $%
\gamma _{l}$ FC independently. This will be referred to as \emph{local
constraint embedding}, and is a main step in the proposed modeling approach.

\subparagraph{Geometric Constraints:}

The submersion theorem ensures that the intra-limb constraints (\ref%
{GeomConsLoopCutJoint}) possess a local solution $%
\mathbold{\vartheta}%
_{\left( \lambda ,l\right) }=\psi _{\left( \lambda ,l\right) }(\mathbf{q}%
_{\left( \lambda ,l\right) })$ in terms of $\delta _{\lambda ,l}:=n_{\lambda
,l}-m_{\lambda ,l}$ independent coordinates $\mathbf{q}_{\left( \lambda
,l\right) }\in M_{\lambda ,l}\subset {\mathbb{V}}^{\delta _{\lambda ,l}}$.
The $\mathbf{q}_{\left( \lambda ,l\right) }$ serve as generalized
coordinates with coordinate manifold $M_{\lambda ,l}$. The solution, i.e.
the map $\psi _{\left( \lambda ,l\right) }:M_{\lambda ,l}\subset {\mathbb{V}}%
^{\delta _{\lambda ,l}}\rightarrow V_{\lambda ,l}\subset {\mathbb{V}}%
^{n_{\lambda ,l}}$, can generally not be determined in closed form. A
computational method is introduced in sec. \ref{secSolLoop} to this end.

There may be $\delta _{0,l}$ joint variables, denoted with $\mathbf{q}%
_{\left( 0,l\right) }\in {\mathbb{V}}^{\delta _{0,l}}$, that are not
included in any FC, and thus serve as further independent generalized
coordinates. The $\mathbf{q}_{\left( 0,l\right) }$ along with $\mathbf{q}%
_{\left( \lambda ,l\right) },\lambda =1,\ldots ,\gamma _{l}$ represents the
overall set of $\delta _{l}:=\delta _{0,l}+\delta _{1,l}+\ldots +\delta
_{\gamma _{l},l}$ generalized coordinates for the separated hybrid limb,
denoted $\mathbf{q}_{\left( l\right) }$. The configuration of limb $l$ is
thus expressed in terms of $\delta _{l}$ independent coordinates $\mathbf{q}%
_{\left( l\right) }\in M_{l}\subset {\mathbb{V}}^{\delta _{l}}$ with the 
\emph{solution map} 
\begin{equation}
\begin{array}{ll}
\psi _{\left( l\right) }: & M_{l}\rightarrow V_{l} \\ 
& 
\hspace{-0.5ex}%
\mathbf{q}_{\left( l\right) }\mapsto 
\mathbold{\vartheta}%
_{\left( l\right) }=\psi _{\left( l\right) }(\mathbf{q}_{\left( l\right) })%
\end{array}
\label{solConL}
\end{equation}%
which summarizes the $\psi _{\left( \lambda ,l\right) },\lambda =1,\ldots
,\gamma _{l}$.

\subparagraph{Velocity and Acceleration Constraints:}

The coordinates are partitioned as $%
\mathbold{\vartheta}%
_{\left( \lambda ,l\right) }=(\mathbf{y}_{\left( \lambda ,l\right) },\mathbf{%
q}_{\left( \lambda ,l\right) })$, where $\mathbf{y}_{\left( \lambda
,l\right) }$ are the remaining $m_{\lambda ,l}$ dependent joint variables in
the FC. The constraints (\ref{VelConsLoopCutJoint}) can then be written as%
\begin{equation}
\mathbf{G}_{\mathbf{y}\left( \lambda ,l\right) }\dot{\mathbf{y}}_{\left(
\lambda ,l\right) }+\mathbf{G}_{\mathbf{q}\left( \lambda ,l\right) }\dot{%
\mathbf{q}}_{\left( \lambda ,l\right) }=\mathbf{0}  \label{GyGq}
\end{equation}%
with the regular $m_{\lambda ,l}\times m_{\lambda ,l}$ submatrix $\mathbf{G}%
_{\mathbf{y}\left( \lambda ,l\right) }$ and the $m_{\lambda ,l}\times \delta
_{\lambda ,l}$ submatrix $\mathbf{G}_{\mathbf{q}\left( \lambda ,l\right) }$
corresponding to the dependent joint rates $\dot{\mathbf{y}}_{\left( \lambda
,l\right) }$ and to the independent joint rates $\dot{\mathbf{q}}_{\left(
\lambda ,l\right) }$, respectively. With the assumed independence of
constraints, the constraint Jacobian $\mathbf{G}_{\left( \lambda ,l\right) }(%
\mathbold{\vartheta}%
_{\left( \lambda ,l\right) })$ is a regular $m_{\lambda ,l}\times n_{\lambda
,l}$ matrix. The solution of velocity and acceleration constraints (\ref%
{VelConsLoopCutJoint}) and (\ref{AccConsLoopCutJoint}) are, respectively,%
\begin{equation}
\left( 
\begin{array}{c}
\dot{\mathbf{y}}_{\left( \lambda ,l\right) } \\ 
\dot{\mathbf{q}}_{\left( \lambda ,l\right) }%
\end{array}%
\right) =\mathbf{H}_{\left( \lambda ,l\right) }\dot{\mathbf{q}}_{\left(
\lambda ,l\right) },\ \ \left( 
\begin{array}{c}
\ddot{\mathbf{y}}_{\left( \lambda ,l\right) } \\ 
\ddot{\mathbf{q}}_{\left( \lambda ,l\right) }%
\end{array}%
\right) =\mathbf{H}_{\left( \lambda ,l\right) }\ddot{\mathbf{q}}_{\left(
\lambda ,l\right) }+\dot{\mathbf{H}}_{\left( \lambda ,l\right) }\dot{\mathbf{%
q}}_{\left( \lambda ,l\right) }.  \label{eta2dH}
\end{equation}%
The $n_{\lambda ,l}\times \delta _{\lambda ,l}$ matrix $\mathbf{H}_{\left(
\lambda ,l\right) }$ is an orthogonal complement of $\mathbf{G}_{\left(
\lambda ,l\right) }$. With the partitioning, it and its derivative is
explicitly%
\begin{align}
\mathbf{H}_{\left( \lambda ,l\right) }& :=\left( 
\begin{array}{c}
-\mathbf{G}_{\mathbf{y}}^{-1}\mathbf{G}_{\mathbf{q}} \\ 
\mathbf{I}%
\end{array}%
\right) _{\left( \lambda ,l\right) }  \label{Hlambda2} \\
\dot{\mathbf{H}}_{\left( \lambda ,l\right) }(%
\mathbold{\vartheta}%
_{\left( \lambda ,l\right) },\dot{%
\mathbold{\vartheta}%
}_{\left( \lambda ,l\right) })& =\left( 
\begin{array}{c}
\mathbf{G}_{\mathbf{y}}^{-1}(\dot{\mathbf{G}}_{\mathbf{y}}\mathbf{G}_{%
\mathbf{y}}^{-1}\mathbf{G}_{\mathbf{q}}%
\hspace{-0.5ex}%
-%
\hspace{-0.5ex}%
\dot{\mathbf{G}}_{\mathbf{q}}) \\ 
\mathbf{0}%
\end{array}%
\right) _{%
\hspace{-0.5ex}%
\left( \lambda ,l\right) }.  \label{Hldot}
\end{align}%
Combining (\ref{eta2dH}) for all FCs yields the overall solution of the
intra-limb constraints 
\begin{equation}
\dot{%
\mathbold{\vartheta}%
}_{\left( l\right) }=\mathbf{H}_{\left( l\right) }\dot{\mathbf{q}}_{\left(
l\right) },\ \ \ddot{%
\mathbold{\vartheta}%
}_{\left( l\right) }=\mathbf{H}_{\left( l\right) }\ddot{\mathbf{q}}_{\left(
l\right) }+\dot{\mathbf{H}}_{\left( l\right) }\dot{\mathbf{q}}_{\left(
l\right) }.  \label{Hl}
\end{equation}

\begin{remark}
It should be noticed that $\mathbf{q}_{\left( \lambda ,l\right) }$ are local
coordinates on the solution variety $V_{\lambda ,l}$. If the PKM exhibits
different motion modes, it may be necessary to switch between coordinates
and different $M_{\lambda ,l}$ accordingly. Moreover, in case of
kinematotropic PKM \cite{HustyZsomborMurray1994,Wohlhart1996} the DOF $%
\delta _{l}$ is different in different motion modes.
\end{remark}

\begin{remark}
\label{remRedConstr}%
The selection of independent velocities (coordinates) is not unique. A
particular choice of $\mathbf{q}_{\left( \lambda ,l\right) }$ corresponds to
a particular choice of $\mathbf{G}_{\mathbf{y}\left( \lambda ,l\right) }$.
The choice determines the numerical conditioning of the matrix $\mathbf{H}%
_{\left( \lambda ,l\right) }$. This problem was addressed in \cite%
{WehageHaug1982,Nikravesh1985,Wehage2015}, were a generalized coordinate
partitioning approach was introduced. Computational aspects were addressed
from a geometric perspective in \cite{Blajer1994,Terze2010,Blajer2011}. In
general, an orthogonal complement of $\mathbf{G}_{\left( \lambda ,l\right) }$%
, and thus $\dot{\mathbf{q}}_{\left( \lambda ,l\right) }$, can be determined
numerically using SVD or QR decompositions to compute the null-space of $%
\mathbf{G}_{\left( \lambda ,l\right) }$. Besides numerical approaches, a
solution of the velocity constraints can be obtained with the reciprocal
screw approach \cite{AngelesBook,MMT2022}. This allows for algebraic
determination, possible avoiding matrix inversions.
\end{remark}

\begin{remark}
\label{remRedConstr2}%
A problem often encountered, is that of constraint redundancy, i.e. some of
the $m_{\lambda ,l}$ constraints are dependent, so that the DOF of $\Lambda
_{\lambda \left( l\right) }$ is $\delta _{\lambda ,l}>n_{\lambda
,l}-m_{\lambda ,l}$, which leads to a singular constraint Jacobian. The FC
is then classified as overconstrained. For most PKM with complex limbs, the
intra-limb constraints are not redundant (see sec. \ref{secCompare}). The
3R[2UU] Delta robot is an example where the intra-limb constraints are
redundant. (The notation 3R[2UU] indicates that the PKM has three limbs,
articulated arm is attached to the base by an actuated R joint, and the arm
and platform are connected by two parallel rods with U joints at both ends).
A limb contains a single loop with four U joints. The loop constraints are
redundant since the axes of the different U joints are parallel.
\end{remark}

\begin{remark}
With the cut-joint formulation, the number of the constraints, and thus the
number of tree-joint variable (i.e. the size of the EOM of a limb), depend
on the selected cut-joint giving rise to specific cut-joint constraints (see
appendix \ref{secCutConstr}). The modeling and constraint formulation could
be simplified by using the \emph{cut-body approach}. Then, the loop
constraints are simply expressed as $f_{\lambda ,l}(%
\mathbold{\vartheta}%
)=\mathbf{I}$, with%
\begin{equation*}
f_{\lambda ,l}(%
\mathbold{\vartheta}%
)=\exp \left( \vartheta _{1}\mathbf{Y}_{1}\right) \cdot \ldots \cdot \exp
\left( \vartheta _{N_{\lambda ,l}}\mathbf{Y}_{N_{\lambda ,l}}\right)
\end{equation*}%
where $\vartheta _{1},\ldots ,\vartheta _{N_{\lambda ,l}}$ denote the joint
angles of all $N_{\lambda ,l}$ joints in the FC $\Lambda _{\lambda \left(
l\right) }$ (including the cut-joint). This, however, always leads to a
system of six equations for all $N_{\lambda ,l}$ joint variables, and thus
to a larger system of constraints and a larger system of EOM. On the other
hand, the computation of the constraints is always the same irrespective of
the type of joints. This approach is usually followed in mechanism theory,
while the cut-joint formulation is exclusively used in MBS dynamics.
\end{remark}

\subsection{Forward Kinematics of Separated Limb}

The forward kinematics problem of limb $l$ is to determine the motion of all
bodies in terms of the generalized coordinates $\mathbf{q}_{\left( l\right)
} $. The configuration of body $k$ of limb $l$ is given in terms of the
tree-joint variables by (\ref{Ckl}), and that of the platform by (\ref{Cp})
with the map $\psi _{\mathrm{p}\left( l\right) }$. Introducing the solution (%
\ref{solConL}) of the intra-limb constraints, the platform motion as part of
the complex limb is given in terms of the generalized coordinates as $%
\mathbf{C}_{\mathrm{p}\left( l\right) }(%
\mathbold{\vartheta}%
_{\left( l\right) })=\varphi _{\mathrm{p}\left( l\right) }(\mathbf{q}%
_{\left( l\right) })$ with the \emph{forward kinematics map of limb} $l$, $%
\varphi _{\mathrm{p}\left( l\right) }:M_{\left( l\right) }\rightarrow
W_{\left( l\right) }$, defined as%
\begin{equation}
\varphi _{\mathrm{p}\left( l\right) }:=\psi _{\mathrm{p}\left( l\right)
}\circ \psi _{\left( l\right) }.  \label{phipl}
\end{equation}%
The forward kinematics map is surjective, and the image of $\varphi _{%
\mathrm{p}\left( l\right) }$ is the workspace $W_{\left( l\right) }\subset
SE\left( 3\right) $ of the separated complex limb. The platform DOF of the
separated limb, denoted with $\delta _{\mathrm{p}\left( l\right) }:=\dim
W_{\left( l\right) }$, is the dimension of the workspace $W_{\left( l\right)
}=\mathrm{im\,}\varphi _{\mathrm{p}\left( l\right) }$ of the limb. As for
serial robotic manipulators, a limb can be kinematically redundant. If the
DOF of the limb is greater than its platform DOF, $\delta _{\mathrm{p}\left(
l\right) }<\delta _{l}$, the \emph{limb is structurally redundant}, i.e. the
limb can perform finite motions even if the platform is locked.

Combining the constraint solution (\ref{Hl}) with the geometric Jacobian (%
\ref{Vk}) of the platform, yields the \emph{forward kinematics Jacobian of
limb} $l$ 
\begin{equation}
\mathbf{L}_{\mathrm{p}\left( l\right) }:=\mathbf{J}_{\mathrm{p}\left(
l\right) }\mathbf{H}_{\left( l\right) }  \label{Vp}
\end{equation}%
which determines the EE-twist as $\mathbf{V}_{\mathrm{p}\left( l\right) }=%
\mathbf{L}_{\mathrm{p}\left( l\right) }\dot{\mathbf{q}}_{\left( l\right) }$.
The rank of $\mathbf{L}_{\mathrm{p}\left( l\right) }$ is $\delta _{\mathrm{p}%
\left( l\right) }$.

For deriving the numerical constraint embedding, it will be helpful to
represent the forward kinematics relation by the following commutation
diagram.

\centerline{\includegraphics[height=2.6cm]{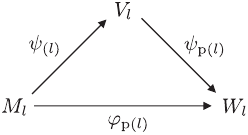}}

\subsection{Inverse Kinematics}

The inverse kinematics problem of the PKM mechanism is to determine the
motion of all bodies, i.e. the joint angles $%
\mathbold{\vartheta}%
_{\left( l\right) }\left( t\right) $, for a given platform motion of the
PKM. In the following, the limbs are assumed to be non-redundant, i.e. $%
\delta _{\mathrm{p}\left( l\right) }=\delta _{l}$.

The inverse of $\varphi _{\mathrm{p}\left( l\right) }$ exists, and $\varphi
_{\mathrm{p}\left( l\right) }^{-1}:W_{\left( l\right) }\rightarrow M_{\left(
l\right) }$ determines the generalized coordinates for given platform pose.
For general PKM, also with non-redundant limbs, the solution of the inverse
kinematics problem of the limb is not unique, since the map $\varphi _{%
\mathrm{p}\left( l\right) }$ is surjective but not bijective, which must be
accounted for when solving the inverse kinematics problem numerically.

When all limbs are assembled, the platform must respect the corresponding
constraints imposed by all $L$ limbs, thus the workspace of the PKM is $%
W=\cap _{l=1}^{L}W_{\left( l\right) }$. The DOF of the platform as part of
the PKM is $\delta _{\mathrm{p}}=\dim W$. It will be important to
distinguish whether the platform DOF of the PKM is lower than the platform
DOF of the separated limb. Limb $l$ is called \emph{equimobile} iff $\delta
_{\mathrm{p}}=\delta _{\mathrm{p}\left( l\right) }$ \cite%
{MuellerAMR2020,MMT2022}, which means that the PKM workspace is an
equidimensional subspace of the workspaces of the limbs. The inverse map $%
\varphi _{\mathrm{p}\left( l\right) }^{-1}$ restricted to $W$ is called the 
\emph{inverse kinematics map of limb} $l$, denoted $f_{\left( l\right)
}:=\varphi _{\mathrm{p}\left( l\right) }^{-1}|_{W}$%
\begin{equation}
\begin{array}{ll}
f_{\left( l\right) }: & W\rightarrow M_{\left( l\right) } \\ 
& 
\hspace{-0.5ex}%
\mathbf{C}_{\mathrm{p}}\mapsto \mathbf{q}_{\left( l\right) }=f_{\left(
l\right) }(\mathbf{C}_{\mathrm{p}})%
\end{array}
\label{fl}
\end{equation}%
which assigns to an admissible platform pose of the PKM the generalized
coordinates of limb $l$. Solving the inverse kinematics problem thus boils
down to evaluating $f_{\left( l\right) }$. This must be performed locally
due to the non-uniqueness. The inverse kinematics problem is solved, i.e.
the map (\ref{fl}) is evaluated, numerically employing the velocity inverse
kinematics.

Consider limb $l$ separated from the PKM but with the platform attached. The
platform twist is determined by $\mathbf{V}_{\mathrm{p}\left( l\right) }=%
\mathbf{L}_{\mathrm{p}\left( l\right) }\dot{\mathbf{q}}_{\left( l\right) }$,
with the Jacobian (\ref{Vp}). The platform attached to a separated limb has $%
\delta _{\mathrm{p}\left( l\right) }$ DOF, and the forward kinematics
Jacobian $\mathbf{L}_{\mathrm{p}\left( l\right) }$ has rank $\delta _{%
\mathrm{p}\left( l\right) }$. Denote with $\mathbf{L}_{\mathrm{t}\left(
l\right) }$ the \emph{task space Jacobian of limb} $l$ defined as the full
rank $\delta _{\mathrm{p}\left( l\right) }\times \delta _{l}$ submatrix of $%
\mathbf{L}_{\mathrm{p}\left( l\right) }$ so that $\mathbf{L}_{\mathrm{t}%
\left( l\right) }\dot{\mathbf{q}}_{\left( l\right) }$ delivers the $\delta _{%
\mathrm{p}\left( l\right) }$ components of the platform velocity when
attached to the separated limb, similarly to the forward kinematics Jacobian
of a serial robot.

Now consider the PKM with all limbs assembled. For an equimobile limb, the
task space velocity $\mathbf{V}_{\mathrm{t}}$ comprises exactly $\delta _{%
\mathrm{p}\left( l\right) }=\delta _{\mathrm{p}}$ components of $\mathbf{V}_{%
\mathrm{p}\left( l\right) }$, while for a non-equimobile PKM, it only
consists of $\delta _{\mathrm{p}}<\delta _{\mathrm{p}\left( l\right) }$
components (the platform DOF of a separated limb is different from the
platform DOF within the PKM). For equimobile limbs, this implies%
\begin{equation}
\mathbf{V}_{\mathrm{t}}=\mathbf{L}_{\mathrm{t}\left( l\right) }\dot{\mathbf{q%
}}_{\left( l\right) }.  \label{Vt}
\end{equation}%
For non-equimobile limbs, in addition to the $\delta _{\mathrm{p}\left(
l\right) }$ rows of $\mathbf{L}_{\mathrm{t}\left( l\right) }$ that determine
the task velocities, there are $\delta _{\mathrm{p}\left( l\right) }-\delta
_{\mathrm{p}}$ rows of $\mathbf{L}_{\mathrm{t}\left( l\right) }$ that
correspond to the components of the platform twist of the separated limb
that must be zero when all limbs are assembled to the PKM. These constraints
on the limb along with the forward kinematics are expressed as 
\begin{equation}
\mathbf{D}_{\mathrm{t}\left( l\right) }\mathbf{V}_{\mathrm{t}}=\mathbf{L}_{%
\mathrm{t}\left( l\right) }\dot{\mathbf{q}}_{\left( l\right) }  \label{VpD}
\end{equation}%
where $\mathbf{D}_{\mathrm{t}\left( l\right) }$ is a $\delta _{\mathrm{p}%
\left( l\right) }\times \delta _{\mathrm{p}}$ velocity distribution matrix,
which assigns the components of the task space velocity $\mathbf{V}_{\mathrm{%
t}}$ to the relevant rows of the task space Jacobian of limb $l$. For
equimobile limbs, $\mathbf{D}_{\mathrm{t}\left( l\right) }$ is the identity
matrix. The platform DOF of the IRSBot-2, for example, is $\delta _{\mathrm{p%
}}=\delta =2$. The platform DOF a separated limb is $\delta _{\mathrm{p}%
\left( l\right) }=3$, however, so that it is non-equimobile (see sec. \ref%
{secIRSBot}). Relative to the platform frame $\mathcal{F}_{\mathrm{p}}$
shown in fig. \ref{figIRSBot}, the task space velocity $\mathbf{V}_{\mathrm{t%
}}=\left( v_{1},v_{3}\right) ^{T}$ consists of the 1- and 3-component of the
translation velocity, and 
\begin{equation}
\mathbf{D}_{\mathrm{t}\left( l\right) }=\left( 
\begin{array}{cc}
1 & 0 \\ 
0 & 0 \\ 
0 & 1%
\end{array}%
\right) .  \label{DtDelta}
\end{equation}%
For non-redundant limbs, $\mathbf{L}_{\mathrm{t}\left( l\right) }$ is square
and invertible. The velocity inverse kinematics solution is thus%
\begin{equation}
\dot{\mathbf{q}}_{\left( l\right) }=\mathbf{F}_{\left( l\right) }\mathbf{V}_{%
\mathrm{t}},\ \ \ \mathrm{with\ \ }\mathbf{F}_{\left( l\right) }=\mathbf{L}_{%
\mathrm{t}\left( l\right) }^{-1}\mathbf{D}_{\mathrm{t}\left( l\right) }
\label{Ftheta2}
\end{equation}%
where $\mathbf{F}_{\left( l\right) }$ is the \emph{inverse kinematics
Jacobian of limb} $l$. With (\ref{Hl}), the velocity of all joints of the
limb is then%
\begin{equation}
\dot{%
\mathbold{\vartheta}%
}_{\left( l\right) }=\mathbf{H}_{\left( l\right) }\mathbf{F}_{\left(
l\right) }\mathbf{V}_{\mathrm{t}}.  \label{HFl}
\end{equation}%
The joint accelerations are then%
\begin{eqnarray}
\ddot{%
\mathbold{\vartheta}%
}_{\left( l\right) } &=&\mathbf{H}_{\left( l\right) }\mathbf{F}_{\left(
l\right) }\dot{\mathbf{V}}_{\mathrm{t}}+(\dot{\mathbf{H}}_{\left( l\right) }-%
\mathbf{H}_{\left( l\right) }\mathbf{F}_{\left( l\right) }\dot{\mathbf{L}}_{%
\mathrm{t}\left( l\right) })\mathbf{F}_{\left( l\right) }\mathbf{V}_{\mathrm{%
t}}  \label{eta2d} \\
&=&\mathbf{H}_{\left( l\right) }\mathbf{F}_{\left( l\right) }\dot{\mathbf{V}}%
_{\mathrm{t}}+(\dot{\mathbf{H}}_{\left( l\right) }-\mathbf{H}_{\left(
l\right) }\mathbf{F}_{\left( l\right) }\dot{\mathbf{L}}_{\mathrm{t}\left(
l\right) })\dot{\mathbf{q}}_{\left( l\right) }  \notag
\end{eqnarray}%
with $\dot{\mathbf{H}}_{\left( l\right) }$ in (\ref{Hldot}), and $\dot{%
\mathbf{L}}_{\mathrm{p}\left( l\right) }(%
\mathbold{\vartheta}%
_{\left( l\right) },\dot{%
\mathbold{\vartheta}%
}_{\left( l\right) })=\dot{\mathbf{J}}_{\mathrm{p}\left( l\right) }\mathbf{H}%
_{\left( l\right) }+\mathbf{J}_{\mathrm{p}\left( l\right) }\dot{\mathbf{H}}%
_{\left( l\right) }$.

\subsection{Inverse Kinematics of PKM}

The above kinematics description of individual limbs can be immediately
employed to the kinematics modeling of the overall PKM. Denote with $%
\mathbold{\vartheta}%
_{\mathrm{act}}\in {\mathbb{V}}^{n_{\mathrm{act}}}$ the $n_{\mathrm{act}}$
joint variables of the actuated joints of the PKM. It is assumed that the
PKM is fully actuated, i.e. the forward kinematics map $\varphi _{\mathrm{FK}%
}:{\mathbb{V}}^{n_{\mathrm{act}}}\rightarrow W$ is well-defined and
surjective. It determines the platform motion for given actuator motion, $%
\mathbf{C}_{\mathrm{p}}=\varphi _{\mathrm{FK}}(%
\mathbold{\vartheta}%
_{\mathrm{act}})$. If the PKM is non-redundantly actuated ($n_{\mathrm{act}%
}=\delta $) \cite{MuellerRobotica2013}, the actuator coordinates represent
generalized coordinates $\mathbf{q}:=%
\mathbold{\vartheta}%
_{\mathrm{act}}$ of the PKM. If it is redundantly actuated ($n_{\mathrm{act}%
}>\delta $), only $\delta $ of them can be used as generalized coordinates.

The inverse kinematics problem of the PKM is to determine the actuator
motion for given platform motion. The inverse $f_{\mathrm{IK}}:=\varphi _{%
\mathrm{FK}}^{-1}$ is the \emph{inverse kinematics map of the PKM}, which
assigns the actuator coordinates to the platform configuration, $%
\mathbold{\vartheta}%
_{\mathrm{act}}=f_{\mathrm{IK}}(\mathbf{C}_{\mathrm{p}})$. This is defined
already by the inverse kinematics map $f_{l}$ in (\ref{fl}) of the
individual limbs.

The actuator velocities are also known already from (\ref{Ftheta2}). The
solution of the velocity inverse kinematics problem is thus%
\begin{equation}
\dot{%
\mathbold{\vartheta}%
}_{\mathrm{act}}=\mathbf{J}_{\mathrm{IK}}\mathbf{V}_{\mathrm{t}}  \label{IK}
\end{equation}%
with $\mathbf{J}_{\mathrm{IK}}$ being the $n_{\mathrm{act}}\times \delta _{%
\mathrm{p}}$ \emph{inverse kinematics Jacobian of the PKM}, which consists
of the rows of $\mathbf{F}_{\left( l\right) },l=1,\ldots ,L$ in (\ref%
{Ftheta2}) corresponding to the actuator coordinates.

\section{Numerical Constraint Embedding%
\label{secConsEmbedding}%
}

\subsection{Nested Solution Scheme}

Local constraint embedding refers to solving the inverse kinematics problem
of the limbs, i.e. evaluating the inverse kinematics map $f_{\left( l\right)
}$ in (\ref{fl}). This involves solving the intra-limb constraints (\ref%
{GeomConsLoopCutJoint}), i.e. evaluating $\psi _{\left( l\right) }$ in (\ref%
{solConL}). An iterative numerical solution scheme is proposed to this end.
Its iterative nature necessitates separate correction steps for the
evaluation of $f_{\left( l\right) }$ and $\psi _{\left( l\right) }$,
respectively. This becomes apparent by the following commutative diagram,
which summarizes the evaluation cycle:

\centerline{\includegraphics[height=2.6cm]{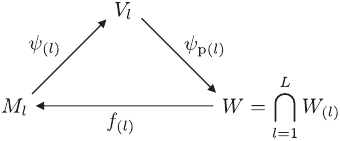}}

The procedure is summarized as follows: Starting with a prescribed $\mathbf{C%
}_{\mathrm{p}}\in W$, an approximate solution $\mathbf{\overline{\mathbf{q}}}%
_{\left( l\right) }=f_{\left( l\right) }(\mathbf{C}_{\mathrm{p}})$ is
obtained numerically. The configuration of limb $l$ must then be adapted by
solving the intra-limb constraints, evaluating $\overline{%
\mathbold{\vartheta}%
}_{\left( l\right) }=\psi _{\left( l\right) }(\mathbf{\overline{\mathbf{q}}}%
_{\left( l\right) })$. The corresponding platform pose is then obtained by
evaluating $\mathbf{\overline{\mathbf{C}}}_{\mathrm{p}}=\psi _{\mathrm{p}%
\left( l\right) }(\overline{%
\mathbold{\vartheta}%
}_{\left( l\right) })$. The discrepancy of $\mathbf{\mathbf{C}}_{\mathrm{p}}$
and $\mathbf{\overline{\mathbf{C}}}_{\mathrm{p}}$ is then used to update $%
\mathbf{q}_{\left( l\right) }$ again in the next iteration cycle.

\subsubsection{Inverse Kinematics of Limb ---Outer-Loop}

The evaluation cycle gives rise to a nested iterative Newton-Raphson scheme.
For prescribed $\mathbf{C}_{\mathrm{p}}\in W$, the outer-loop computes $%
\mathbf{q}_{\left( l\right) }=f_{\left( l\right) }(\mathbf{C}_{\mathrm{p}})$
so that $\mathbf{C}_{\mathrm{p}}=\varphi _{\mathrm{p}\left( l\right) }(%
\mathbf{q}_{\left( l\right) })$, while the inner-loop evaluates $%
\mathbold{\vartheta}%
_{\left( l\right) }=\psi _{\left( l\right) }(\mathbf{q}_{\left( l\right) })$
by solving the intra-limb constraints, which is used in the outer-loop to
evaluate $\varphi _{\mathrm{p}\left( l\right) }=\psi _{\left( l\right)
}\circ \psi _{\mathrm{p}\left( l\right) }(\mathbf{q}_{\left( l\right) })$.

Consider the PKM in an initial state. Let $%
\mathbold{\vartheta}%
_{\left( l\right) ,0}\in V_{l}$ be an admissible initial configuration of
the PKM, and $\mathbf{C}_{\mathrm{p},0}$ be the corresponding platform pose.
Given a desired platform pose $\mathbf{C}_{\mathrm{p}}\in W$, the task is to
compute the corresponding configuration $%
\mathbold{\vartheta}%
_{\left( l\right) }=%
\mathbold{\vartheta}%
_{\left( l\right) ,0}+\Delta 
\mathbold{\vartheta}%
_{\left( l\right) }$ starting from $%
\mathbold{\vartheta}%
_{\left( l\right) ,0}$. Firstly, this necessitates are geometrically
consistent expression for the increment of the platform pose. The
configuration increment, represented in the platform frame $\mathcal{F}_{%
\mathrm{p}}$, is $\Delta \mathbf{C}_{\mathrm{p}}=\mathbf{C}_{\mathrm{p}%
,0}^{-1}\mathbf{C}_{\mathrm{p}}\in SE\left( 3\right) $. Its first-order
approximation yields $\Delta \mathbf{C}_{\mathrm{p}}\approx \mathbf{I}-%
\widehat{\mathbf{L}_{\mathrm{p}\left( l\right) }\Delta \mathbf{q}_{\left(
l\right) }}\in se\left( 3\right) $, and thus in vector form $\left( \mathbf{I%
}-\Delta \mathbf{C}_{\mathrm{p}}\right) ^{\vee }=\mathbf{L}_{\mathrm{p}%
\left( l\right) }\Delta \mathbf{q}_{\left( l\right) }$\footnote{$\mathbf{A}%
^{\vee }\in {\mathbb{R}}^{6}$ is the vector corresponding to matrix $\mathbf{%
A}\in se\left( 3\right) $. The inverse operation $\widehat{\mathbf{X}}\in
se\left( 3\right) $ assigns a $se\left( 3\right) $-matrix to vector $\mathbf{%
X}\in {\mathbb{R}}^{6}$. See appendix \ref{secSymbols}.}. The $\delta _{%
\mathrm{p}\left( l\right) }$ components of this configuration increment
corresponding to the platform motion are extracted with a $\delta _{\mathrm{p%
}\left( l\right) }\times 6$ matrix $\mathbf{P}_{\mathrm{t}\left( l\right) }$%
. Relation (\ref{VpD}) then leads to $\mathbf{D}_{\mathrm{t}\left( l\right) }%
\mathbf{P}_{\mathrm{t}\left( l\right) }\left( \mathbf{I}-\Delta \mathbf{C}_{%
\mathrm{p}}\right) ^{\vee }=\mathbf{L}_{\mathrm{t}\left( l\right) }\Delta 
\mathbf{q}_{\left( l\right) }$. The increment of generalized coordinates is
thus approximated as%
\begin{equation}
\Delta \mathbf{q}_{\left( l\right) }=\mathbf{F}_{\mathrm{t}\left( l\right) }%
\mathbf{P}_{\mathrm{t}\left( l\right) }\left( \mathbf{I}-\Delta \mathbf{C}_{%
\mathrm{p}}\right) ^{\vee }=\mathbf{F}_{\mathrm{t}\left( l\right) }\Delta 
\mathbf{x}_{\mathrm{p}},\ \ \mathrm{with}\ \ \Delta \mathbf{x}_{\mathrm{p}}:=%
\mathbf{P}_{\mathrm{t}\left( l\right) }\left( \mathbf{I}-\Delta \mathbf{C}_{%
\mathrm{p}}\right) ^{\vee }
\end{equation}%
and the generalized coordinates are approximated as $\mathbf{q}_{\left(
l\right) }=\mathbf{q}_{\left( l\right) ,0}+\Delta \mathbf{q}_{\left(
l\right) }$.

In case of the IRSBot-2, the platform at a separated limb has $\delta _{%
\mathrm{p}\left( l\right) }=3$ DOF, and its spatial translation is used as
independent motion component. The corresponding $3\times 6$ selection matrix
is%
\begin{equation}
\mathbf{P}_{\mathrm{t}\left( l\right) }=\left( 
\begin{array}{cccccc}
0 & 0 & 0 & 1 & 0 & 0 \\ 
0 & 0 & 0 & 0 & 1 & 0 \\ 
0 & 0 & 0 & 0 & 0 & 1%
\end{array}%
\right) .
\end{equation}

Invoking (\ref{Hl}), the increment of tree-joint variables is approximated
as $\Delta 
\mathbold{\vartheta}%
_{\left( l\right) }=\mathbf{H}_{\left( l\right) }\Delta \mathbf{q}_{\left(
l\right) }$. This step is applied iteratively until $\mathbf{C}_{\mathrm{p}%
}=\varphi _{\mathrm{p}\left( l\right) }(\mathbf{q}_{\left( l\right) })$ is
satisfied. Since $\mathbf{F}_{\mathrm{t}\left( l\right) }(%
\mathbold{\vartheta}%
_{\left( l\right) })$ depends on the tree-joint variables, and $\varphi _{%
\mathrm{p}\left( l\right) }$ rests on a constraint solution, iterative
application of this step must involve a subsequent correction of $%
\mathbold{\vartheta}%
_{\left( l\right) }$ in order to satisfy the intra-limb constraints. To this
end, a nested iterative solution scheme is introduced as Algorithm 1.

\begin{tabular}[t]{l}
\\ 
\textbf{Algorithm 1: Iterative Solution of Inverse Kinematics of Limb
---Outer-Loop}%
\vspace{1ex}
\\ 
\begin{tabular}[t]{l}
\hspace{-2ex}%
\begin{tabular}[t]{ll}
\underline{Input:} & 
\hspace{-2ex}%
$%
\mathbold{\vartheta}%
_{\left( l\right) ,0},\mathbf{C}_{\mathrm{p}}$%
\vspace{0.7ex}
\\ 
\underline{Initialization:} & 
\hspace{-2ex}%
$\mathbf{q}_{\left( l\right) }:=\mathbf{q}_{\left( l\right) ,0},%
\mathbold{\vartheta}%
_{\left( l\right) }:=%
\mathbold{\vartheta}%
_{\left( l\right) ,0},\overline{\mathbf{C}}_{\mathrm{p}}:=\psi _{\mathrm{p}%
\left( l\right) }(%
\mathbold{\vartheta}%
_{\left( l\right) ,0}),\Delta \mathbf{C}_{\mathrm{p}}:=\overline{\mathbf{C}}%
_{\mathrm{p}}^{-1}\mathbf{C}_{\mathrm{p}},\Delta \mathbf{x}_{\mathrm{t}}:=%
\mathbf{P}_{\mathrm{t}\left( l\right) }\left( \mathbf{I}-\Delta \mathbf{C}_{%
\mathrm{p}}\right) ^{\vee }$%
\end{tabular}%
\vspace{1ex}
\\ 
\hspace{-2ex}%
\begin{tabular}{l}
\hspace{-1.5ex}%
\begin{tabular}[t]{ll}
\underline{Outer-Loop:} & WHILE $\left\Vert \Delta \mathbf{x}_{\mathrm{t}%
}\right\Vert >\varepsilon $ DO \\ 
& \ 
\begin{tabular}{rrll}
& $\Delta \mathbf{q}_{\left( l\right) }$ & 
\hspace{-2ex}%
$:=$ & 
\hspace{-2ex}%
$\mathbf{F}_{\mathrm{t}\left( l\right) }\Delta \mathbf{x}_{\mathrm{t}}$%
\vspace{0.7ex}
\\ 
& $\mathbf{q}_{\left( l\right) }$ & 
\hspace{-2ex}%
$:=$ & 
\hspace{-2ex}%
$\mathbf{q}_{\left( l\right) }+\Delta \mathbf{q}_{\left( l\right) }$%
\vspace{0.7ex}
\\ 
& $%
\mathbold{\vartheta}%
_{\left( l\right) }$ & 
\hspace{-2ex}%
$:=$ & 
\hspace{-2ex}%
$\psi _{\left( l\right) }(\mathbf{q}_{\left( l\right) })$ \ \ \ Algorithm 2
(Inner-loop, evaluated with $%
\mathbold{\vartheta}%
_{\left( l\right) }$)%
\vspace{0.7ex}
\\ 
& $\overline{\mathbf{C}}_{\mathrm{p}}$ & 
\hspace{-2ex}%
$:=$ & 
\hspace{-2ex}%
$\psi _{\mathrm{p}\left( l\right) }(%
\mathbold{\vartheta}%
_{\left( l\right) })$ \ (evaluate eq. (\ref{Cp}))%
\vspace{0.4ex}
\\ 
& $\Delta \mathbf{C}_{\mathrm{p}}$ & 
\hspace{-2ex}%
$:=$ & 
\hspace{-2ex}%
$\overline{\mathbf{C}}_{\mathrm{p}}^{-1}\mathbf{C}_{\mathrm{p}}$%
\vspace{0.7ex}
\\ 
& $\Delta \mathbf{x}_{\mathrm{t}}$ & 
\hspace{-2ex}%
$:=$ & 
\hspace{-2ex}%
$\mathbf{P}_{\mathrm{t}\left( l\right) }\left( \mathbf{I}-\Delta \mathbf{C}_{%
\mathrm{p}}\right) ^{\vee }$%
\end{tabular}
\\ 
& END%
\end{tabular}%
\vspace{0.3ex}%
\end{tabular}
\\ 
\hspace{-2ex}%
\begin{tabular}[t]{ll}
\underline{Output:} & $%
\mathbold{\vartheta}%
_{\left( l\right) }$%
\end{tabular}%
\end{tabular}
\\ 
\ 
\end{tabular}

Notice that there is no bi-invariant metric on $SE\left( 3\right) $. For
computing $\left\Vert \Delta \mathbf{x}_{\mathrm{t}}\right\Vert $ the
left-invariant metric is used, which depends on the scaling \cite%
{Murray,SeligBook2005}. The corresponding norm of $\mathbf{X}=\left( 
\mathbold{\xi}%
,%
\mathbold{\eta}%
\right) ^{T}\in {\mathbb{R}}^{6}\cong se\left( 3\right) $ is $\left\Vert 
\mathbf{X}\right\Vert =\alpha \left\Vert 
\mathbold{\xi}%
\right\Vert +\beta \left\Vert 
\mathbold{\eta}%
\right\Vert $.

\subsubsection{Iterative Solution of Intra-Limb Constraints ---Inner-Loop%
\label{secSolLoop}%
}

As the solution is generally not unique, this solution $%
\mathbold{\vartheta}%
_{\left( \lambda ,l\right) }=\psi _{\left( \lambda ,l\right) }(\mathbf{q}%
_{\left( \lambda ,l\right) })$ is valid in a neighborhood of the starting
configuration. The constraints are solved with an iterative Newton-Raphson
scheme. The solution of constraints (\ref{GeomConsLoopCutJoint}) is
evaluated numerically for given $\mathbf{q}_{\left( \lambda ,l\right) }$
starting from an admissible initial configuration $%
\mathbold{\vartheta}%
_{\left( \lambda ,l\right) ,0}=(\mathbf{y}_{\left( \lambda ,l\right) ,0},%
\mathbf{q}_{\left( \lambda ,l\right) ,0})\in V_{l}$. The prescribed
increment of generalized coordinates is $\Delta \mathbf{q}_{\left( \lambda
,l\right) }=\mathbf{q}_{\left( \lambda ,l\right) }-\mathbf{q}_{\left(
\lambda ,l\right) ,0}$. The solution for the dependent coordinates is
expressed as $\mathbf{y}_{\left( \lambda ,l\right) }=\mathbf{y}_{\left(
\lambda ,l\right) ,0}+\Delta \mathbf{y}_{\left( \lambda ,l\right) }$. A
Newton-Raphson iteration scheme is constructed by linearizing the geometric
constraints at the current configuration assuming sufficiently small $\Delta 
\mathbf{y}_{\left( \lambda ,l\right) }$. First-order Taylor expansion of (%
\ref{GeomConsLoopCutJoint}) yields $g_{\left( \lambda ,l\right) }(%
\mathbold{\vartheta}%
_{\left( \lambda ,l\right) ,0})+\mathbf{G}_{\mathbf{y}\left( \lambda
,l\right) }(%
\mathbold{\vartheta}%
_{\left( \lambda ,l\right) ,0})\Delta \mathbf{y}_{\left( \lambda ,l\right) }+%
\mathbf{G}_{\mathbf{q}\left( \lambda ,l\right) }(%
\mathbold{\vartheta}%
_{\left( \lambda ,l\right) ,0})\Delta \mathbf{q}_{\left( \lambda ,l\right) }=%
\mathbf{0}$. In the following, the index $\left( \lambda ,l\right) $ is
omitted for sake of simplicity, as the derivation always refers to FC $%
\lambda $ of limb $l$. The solution of the overall system of linearized
constraints is%
\begin{equation}
\left( 
\begin{array}{c}
\Delta \mathbf{y} \\ 
\Delta \mathbf{q}%
\end{array}%
\right) =\left( 
\begin{array}{cc}
-\mathbf{G}_{\mathbf{y}}^{-1}(%
\mathbold{\vartheta}%
_{0}) & -\mathbf{G}_{\mathbf{y}}^{-1}(%
\mathbold{\vartheta}%
_{0})\mathbf{G}_{\mathbf{q}}(%
\mathbold{\vartheta}%
_{0}) \\ 
\mathbf{0} & \mathbf{I}%
\end{array}%
\right) \left( 
\begin{array}{c}
g(%
\mathbold{\vartheta}%
_{0}) \\ 
\mathbf{q}-\mathbf{q}_{0}%
\end{array}%
\right) .
\end{equation}%
The so obtained coordinates $%
\mathbold{\vartheta}%
_{0}+\Delta 
\mathbold{\vartheta}%
$ will not satisfy the constraints. Starting from an admissible
configuration $%
\mathbold{\vartheta}%
_{0}$, i.e. $g(%
\mathbold{\vartheta}%
_{0})=\mathbf{0}$, this gives rise to the following Newton-Raphson iteration
scheme in Algorithm 2.

\begin{tabular}[t]{l}
\\ 
\textbf{Algorithm 2: Iterative Solution of Loop Constraints ---Inner-Loop}%
\vspace{1ex}
\\ 
\begin{tabular}[t]{ll}
\underline{Input:} & $%
\mathbold{\vartheta}%
_{0}=(\mathbf{y}_{0},\mathbf{q}_{0}),\ \Delta \mathbf{q}$%
\vspace{2ex}
\\ 
\hspace{-1.3ex}%
\begin{tabular}{r}
\underline{Initial Step:} \\ 
\  \\ 
\ \ 
\end{tabular}
& 
\begin{tabular}{rll}
$\mathbf{q}$ & 
\hspace{-2ex}%
$:=$ & 
\hspace{-2ex}%
$\mathbf{q}_{0}+\Delta \mathbf{q}$ \\ 
$\Delta \mathbf{y}$ & 
\hspace{-2ex}%
$:=$ & 
\hspace{-2ex}%
$-\mathbf{G}_{\mathbf{y}}^{-1}(%
\mathbold{\vartheta}%
_{0})\mathbf{G}_{\mathbf{q}}(%
\mathbold{\vartheta}%
_{0})\Delta \mathbf{q}$ \\ 
$%
\mathbold{\vartheta}%
$ & 
\hspace{-2ex}%
$:=$ & 
\hspace{-2ex}%
$(\mathbf{y}_{0}+\Delta \mathbf{y},\mathbf{q})$%
\end{tabular}%
\vspace{2ex}
\\ 
\underline{Correction Steps:} & WHILE $\left\Vert g\left( 
\mathbold{\vartheta}%
\right) \right\Vert >\varepsilon $ DO \\ 
& \ \ 
\begin{tabular}{rll}
$\Delta \mathbf{y}$ & 
\hspace{-2ex}%
$:=$ & 
\hspace{-2ex}%
$-\mathbf{G}_{\mathbf{y}}^{-1}(%
\mathbold{\vartheta}%
)\mathbf{g}\left( 
\mathbold{\vartheta}%
\right) $ \\ 
$%
\mathbold{\vartheta}%
$ & 
\hspace{-2ex}%
$:=$ & 
\hspace{-2ex}%
$(\mathbf{y}+\Delta \mathbf{y},\mathbf{q})$%
\end{tabular}
\\ 
& END%
\vspace{2ex}
\\ 
\underline{Output:} & $%
\mathbold{\vartheta}%
=(\mathbf{y},\mathbf{q})$%
\end{tabular}
\\ 
\ 
\end{tabular}

This evaluates the solution $%
\mathbold{\vartheta}%
_{\left( \lambda ,l\right) }=\psi _{\left( \lambda ,l\right) }(\mathbf{q}%
_{\left( \lambda ,l\right) })$ of loop constraints of FC $\Lambda _{\left(
\lambda ,l\right) }$, and thus the overall solution (\ref{solConL}) in terms
of the $\delta _{l}$ generalized coordinates $\mathbf{q}_{\left( l\right) }$%
, as they are independent for a hybrid limb. The iteration scheme is a
computationally efficient and numerically stable solution method, which
converges after a few iteration steps, provided that $\mathbf{G}_{\mathbf{y}%
\left( \lambda ,l\right) }$ or $\mathbf{G}_{\mathbf{q}\left( \lambda
,l\right) }$ are well conditioned (which is not the case near singularities).

\begin{remark}
A solution can also be obtained by numerical time integration of the
velocity constraints. This is pursued for numerical time integration of the
EOM of general MBS, e.g. in index 2 formulation. Then, constraint violations
due to numerical drift must be stabilized, however \cite%
{Blajer1994,Terze2010,Blajer2011}. A simple method is the Baumgarte
stabilization, which ensures asymptotic decay of the error over time. When
the model is used to compute the feedforward command in model-based control
schemes, an asymptotic decay is not sufficient.
\end{remark}

\subsection{Simplified iteration scheme ---Compound iteration step for
inverse kinematics and loop constraints%
\label{secOneStep}%
}

The nested solution scheme separates the solution of loop constraints from
that of the inverse kinematics of the limbs. Algorithm 2 performs exactly as
many steps as necessary for the individual constraints. This allows dealing
with ill-conditioned or redundant loop constraints for particular loops,
while not increasing the number of iterations for solving the limb inverse
kinematics. On the other hand, if all subproblems are well-conditioned, the
loop constraints and inverse kinematics can be solved using a one overall
iteration step. In other words, one single step in algorithm 2, for solving
loop constraints, is performed per step of algorithm 1. This is summarized
as Algorithm 3.

\begin{tabular}[t]{l}
\\ 
\textbf{Algorithm 3: Compound Scheme solving Inverse Kinematics of Limb
along with Loop Constraints}%
\vspace{1ex}
\\ 
\begin{tabular}[t]{l}
\hspace{-2ex}%
\begin{tabular}[t]{ll}
\underline{Input:} & 
\hspace{-2ex}%
$%
\mathbold{\vartheta}%
_{\left( l\right) ,0},\ \mathbf{C}_{\mathrm{p}}$%
\vspace{0.7ex}
\\ 
\underline{Initialization:} & 
\hspace{-2ex}%
$%
\mathbold{\vartheta}%
_{\left( l\right) }:=%
\mathbold{\vartheta}%
_{\left( l\right) ,0},\ \overline{\mathbf{C}}_{\mathrm{p}}:=\psi _{\mathrm{p}%
\left( l\right) }(%
\mathbold{\vartheta}%
_{\left( l\right) ,0}),\ \Delta \mathbf{C}_{\mathrm{p}}:=\overline{\mathbf{C}%
}_{\mathrm{p}}^{-1}\mathbf{C}_{\mathrm{p}},\ \Delta \mathbf{x}_{\mathrm{t}}:=%
\mathbf{P}_{\mathrm{t}\left( l\right) }\left( \mathbf{I}-\Delta \mathbf{C}_{%
\mathrm{p}}\right) ^{\vee }$%
\end{tabular}%
\vspace{1ex}
\\ 
\hspace{-2ex}%
\begin{tabular}{l}
\hspace{-1.5ex}%
\begin{tabular}[t]{ll}
\underline{Outer-Loop:} & WHILE $\left\Vert \Delta \mathbf{x}_{\mathrm{t}%
}\right\Vert >\varepsilon _{1}\vee \left\Vert g(%
\mathbold{\vartheta}%
_{\left( l\right) })\right\Vert >\varepsilon _{2}$ DO \\ 
& \ 
\begin{tabular}{rrll}
& $\Delta 
\mathbold{\vartheta}%
_{\left( l\right) }$ & 
\hspace{-2ex}%
$:=$ & 
\hspace{-2ex}%
$\mathbf{H}_{\left( l\right) }(%
\mathbold{\vartheta}%
_{\left( l\right) })\mathbf{F}_{\mathrm{t}\left( l\right) }(%
\mathbold{\vartheta}%
_{\left( l\right) })\mathbf{P}_{\mathrm{t}}\left( \mathbf{I}-\Delta \mathbf{C%
}_{\mathrm{p}}\right) ^{\vee }$%
\vspace{0.7ex}
\\ 
& $%
\mathbold{\vartheta}%
_{\left( l\right) }$ & 
\hspace{-2ex}%
$:=$ & 
\hspace{-2ex}%
$%
\mathbold{\vartheta}%
_{\left( l\right) }+\Delta 
\mathbold{\vartheta}%
_{\left( l\right) }$%
\vspace{0.7ex}
\\ 
& $\overline{\mathbf{C}}_{\mathrm{p}}$ & 
\hspace{-2ex}%
$:=$ & 
\hspace{-2ex}%
$\psi _{\mathrm{p}\left( l\right) }(%
\mathbold{\vartheta}%
_{\left( l\right) })$ \ (evaluate eq. (\ref{Cp}))%
\vspace{0.7ex}
\\ 
& $\Delta \mathbf{C}_{\mathrm{p}}$ & 
\hspace{-2ex}%
$:=$ & 
\hspace{-2ex}%
$\overline{\mathbf{C}}_{\mathrm{p}}^{-1}\mathbf{C}_{\mathrm{p}}$ \\ 
& $\Delta \mathbf{x}_{\mathrm{t}}$ & 
\hspace{-2ex}%
$:=$ & 
\hspace{-2ex}%
$\mathbf{P}_{\mathrm{t}\left( l\right) }\left( \mathbf{I}-\Delta \mathbf{C}_{%
\mathrm{p}}\right) ^{\vee }$%
\end{tabular}
\\ 
& END%
\end{tabular}%
\vspace{0.3ex}%
\end{tabular}
\\ 
\hspace{-2ex}%
\begin{tabular}[t]{ll}
\underline{Output:} & $%
\mathbold{\vartheta}%
_{\left( l\right) }$%
\end{tabular}%
\end{tabular}
\\ 
\ 
\end{tabular}

\subsection{Comparison with standard MBS formulations%
\label{secCompare}%
}

Many of the proposed lower-mobility PKM with complex limbs \cite%
{BriotKhalilBook,KongGosselin2002,HuangLiDing2013} are (globally)
overconstrained, which is reflected by a redundant overall system of
constraints. This is problematic when using standard MBS modeling
approaches. The local constraint embedding alleviates this problem as for
many (globally) overconstrained PKM, the intra-limb constraints are not
redundant. This is briefly discussed next.

\subparagraph{Relative coordinate formulation}

\begin{figure}[h]
\centerline{\includegraphics[height=5.6cm]{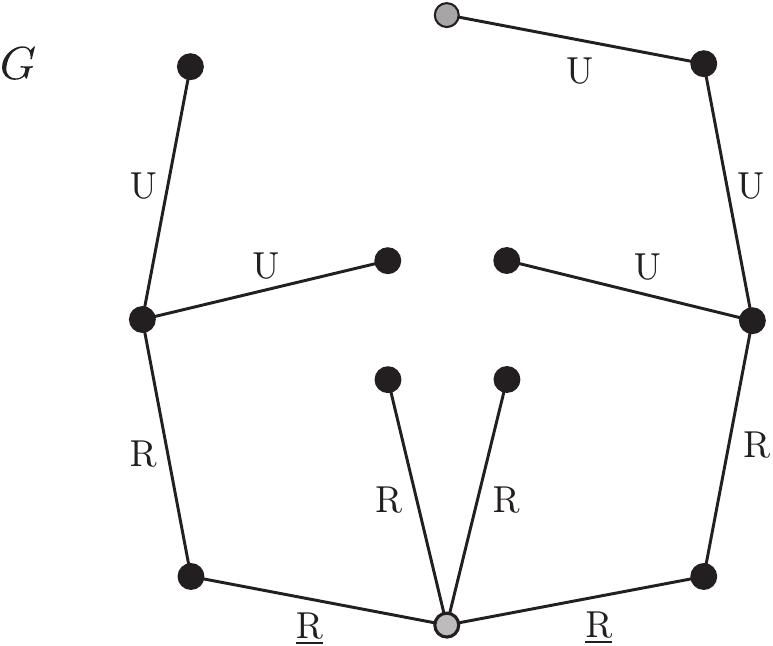}}
\caption{A possible tree $G$ on the topological graph $\Gamma $ of the
IRSBot-2 in fig. \protect\ref{figIRSBotGraph}a).}
\label{figTreeMBS}
\end{figure}

In the relative coordinate formulation, a spanning tree is introduced for
the PKM as a whole. For example, a possible tree on the topological graph $%
\Gamma $ in fig. \ref{figIRSBotGraph}a) is shown in fig. \ref{figTreeMBS}.
Three U joints and two R joints are removed, which leaves a tree-topology
MBS with $n=5\cdot 2+3\cdot 1=13$ tree-joint variables (5 U- \& 3 R-joints).
A system of $m_{\mathrm{rel}}=3\cdot 4+2\cdot 5=22$ cut-joint constraints (3
U- \& 2 R-joints) is introduced. Clearly, 7 of the 22 constraints are
redundant as the PKM DOF is $\delta =2$. If the information that the
R-joints form planar mechanisms, and only 2 constraints are imposed per
loop, then there are $16$ cut-joint constraints and still 5 of them are
redundant. 

\subparagraph{Absolute coordinate formulation}

The absolute coordinate formulation is another established approach in
computational MBS dynamics \cite{ShabanaBook}. It leads to a large but
sparse system of EOM where the constraint Jacobian is sparse reflecting the
local topological loops and the overall PKM topology. The rank deficiency is
the same as in the relative coordinate formulation. The absolute coordinate
model explicitly involves the constraint reactions (Lagrange multipliers).
This information is useful for trajectory planning and can be used for
limiting the joint loads \cite{ICRA2022}, while for control purposes, the
minimal coordinate model is preferable.

\subparagraph{Local constraint embedding}

The $L$ inter-limb constraints give rise to the overall system of $%
m=m_{1}+\ldots +m_{l}$ constraints. As with the relative coordinate model,
these constraints are redundant for overconstrained (lower-mobility) PKM.
Instead of imposing a system of $m$ constraints, the local constraint
embedding technique imposes the particular intra-limb constraints to the
separated kinematic loops, and incorporates their solutions into the model.
It thus avoids dealing with a singular overall constraints system. Even if
the intra-limb constraints are redundant, reflected by singular Jacobians $%
\mathbf{G}_{\left( \lambda ,l\right) }$ in (\ref{VelConsLoopCutJoint}), the
local constraint embedding method allows dealing with this locally.
Moreover, many lower-mobility PKM are globally overconstrained but the
separated limbs are not overconstrained. The IRSBot-2, for instance, is
overconstrained leading to a redundant overall system of $m_{\mathrm{rel}%
}=22 $ constraints (see above). On the other hand, the intra-limb
constraints for the 4U loop is regular, and those for the planar 4R loop can
be non-redundantly formulated (see section \ref{secIRSBot}). That is, the
overconstrained problem is avoided by the local constraint embedding.
Moreover, the embedding technique renders the PKM model as a model of PKM
with simple limbs. This allows for modeling in terms of task space
coordinates, by combining the models of the $L$ limbs, and thus treating
complex \cite{MMT2022} and simple limbs \cite{MuellerAMR2020} in a unified
way.

\section{Dynamics Equations of Motion%
\label{secEOMPKM}%
}

\begin{figure}[h]
\centerline{%
\includegraphics[height=5.6cm]{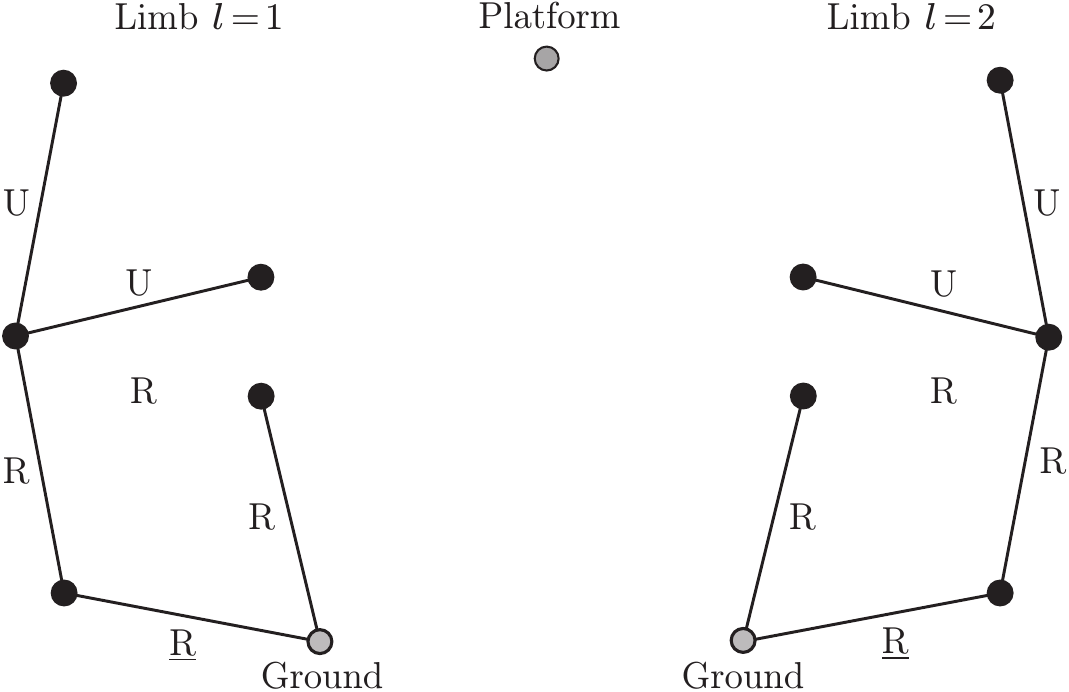}}
\caption{Kinematic topology of the system for which EOM are derived in case
of the IRSBot-2.}
\label{figLimbsPlatform}
\end{figure}

A task space formulation of the dynamic EOM of PKM with complex limbs was
derived in \cite{MMT2022}, and for PKM with simple limbs in \cite%
{AbdellatifHeimann_MMT2009,BriotKhalilBook,MuellerAMR2020} by combining the
EOM of the $L$ limbs using the inverse kinematics solution (\ref{Ftheta2})
in terms of the generalized coordinates $\mathbf{q}_{\left( l\right) }$. In
the following, the inverse kinematics solution computed with the constraint
embedding technique is used. In this formulation, the platform is treated
separately, and the EOM of limb $l$ in terms of $\mathbf{q}_{\left( l\right)
}$ are constructed from the EOM of the tree-topology system using the
solution (\ref{Hl}) of the intra-limb constraints. For the IRSBot-2 example,
Fig. \ref{figLimbsPlatform} shows the kinematic topology of the system for
which the EOM are constructed that are then combined to the overall EOM,
which comprises two tree-systems for the limbs and the single platform body. 

\subsection{EOM of Tree-Topology System of a Limb}

The platform and the joint connecting it to limb $l$ are removed from the
tree-topology system associated to the limb. Denote with $\bar{%
\mathbold{\vartheta}%
}_{\left( l\right) }\in {\mathbb{V}}^{\bar{n}_{l}}$ the vector of the
remaining $\bar{n}_{l}$ tree-joint coordinates. The EOM of this system can
be written in the standard form for tree-topology MBS%
\begin{equation}
\bar{\mathbf{M}}_{\left( l\right) }\ddot{\bar{%
\mathbold{\vartheta}%
}}_{\left( l\right) }+\bar{\mathbf{C}}_{\left( l\right) }\dot{\bar{%
\mathbold{\vartheta}%
}}_{\left( l\right) }+\bar{\mathbf{Q}}_{\left( l\right) }^{\mathrm{grav}}+%
\bar{\mathbf{Q}}_{\left( l\right) }=\bar{\mathbf{Q}}_{\left( l\right) }^{%
\mathrm{act}}  \label{EOMLimb}
\end{equation}%
Therein, $\bar{\mathbf{M}}_{\left( l\right) }(\bar{%
\mathbold{\vartheta}%
}_{\left( l\right) })$ is the generalized mass matrix, $\bar{\mathbf{C}}%
_{\left( l\right) }(\dot{\bar{%
\mathbold{\vartheta}%
}}_{\left( l\right) },\dot{%
\mathbold{\vartheta}%
}_{\left( l\right) })$ is the Coriolis/centrifugal matrix, $\bar{\mathbf{Q}}%
_{\left( l\right) }^{\mathrm{grav}}(\bar{%
\mathbold{\vartheta}%
}_{\left( l\right) })$ represents generalized gravity forces, and $\bar{%
\mathbf{Q}}_{\left( l\right) }$ accounts for all remaining forces (e.g.
friction). The vector $\bar{\mathbf{Q}}_{\left( l\right) }^{\mathrm{act}%
}\left( t\right) $ accounts for actuation forces. Its non-zero entries are
the torques/forces at the actuated joints. The EOM (\ref{EOMLimb}) can be
expressed in a compact and readily implemented form employing the system
matrices $\mathsf{J}_{\left( l\right) },\mathsf{A}_{\left( l\right) },%
\mathsf{a}_{\left( l\right) }$ used for the kinematics modeling in sec. \ref%
{secLimbKin}. Details can be found in \cite%
{MMT2022,ModernRobotics,MUBOScrew1}.

\subsection{EOM of Complex Limb without Platform}

The joint velocity of the tree-system is determined with (\ref{Hl}) by the
generalized velocities. The joint velocities of the system without platform
can thus be written as 
\begin{equation}
\dot{\bar{%
\mathbold{\vartheta}%
}}_{\left( l\right) }=\bar{\mathbf{H}}_{\left( l\right) }\dot{\bar{\mathbf{q}%
}}_{\left( l\right) }.  \label{etaq2}
\end{equation}%
where $\bar{\mathbf{H}}_{\left( l\right) }(%
\mathbold{\vartheta}%
_{\left( l\right) })$ is the $\bar{n}_{l}\times \delta _{l}$ submatrix of $%
\mathbf{H}_{\left( l\right) }$ with rows corresponding to the $\bar{n}_{l}$
tree-joint variables. Using variations $\dot{\delta \bar{%
\mathbold{\vartheta}%
}}_{\left( l\right) }=\bar{\mathbf{H}}_{\left( l\right) }\delta \dot{\mathbf{%
q}}_{\left( l\right) }$, along with (\ref{etaq2}) and its derivative, the
principle of virtual power yields the EOM of the complex limb

\begin{equation}
\bar{\bar{\mathbf{M}}}_{\left( l\right) }\ddot{\bar{\mathbf{q}}}_{\left(
l\right) }+\bar{\bar{\mathbf{C}}}_{\left( l\right) }\dot{\bar{\mathbf{q}}}%
_{\left( l\right) }+\bar{\bar{\mathbf{Q}}}_{\left( l\right) }^{\mathrm{grav}%
}+\bar{\bar{\mathbf{Q}}}_{\left( l\right) }=\bar{\bar{\mathbf{Q}}}_{\left(
l\right) }^{\mathrm{act}}  \label{EOMLimb2}
\end{equation}%
where%
\begin{align}
\bar{\bar{\mathbf{M}}}_{\left( l\right) }(%
\mathbold{\vartheta}%
_{\left( l\right) })& :=\bar{\mathbf{H}}_{\left( l\right) }^{T}\bar{\mathbf{M%
}}_{\left( l\right) }\bar{\mathbf{H}}_{\left( l\right) }  \notag \\
\bar{\bar{\mathbf{C}}}_{\left( l\right) }(%
\mathbold{\vartheta}%
_{\left( l\right) },\dot{%
\mathbold{\vartheta}%
}_{\left( l\right) })& :=\bar{\mathbf{H}}_{\left( l\right) }^{T}(\bar{%
\mathbf{M}}_{\left( l\right) }\dot{\bar{\mathbf{H}}}_{\left( l\right) }+\bar{%
\mathbf{C}}_{\left( l\right) }\bar{\mathbf{H}}_{\left( l\right) })
\label{MCQLimb} \\
\bar{\bar{\mathbf{Q}}}_{\left( l\right) }(%
\mathbold{\vartheta}%
_{\left( l\right) })& :=\bar{\mathbf{H}}_{\left( l\right) }^{T}\bar{\mathbf{Q%
}}_{\left( l\right) },\ \bar{\bar{\mathbf{Q}}}_{\left( l\right) }^{\mathrm{%
act}}(%
\mathbold{\vartheta}%
_{\left( l\right) }):=\bar{\mathbf{H}}_{\left( l\right) }^{T}\bar{\mathbf{Q}}%
_{\left( l\right) }^{\mathrm{act}}  \notag \\
\bar{\bar{\mathbf{Q}}}_{\left( l\right) }^{\mathrm{grav}}(%
\mathbold{\vartheta}%
)& :=\bar{\mathbf{H}}_{\left( l\right) }^{T}\bar{\mathbf{Q}}_{\left(
l\right) }^{\mathrm{grav}}.  \notag
\end{align}%
The EOM (\ref{EOMLimb2}) constitute a system of $\delta _{l}$ ODEs in terms
of the generalized velocities $\dot{\mathbf{q}}_{\left( l\right) }$ and
acceleration $\ddot{\mathbf{q}}_{\left( l\right) }$. Since the geometric
loop constraints cannot be solved explicitly, all terms remain dependent on
the tree-joint variables $%
\mathbold{\vartheta}%
_{\left( l\right) }$ (the joint variables of the complete tree-system,
including the joints connecting the platform with the limbs). The latter are
determined in terms of the generalized coordinates with the numerical
solution of the form $%
\mathbold{\vartheta}%
_{\left( l\right) }=\psi _{\left( l\right) }(\mathbf{q}_{\left( l\right) })$%
, computed with the solution scheme of section \ref{secSolLoop}.

\subsection{Task Space Formulation of PKM Dynamics}

The platform dynamics is not governed by the above EOM. The dynamics of the
platform as a floating rigid body is governed by the Newton-Euler equations,
which are expressed as%
\begin{equation}
\mathbf{M}_{\mathrm{p}}\dot{\mathbf{V}}_{\mathrm{p}}+\mathbf{G}_{\mathrm{p}}%
\mathbf{M}_{\mathrm{p}}\mathbf{V}_{\mathrm{p}}+\mathbf{W}_{\mathrm{p}}^{%
\mathrm{grav}}=\mathbf{W}_{\mathrm{p}}^{\mathrm{EE}}.  \label{EOMPlat}
\end{equation}%
Here $\mathbf{M}_{\mathrm{p}}$ is the constant mass matrix expressed in the
platform frame $\mathcal{F}_{\mathrm{p}}$, and matrix $\mathbf{G}_{\mathrm{p}%
}\left( \mathbf{V}_{\mathrm{p}}\right) =-\mathbf{ad}_{\mathbf{V}_{\mathrm{p}%
}}^{T}$ accounts for gyroscopic and centrifugal effects \cite{MMT2022}. The
wrenches $\mathbf{W}_{\mathrm{p}}^{\mathrm{grav}}$ and $\mathbf{W}_{\mathrm{p%
}}$ account for gravity and the interaction of the PKM via the EE.

With the EOM of all components in place, the overall EOM in terms of the
task space velocity are obtained with the inverse kinematics solution (\ref%
{Ftheta2}). Accordingly, the principle of virtual power with variations $%
\delta \dot{\bar{\mathbf{q}}}_{\left( l\right) }=\bar{\mathbf{F}}_{\left(
l\right) }\delta \mathbf{V}_{\mathrm{t}}$ yields the EOM%
\begin{equation}
\mathbf{M}_{\mathrm{t}}\dot{\mathbf{V}}_{\mathrm{t}}+\mathbf{C}_{\mathrm{t}}%
\mathbf{V}_{\mathrm{t}}+\mathbf{W}_{\mathrm{t}}^{\mathrm{grav}}+\mathbf{W}_{%
\mathrm{t}}=\mathbf{W}_{\mathrm{t}}^{\mathrm{EE}}+\mathbf{J}_{\mathrm{IK}%
}^{T}\mathbf{u}\left( t\right)  \label{EOMTask}
\end{equation}%
with the $\delta \times \delta $ generalized mass matrix and Coriolis matrix%
\begin{align}
\mathbf{M}_{\mathrm{t}}(%
\mathbold{\vartheta}%
):=& \sum_{l=1}^{L}\bar{\mathbf{F}}_{\left( l\right) }^{T}\bar{\bar{\mathbf{M%
}}}_{\left( l\right) }\bar{\mathbf{F}}_{\left( l\right) }+\mathbf{P}_{%
\mathrm{p}}^{T}\mathbf{M}_{\mathrm{p}}\mathbf{P}_{\mathrm{p}}  \label{Mbar}
\\
\mathbf{C}_{\mathrm{t}}(%
\mathbold{\vartheta}%
,\dot{%
\mathbold{\vartheta}%
}):=& \sum_{l=1}^{L}\bar{\mathbf{F}}_{\left( l\right) }^{T}(\bar{\bar{%
\mathbf{C}}}_{\left( l\right) }\bar{\mathbf{F}}_{\left( l\right) }+\bar{\bar{%
\mathbf{M}}}_{\left( l\right) }\dot{\bar{\mathbf{F}}}_{\left( l\right) })+%
\mathbf{P}_{\mathrm{p}}^{T}\mathbf{G}_{\mathrm{p}}\mathbf{M}_{\mathrm{p}}%
\mathbf{P}_{\mathrm{p}}  \label{Cbar}
\end{align}%
where the gyroscopic matrix in (\ref{EOMPlat}) is evaluated with the task
space velocity, using (\ref{VpVt}), as $\mathbf{G}_{\mathrm{p}}=\mathbf{G}_{%
\mathrm{p}}\left( \mathbf{P}_{\mathrm{p}}\mathbf{V}_{\mathrm{t}}\right) $.
The generalized forces accounting for EE-loads, gravity, and all remaining
forces are%
\begin{align}
\mathbf{W}_{\mathrm{t}}^{\mathrm{EE}}\left( t\right) & :=\mathbf{P}_{\mathrm{%
p}}^{T}\mathbf{W}_{\mathrm{p}}^{\mathrm{EE}}\left( t\right)  \label{WEE} \\
\mathbf{W}_{\mathrm{t}}^{\mathrm{grav}}(%
\mathbold{\vartheta}%
,\mathbf{x})& :=\sum_{l=1}^{L}\bar{\mathbf{F}}_{\left( l\right) }^{T}\bar{%
\bar{\mathbf{Q}}}_{\left( l\right) }^{\mathrm{grav}}+\mathbf{P}_{\mathrm{p}%
}^{T}\mathbf{W}_{\mathrm{p}}^{\mathrm{grav}}  \label{Wgrav} \\
\mathbf{W}_{\mathrm{t}}(%
\mathbold{\vartheta}%
,\dot{%
\mathbold{\vartheta}%
},t)& :=\sum_{l=1}^{L}\bar{\mathbf{F}}_{\left( l\right) }^{T}\bar{\bar{%
\mathbf{Q}}}_{\left( l\right) }.  \label{W}
\end{align}%
The vector $\mathbf{u}\in {\mathbb{R}}^{N_{\mathrm{act}}}$ of $N_{\mathrm{act%
}}\geq \delta $ actuator forces/torques yields the corresponding generalized
forces via the inverse kinematics Jacobian $\mathbf{J}_{\mathrm{IK}}(%
\mathbold{\vartheta}%
)$ in (\ref{IK}). All terms in (\ref{EOMTask}) dependent on the tree-joint
variables $%
\mathbold{\vartheta}%
$ of all limbs (including joint connecting limbs and platform).

\subsection{Implementation and Application Aspects}

The task space formulation is particularly advantageous for model-based
control where the solution to the inverse dynamics problem delivers the
feed-forward control command%
\begin{equation}
\mathbf{u}=\mathbf{J}_{\mathrm{IK}}^{-T}(\mathbf{M}_{\mathrm{t}}\dot{\mathbf{%
V}}_{\mathrm{t}}+\mathbf{C}_{\mathrm{t}}\mathbf{V}_{\mathrm{t}}+\mathbf{W}_{%
\mathrm{t}}^{\mathrm{grav}}+\mathbf{W}_{\mathrm{t}}-\mathbf{W}_{\mathrm{t}}^{%
\mathrm{EE}}).  \label{invDyn}
\end{equation}%
In this regard, its computational efficiency is crucial. The formulation
efficiently separates the kinematics modeling of limbs and PKM from the
dynamics modeling. Consequently, the formalism to evaluate the EOM (\ref%
{EOMLimb}) of the limbs can be chosen as appropriate or according to
computational preferences. This could be a classical MBS relative coordinate
formalism \cite{Rodriguez1987,WittenburgBook,AngelesBook} or a Lie group
method \cite{MMT2022,ModernRobotics,MUBOScrew1}. The latter offer a
user-friendly modeling approach while ensuring computational efficiency. A
Lie group formulation employing the notation used in this paper can be found
in \cite[appendix B]{MMT2022}.

Depending on the complexity of the PKM, the computational efficiency can
further be improved by recursively evaluating the EOM (\ref{EOMLimb}) of the
limbs using $O\left( n\right) $ algorithms \cite{ParkBobrowPloen1995,RAL2020}%
, instead of deriving them in closed form. Moreover, the topological
independence of the limbs allows for evaluating the EOM (\ref{EOMLimb2}) in
parallel. Invoking the constraint embedding method, an $O\left( n\right) $
algorithm was proposed in \cite{Jain2009,JainBook,Jain2012} where for each
FC $\Lambda _{\lambda \left( l\right) }$ an aggregated mass matrix is
introduced, which is obtained by introducing the solution of the
constraints. Then, for a hybrid limb, the limb reduces to a serial chain of
aggregated bodies (each representing a FC) whose dynamics can be evaluated
in $O\left( n\right) $.

The topology of PKM naturally allows for a modular modeling approach.
Moreover, since (\ref{EOMLimb}) are the EOM of a separated limb without the
platform, and since the majority of PKM consists of structurally identical
limbs, the EOM (\ref{EOMLimb}) and (\ref{EOMLimb2}) can be derived (or
evaluated) for a prototypical limb, and incorporated into the overall EOM (%
\ref{EOMTask}). This has been described in detail in \cite%
{MuellerAMR2020,MMT2022}.

Another benefit of formulating the EOM in task space in general is that the
terms in the EOM reveal the dynamic properties of the PKM as apparent in
task space. In particular the mass matrix $\mathbf{M}_{\mathrm{t}}$
characterizes the apparent inertia at the EE. This is also called the
reflected mass, which is an information used to assess a PKM when used as
collaborative robot and in human machine interaction (HMI) scenarios \cite%
{Abdallah2022}.

\subsection{Constraint Embedding, Aggregated bodies, and $O\left( n\right) $
Formulations}

Since the constraint embedding method casts the complex limbs of a PKM as
serial kinematic chain obtained by aggregating all bodies of a FC. The FC
are represented by a submechanism with its own internal dynamics, described
by a set of independent generalized coordinates. Each of this submechanisms
is referred to as an aggregated body \cite{JainBook}. That is, the hybrid
limb can be regarded as an aggregated tree where each aggregated body, i.e.
each loop of the hybrid limb, is represented by a (configuration dependent)
mass matrix. The latter are $\mathbf{H}_{\left( \lambda ,l\right) }^{T}%
\mathbf{M}_{\left( \lambda ,l\right) }\mathbf{H}_{\left( \lambda ,l\right) }$%
, where $\mathbf{M}_{\left( \lambda ,l\right) }$ is the submatrix of $\bar{%
\mathbf{M}}_{\left( l\right) }$ associated to the FC $\Lambda _{\lambda
\left( l\right) }$.

\section{Example: IRSBot-2%
\label{secIRSBot}%
}

In the following, numerical results are presented for the IRSBot-2 when the
inverse kinematics is solved with the constraint embedding method, and the
inverse dynamics is solved subsequently. The dynamic parameters are
simplified for simplicity, and since no particular data is available.

\begin{figure}[h]
\caption{Geometry of the IRSBot-Model -- Front view of right limb. $\protect%
\varphi _{0}$ and $\protect\psi _{0}$ describe the orientation of the
proximal and distal limb, and $h_{0}$ is the platform hight in the reference
configuration $%
\mathbold{\vartheta}%
=\mathbf{0}$, respectively.}
\label{figGeom1}\centerline{%
\includegraphics[height=9.1cm]{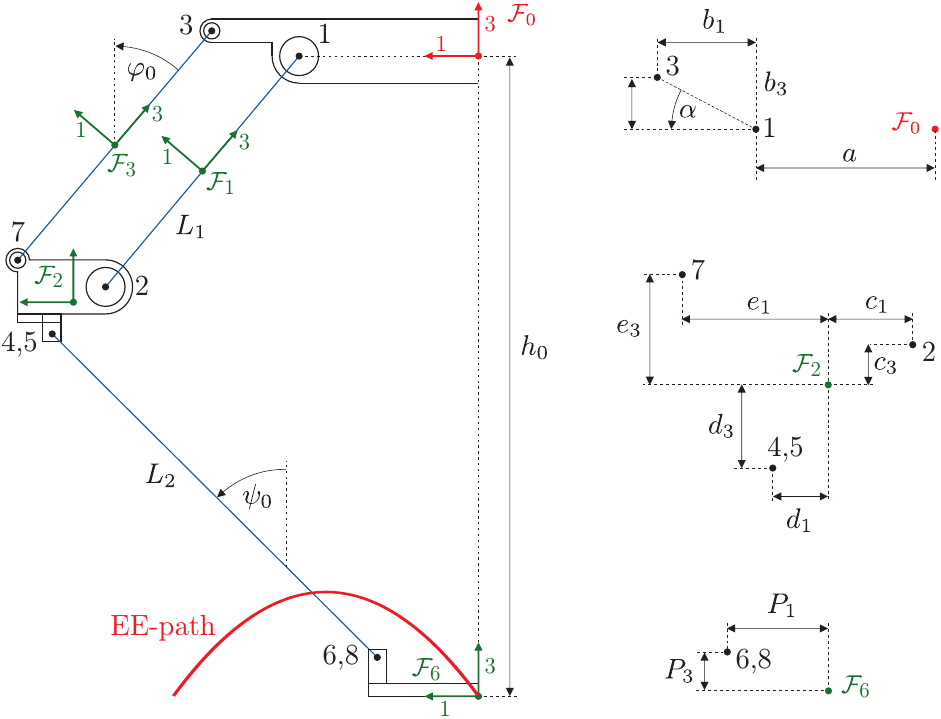}}
\end{figure}
\begin{figure}[h]
\centerline{\includegraphics[height=9.1cm]{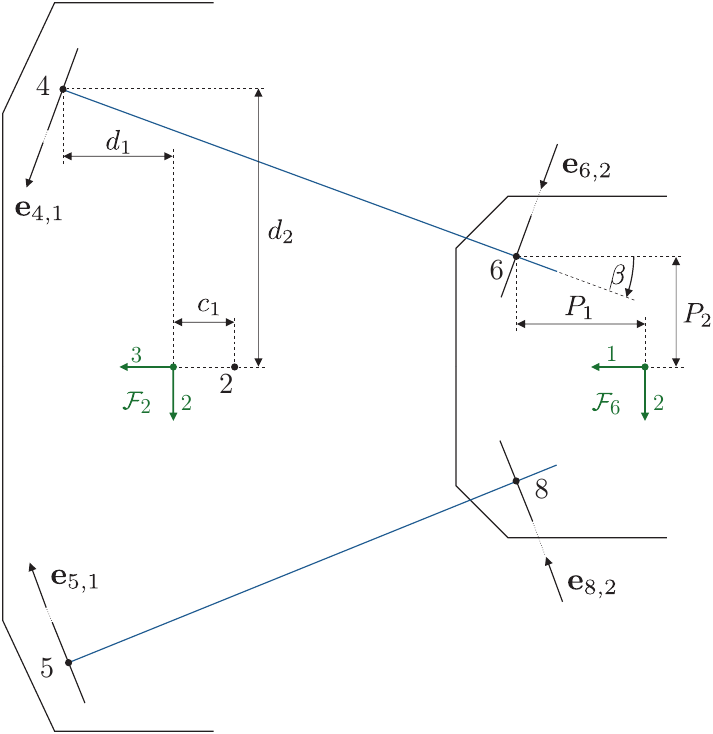}}
\caption{Geometry of the distal limb of the IRSBot-Model -- Top view of the
right limb. $\mathbf{e}_{4,1}$ and $\mathbf{e}_{5,1}$ are unit vectors along
the first revolute axis of U-joint 4 and 5, and $\mathbf{e}_{6,2}$ and $%
\mathbf{e}_{8,2}$ are unit vectors along the second revolute axis of U-joint
6 and 8 respectively. }
\label{figGeom2}
\end{figure}

\subsection{Kinematics of tree-topology system}

\textbf{Joint variables:} The tree-topology system for limb $l=1,2$ is
introduced according to the spanning tree $\vec{G}_{\left( l\right) }$ in
Fig. \ref{figIRSBotGraph}c). It consists of the three R-joints 1,2,3, and
the three U-joints 4,5,6. The revolute joint angles are denoted with $%
\varphi _{1\left( l\right) },\varphi _{2\left( l\right) },\varphi _{3\left(
l\right) }$. The U-joints are modeled as combination of two R-joints with
revolute angles $\varphi _{4,1\left( l\right) },\varphi _{4,2\left( l\right)
},\varphi _{5,1\left( l\right) },\varphi _{5,2\left( l\right) },\varphi
_{6,1\left( l\right) },\varphi _{6,2\left( l\right) }$. The nine revolute
angles form the joint coordinate vector $%
\mathbold{\vartheta}%
_{\left( l\right) }\in {\mathbb{V}}^{9}$. All joint angles are measured
according to the spanning tree.

\textbf{Geometric parameters:} The limb geometry is parameterized as shown
in Fig. \ref{figGeom1} and \ref{figGeom2}. The parameters satisfy the
following relations%
\begin{eqnarray}
b_{1} &=&b\cos \alpha ,\ b_{3}=b\sin \alpha ,\ e_{1}=b_{1}-c_{1},\
e_{3}=b_{3}+c_{3}  \notag \\
\sin \beta &=&\frac{d_{2}-p_{2}}{\sqrt{u^{2}+\left( d_{2}-p_{2}\right) ^{2}}}%
=\frac{1}{\sqrt{u^{2}+1/\tan ^{2}\beta }},\ \cos \beta =\frac{u}{\sqrt{%
u^{2}+\left( d_{2}-p_{2}\right) ^{2}}}=\frac{1}{\sqrt{1+\tan ^{2}\beta }},\
\tan \beta =\frac{d_{2}-p_{2}}{u} \\
\sin \psi _{0} &=&\frac{\sqrt{\left( d_{2}-p_{2}\right) ^{2}+u^{2}}}{L_{2}}%
,\ \cos \psi _{0}=\frac{\sqrt{L_{2}^{2}-\left( d_{2}-p_{2}\right) ^{2}-u^{2}}%
}{L_{2}},\ \ \ \ \mathrm{with}\ \ u:=a+c_{1}+d_{1}-p_{1}+L_{1}\sin \varphi
_{0}.  \notag
\end{eqnarray}%
For numerical computations, the following parameter values are selected so
to approximately resemble the IRSBot-2 design%
\begin{eqnarray}
a &=&1/8\,\mathrm{m},\ b=1/12\,\mathrm{m},\ c_{1}=b_{1}/2,\ c_{3}=b_{3},\
d_{1}=b_{1}/2,\ d_{2}=a,\ d_{3}=c_{3},  \notag \\
L_{1} &=&1/4\,\mathrm{m},\ L_{2}=1/2\,\mathrm{m},\ P_{1}=a/2,\
P_{2}=d_{2}/2,\ P_{3}=0,\ \alpha =\pi /6,\ \varphi _{0}=\pi /4.
\end{eqnarray}

\textbf{Joint screw coordinates:} Basis for the kinematics modeling, are the
joint screw coordinates in (\ref{fk}), which are defined as%
\begin{equation}
\mathbf{Y}_{i\left( l\right) }=\left( \mathbf{e}_{i\left( l\right) },\mathbf{%
y}_{i\left( l\right) }\times \mathbf{e}_{i\left( l\right) }\right) ^{T},l=1,2
\end{equation}%
where $\mathbf{y}_{i\left( l\right) }\in {\mathbb{R}}^{3}$ denotes the
position vector of a point on the axis of joint $i$, and $\mathbf{e}%
_{i\left( l\right) }\in {\mathbb{R}}^{3}$ is a unit vector along the axis.
Both vectors are resolved in the spatial IFR $\mathcal{F}_{0}$ shown in Fig. %
\ref{figIRSBot}b). In the zero reference configuration in Fig. \ref%
{figIRSBot}, the joint position and unit direction vectors are thus%
\begin{eqnarray*}
\mathbf{y}_{{1}} &=&(a,0,0)^{T},\ \mathbf{y}_{{2}}=(a+{L}_{{1}}\sin {\varphi 
}_{{0}},0,-{L}_{{1}}\cos {\varphi }_{{0}})^{T},\ \mathbf{y}_{{3}}=(a+{b}_{{1}%
},0,{b}_{{3}})^{T} \\
\mathbf{y}_{{4,1}} &=&\mathbf{y}_{{4,2}}=(a+{c}_{{1}}+{d}_{{1}}+{L}_{{1}%
}\sin {\varphi }_{{0}},{d}_{{2}},-{c}_{{3}}-{d}_{{3}}-{L}_{{1}}\cos {\varphi 
}_{{0}})^{T} \\
\mathbf{y}_{{5,1}} &=&\mathbf{y}_{{5,2}}=(a+{c}_{{1}}+{d}_{{1}}+{L}_{{1}%
}\sin {\varphi }_{{0}},-{d}_{{2}},-{c}_{{3}}-{d}_{{3}}-{L}_{{1}}\cos {%
\varphi }_{{0}})^{T} \\
\mathbf{y}_{{6,1}} &=&\mathbf{y}_{{6,2}}=({p}_{{1}},-{p}_{{2}},-{c}_{{3}}-{d}%
_{{3}}-{L}_{{1}}\cos {\varphi }_{{0}}-{L}_{{2}}\cos \psi _{0})^{T}\  \\
\mathbf{e}_{{1}} &=&\mathbf{e}_{{2}}=\mathbf{e}_{{3}}=(0,1,0),\ \mathbf{e}_{{%
4,2}}=\mathbf{e}_{{6,1}}=(\cos \beta \cos {\psi }_{{0}},-\sin \beta \cos {%
\psi }_{{0}},\sin {\psi }_{{0}}),\ \mathbf{e}_{{4,1}}=\mathbf{e}_{{6,2}%
}=(\sin \beta ,\cos \beta ,0) \\
\mathbf{e}_{{5,1}} &=&(-\sin (\beta ),\cos (\beta ),0),\ \mathbf{e}_{{5,2}%
}=(\cos \beta \cos {\psi }_{{0}},\sin \beta \cos {\psi }_{{0}},\sin {\psi }_{%
{0}}).
\end{eqnarray*}%
Therein, and in the following, indexes ${i,1}$ and ${i,2}$ refer to the two
revolute axes of U-joint $i$.

\textbf{Reference configurations:} The reference configuration $\mathbf{A}%
_{i\left( l\right) }$ of body $i\left( l\right) $ in (\ref{Ckl}) involves
the rotation matrix $\mathbf{R}_{i\left( l\right) ,0}\in SO\left( 3\right) $
and position vector $\mathbf{r}_{i\left( l\right) ,0}\in {\mathbb{R}}^{3}$
of the body-fixed frame relative to the IFR $\mathcal{F}_{0}$. According to
Figs. \ref{figGeom1} and \ref{figGeom2}, these are%
\begin{eqnarray*}
\mathbf{R}_{1\left( l\right) ,0} &=&\mathbf{R}_{3\left( l\right) ,0}=\mathrm{%
Rot}\left( y,-\varphi _{0}\right) ,\ \mathbf{R}_{2\left( l\right) ,0}=%
\mathbf{R}_{6\left( l\right) ,0}=\mathbf{I},\ \mathbf{R}_{4\left( l\right)
,0}=\mathrm{Rot}\left( z,-\beta \right) \mathrm{Rot}\left( y,\psi
_{0}\right) ,\ \mathbf{R}_{5\left( l\right) ,0}=\mathrm{Rot}\left( z,\beta
\right) \mathrm{Rot}\left( y,\psi _{0}\right) \\
\mathbf{r}_{1\left( l\right) ,0} &=&\left( a+L_{1}/2\sin \varphi
_{0},0,-L_{1}/2\cos \varphi _{0}\right) ^{T},\ \mathbf{r}_{2\left( l\right)
,0}=\left( a+b_{1}+L_{1}/2\sin \varphi _{0},0,b_{3}-L_{1}/2\cos \varphi
_{0}\right) ^{T} \\
\mathbf{r}_{3\left( l\right) ,0} &=&\left( a+c_{1}+L_{1}\sin \varphi
_{0},0,b_{3}-L_{1}/2\cos \varphi _{0}\right) ^{T},\ \mathbf{r}_{6\left(
l\right) ,0}=\left( 0,0,-c_{3}-d_{1}-P_{3}-L_{1}\cos \varphi _{0}-L_{2}\cos
\psi _{0}\right) ^{T} \\
\mathbf{r}_{4\left( l\right) ,0} &=&\left( (a+c_{1}+d_{1}+p_{1}+L_{1}\sin
\varphi _{0})/2,-(p_{2}+d_{2})/2,-c_{3}-d_{3}-L_{1}\cos \varphi
_{0}-L_{2}/2\cos \psi _{0}\right) ^{T} \\
\mathbf{r}_{5\left( l\right) ,0} &=&\left( (a+c_{1}+d_{1}+p_{1}+L_{1}\sin
\varphi _{0})/2,(p_{2}+d_{2})/2,-c_{3}-d_{3}-L_{1}\cos \varphi
_{0}-L_{2}/2\cos \psi _{0}\right) ^{T}
\end{eqnarray*}%
where $\mathrm{Rot}\left( y,\varphi \right) $ and $\mathrm{Rot}\left(
z,\varphi \right) $ denote the rotation matrices for rotations about the $y$%
- and $z$-axis, respectively.

\subsection{Loop constraints}

A single limb has two independent FC, referred to as the proximal (close to
the base) and the distal loop. According to the spanning tree $\vec{G}%
_{\left( l\right) }$ in fig. \ref{figIRSBotGraph}c) the tree-topology system
is obtained by removing revolute joint 7 and universal joint 8, for whish
cut-joint constraints are introduced.

\textbf{Proximal loop:} The proximal loop, which corresponds to FC $\Lambda
_{1,l}$, is a planar parallelogram 4-bar linkage formed by the R-joints
1,2,3,7. According the spanning tree in Fig. \ref{figIRSBotGraph}c), joints
1,2,3 are tree-joints, and the vector of tree-joint coordinates consists of
the $n_{1,l}$ joint angles $%
\mathbold{\vartheta}%
_{\left( 1,l\right) }=\left( \varphi _{1\left( l\right) },\varphi _{2\left(
l\right) },\varphi _{3\left( l\right) }\right) ^{T}$. The cut-joint
constraints for the revolute joint 7 thus yields $m_{1,l}=2$ independent
(translation) constraints, so that FC $\Lambda _{1,l}$ has $\delta _{1,l}=1$
DOF. The revolute joint angle $\varphi _{1\left( l\right) }$ of the actuated
joint 1 is used as independent coordinate, so that $\mathbf{q}_{\left(
1,l\right) }=\left( \varphi _{1\left( l\right) }\right) $ and $\mathbf{y}%
_{\left( 1,l\right) }=\left( \varphi _{2\left( l\right) },\varphi _{3\left(
l\right) }\right) ^{T}$. The planar parallelogram 4-bar is treated ad hoc as
the solution is readily known. The solution of the geometric loop
constraints for FC $\Lambda _{1,l}$ is readily found as $\varphi _{1\left(
l\right) }=\varphi _{3\left( l\right) }=-\varphi _{2\left( l\right) }$, and
the velocity constraints are solved with%
\begin{equation}
\mathbf{H}_{\left( 1,l\right) }=\left( 
\begin{array}{c}
1 \\ 
-1 \\ 
1%
\end{array}%
\right) .  \label{H1lDelta}
\end{equation}%
\textbf{Distal loop:} The distal loop is formed by U-joints 4,5,6,8. The
loop constraints for FC $\Lambda _{2,l}$ thus involves the $n_{2,l}=3\cdot 2$
tree-joint coordinates $%
\mathbold{\vartheta}%
_{\left( 2,l\right) }=\left( \varphi _{4,1\left( l\right) },\varphi
_{4,2\left( l\right) },\varphi _{5,1\left( l\right) },\varphi _{5,2\left(
l\right) },\varphi _{6,1\left( l\right) },\varphi _{6,2\left( l\right)
}\right) ^{T}$. The universal joint 8 imposes $m_{2,l}=4$ cut-joint
constraints. These constraints are non-redundant, and the FC $\Lambda _{2,l}$
has $\delta _{2,l}=2$ DOF. The angles of universal joint 6 are used as
independent coordinates, i.e. $\mathbf{q}_{\left( 2,l\right) }=\left(
\varphi _{6,1\left( l\right) },\varphi _{6,2\left( l\right) }\right) ^{T},%
\mathbf{y}_{\left( 2,l\right) }=\left( \varphi _{4,1\left( l\right)
},\varphi _{4,2\left( l\right) },\varphi _{5,1\left( l\right) },\varphi
_{5,2\left( l\right) }\right) ^{T}$. Constraints are introduced with the
general formulation in appendix \ref{secCutConstr}.

\begin{figure}[h]
\centerline{%
\includegraphics[height=10cm]{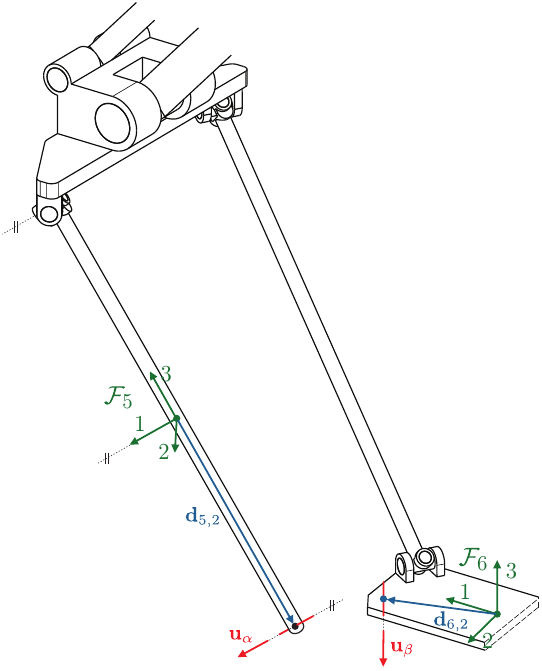}}
\caption{Geometry of cut-joint (U-joint 8).}
\label{figCutGeom}
\end{figure}

The U-joint imposes three position constraints between body $k=6=\mathrm{P}$
(the platform) and $r=5$ of the form (\ref{PosConst}). In the following, the
index $\left( l\right) $ is omitted. The body-fixed position vectors,
expressed in frame $\mathcal{F}_{r=5}$ and platform frame $\mathcal{F}_{k=6}$
(see Fig. \ref{figCutGeom}), in (\ref{PosConst2}) are ${^{k}}\mathbf{d}%
_{k,\lambda }={^{6}}\mathbf{d}_{6,2}=\left( P_{1},P_{2},P_{3}\right) ^{T},\ {%
^{r}}\mathbf{d}_{r,\lambda }={^{5}}\mathbf{d}_{5,2}=\left(
0,0,-L_{2}/2\right) ^{T}$, with the parameters introduced above. The
relative displacement vector expressed in platform frame $\mathcal{F}_{k=%
\mathrm{P}}$ is ${^{k}}\mathbf{d}_{\lambda }=\mathbf{R}_{k,r}{(}\mathbf{r}%
_{r}-\mathbf{r}_{k})+\mathbf{R}_{k,r}{^{r}}\mathbf{d}_{r,\lambda }-{^{k}}%
\mathbf{d}_{k,\lambda }$, where $\mathbf{R}_{k,r}=\mathbf{R}_{k}^{T}\mathbf{R%
}_{r}$, with the absolute rotation matrices $\mathbf{R}_{k},\mathbf{R}_{r}$
of body $r=5$ and platform $r=6$, and their position vectors $\mathbf{r}_{r},%
\mathbf{r}_{k}$, respectively, determined by (\ref{Ckl}). As no relative
translation is permitted, the three position constraints are ${^{k}}\mathbf{d%
}_{k,\lambda }=\mathbf{0}$. The constraint Jacobian in (\ref{Jlambda})
comprises the submatrices $\mathbf{J}_{5,2}=\left( \mathbf{0},\mathbf{0},%
\mathbf{J}_{5,1},\mathbf{J}_{5,2},\mathbf{0},\mathbf{0}\right) $ and $%
\mathbf{J}_{6,2}=\left( \mathbf{J}_{4,1},\mathbf{J}_{4,2},\mathbf{0},\mathbf{%
0},\mathbf{J}_{6,1},\mathbf{J}_{6,2}\right) $.

The orientation constraints (\ref{OriConstr}) are formulated with ${^{r}}%
\mathbf{u}_{\alpha }={^{5}}\mathbf{u}_{\alpha }=\left( 1,0,0\right) ^{T}$,
which is the unit vector along the rotation axis of the U-joint 8 that is
constant relative to body $r=5$, and ${^{k}}\mathbf{u}_{\beta }={^{6}}%
\mathbf{u}_{\beta }=\left( -\sin \beta ,\cos \beta ,0\right) ^{T}$ is the
unit vector along the other rotation axis of the U-joint that is constant
relative to the platform $k=6$. The single orientation constraint is $%
g_{\alpha ,\beta }^{\mathrm{rot}}:={^{k}}\mathbf{u}_{\alpha }^{T}\mathbf{R}%
_{k,r}{^{r}}\mathbf{u}_{\beta }$. This, along with the three position
constraints, gives rise to $m_{2,l}=4$ loop constraints for $\Lambda
_{2\left( l\right) }$. The corresponding velocity and acceleration
constraints are found with (\ref{gRotDot}) and (\ref{gRot2Dot}).

Overall, the kinematics of limb $l$ is described by $n_{l}=9$ joint
variables subjected to $m_{l}=6$ constraints, and has $\delta _{l}=3$ DOF.
The solution of velocity constraints is given by (\ref{Hl}) in terms of the
independent velocities $\dot{\mathbf{q}}_{\left( l\right) }=\left( \dot{%
\varphi}_{1\left( l\right) },\dot{\varphi}_{6,1\left( l\right) },\dot{\varphi%
}_{6,2\left( l\right) }\right) ^{T}$.

\subsection{Inverse Kinematics}

The platform attached to a separated limb of the IRSBot-2 has $\delta _{l}=3$
DOF, and the spatial position of the platform can be used as independent
components of the platform motion. The task space Jacobian $\mathbf{L}_{%
\mathrm{t}\left( l\right) }$ in (\ref{VpD}) thus comprises the last three
rows of $\mathbf{L}_{\mathrm{p}\left( l\right) }$. Since the platform DOF is 
$\delta _{\mathrm{P}}=2$, and the task velocity vector $\mathbf{V}_{\mathrm{t%
}}=\left( v_{1},v_{3}\right) ^{T}$ contains two translation components only,
the limbs (and thus the PKM) are non-equimobile. The generalized velocity of
the separated limb $l=1,2$ is computed from the prescribed task velocity via
(\ref{Ftheta2}), and the tree-joint velocities via (\ref{HFl}), with the
velocity distribution matrix $\mathbf{D}_{\mathrm{t}\left( l\right) }$ in (%
\ref{DtDelta}). The $2\times 2$ inverse kinematics Jacobian $\mathbf{J}_{%
\mathrm{IK}}$ in (\ref{IK}) consists of the first column of $\mathbf{F}%
_{\left( 1\right) }$ and of $\mathbf{F}_{\left( 2\right) }$.

\subsection{Numerical Results}

The inverse dynamics is solved with a sampling time of $\Delta t=0.01\,$s.
The EOM (\ref{EOMTask}) were implemented in closed form, and numerically
evaluated for a prescribed EE-motion using Matlab. All results were
validated against the commercial MBS simulation tool Alaska \cite{AlaskaWeb}%
. With the chosen parameters, the reference position of the platform for the
zero reference configuration $%
\mathbold{\vartheta}%
=\mathbf{0}$ is $\mathbf{r}_{6,0}=(0,0,-1/24(4+6\sqrt{2}+\sqrt{6(37-6\sqrt{2}%
-2\sqrt{3}-4\sqrt{6})}))^{T}$. The platform translation was prescribed as a
parabolic curve $\mathbf{r}_{6}\left( t\right) =\mathbf{r}_{6,0}+(\Delta
x\cdot s\left( t\right) ,0,\Delta z\cdot 4s\left( t\right) \left( 1-s\left(
t\right) \right) )^{T}$, with motion profile $s\left( t\right) =\sin
^{2}\left( \nu \pi t/T\right) $, and $\Delta x=0.25$\thinspace m$,\Delta
z=0.45$\thinspace m, where the EE has to perform $\nu =3$ cycles per second.
The EE-path is indicated in fig. \ref{figGeom1}. In the following only the
results for limb $l=1$ are presented.

\subsubsection{Inverse kinematics and constraint satisfaction}

\begin{figure}[h]
\centerline{a)\includegraphics[height=5.1cm]{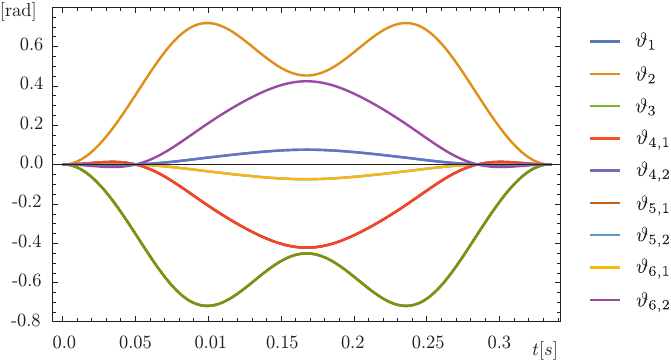}} \; \newline
\centerline{b)\includegraphics[height=5.1cm]{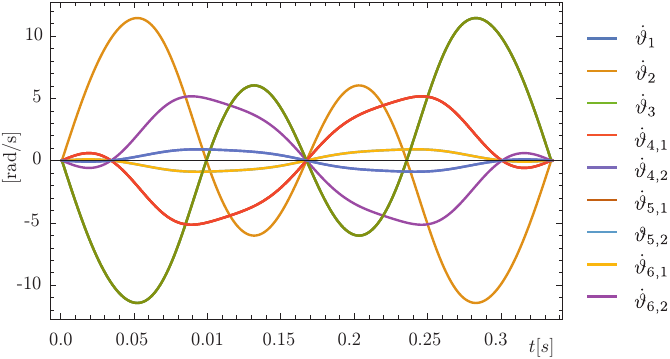}}
\caption{Inverse kinematics solution for a) tree-joint angles, and b)
velocities of limb $l=1$.}
\label{figIK1}
\end{figure}

\begin{figure}[h]
\centerline{a)\includegraphics[height=5.1cm]{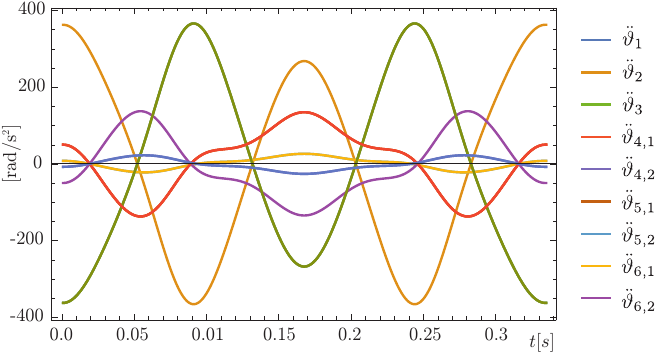}} \; \newline
\centerline{b)\includegraphics[height=5.1cm]{figures/QQFig.pdf}}
\caption{a) Inverse kinematics solution for tree-joint accelerations of limb 
$l=1$. b) Inverse dynamics solution (drive torques $u_{1}$ and $u_{2}$ at
joint 1 of limb $l=1$ and $l=2$).}
\label{figIK2}
\end{figure}

\begin{figure}[h]
\centerline{a)\includegraphics[height=5.1cm]{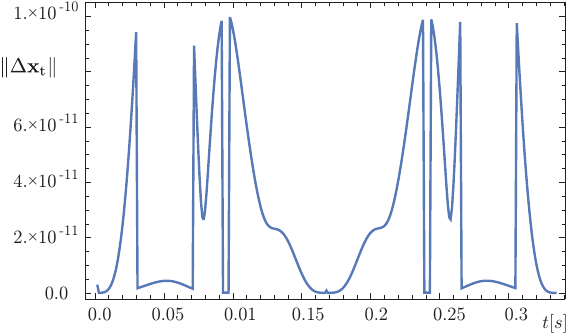} ~~\hfil
b)\includegraphics[height=5.1cm]{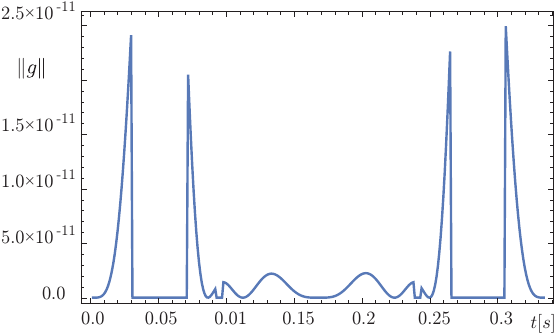}}
\caption{a) Error of outer loop solving the inverse kinematics of limb, and
b) of inner loop solving the loop constraints of FC $\Lambda _{2\left(
l\right) }$.}
\label{figErrorAlg}
\end{figure}

\begin{figure}[h]
\centerline{a)%
\includegraphics[width=8.4cm]{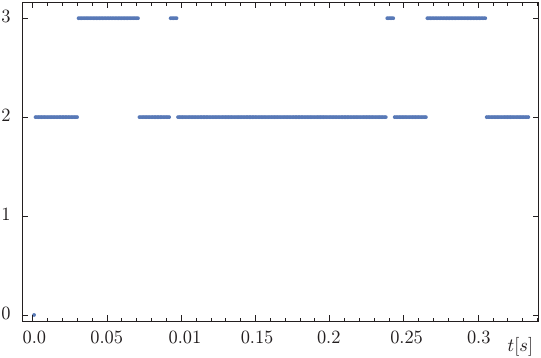}~~~~~b)%
\includegraphics[width=8.4cm]{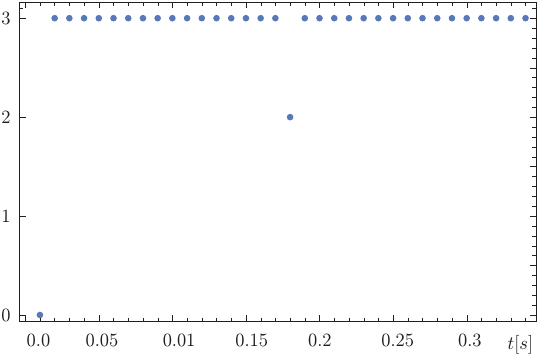}}
\caption{Number of iterations required for solving the inverse kinematics of
limb (algorithm 1) with precision goal $\protect\varepsilon =10^{-10}$,
where a) $\Delta t=0.001\,$s, and b) $\Delta t=0.01\,$s.}
\label{figIterAlg1}
\end{figure}

\begin{figure}[h]
\centerline{%
\includegraphics[width=14.1cm]{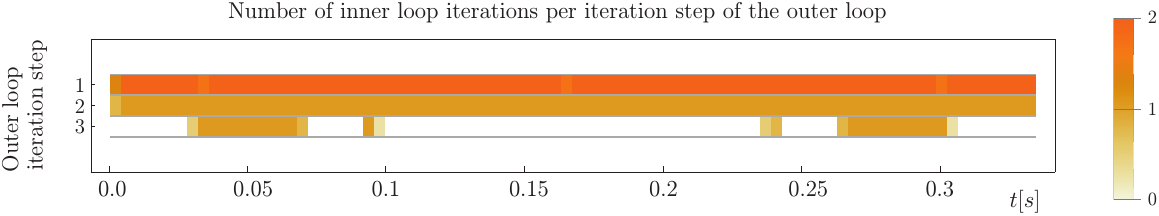}}
\caption{Number of iterations required for solving the loop constraints for
FC $\Lambda _{2\left( l\right) }$ (algorithm 2) within each iteration of
algorithm 1 solving the inverse kinematics of limb with accuracy goal $%
\protect\varepsilon =10^{-10}$. }
\label{figIterAlg2}
\end{figure}

\textbf{Application of Nested Algorithms 1 and 2:} The inverse kinematics is
solved for the trajectory sampled with time step size $\Delta t=0.001\,$s.
Fig. \ref{figIK1} and Fig. \ref{figIK2}a) show the trajectories of all
tree-joint variables. The inverse kinematics problem of the limb along with
the loop constraints for FC $\Lambda _{2\left( l\right) }$ are solved with
the nested algorithms 1 and 2, respectively. The error threshold was set to $%
\varepsilon =10^{-10}$ in both algorithms. Fig. \ref{figErrorAlg} shows the
evolution of the error $\left\Vert \Delta \mathbf{x}_{\mathrm{t}}\right\Vert 
$ of the inverse kinematics (algorithm 1) and the error $\left\Vert
g\right\Vert $ of the cut-joint constraints of FC $\Lambda _{2\left(
l\right) }$ (algorithm 2). The number of outer-loop iteration steps
necessary for solving the limb inverse kinematics is shown in Fig. \ref%
{figIterAlg1}a). Between 1 and 3 iterations of the outer-loop (algorithm 1)
are needed, which depends on the rate of change of the EE configuration and
the corresponding generalized coordinates within one time step. The number
of inner-loop iterations needed for solving the cut-joint constraints
(algorithm 2) is shown in Fig. \ref{figIterAlg2}. Shown is the number of
inner-loop iterations needed to solve the loop constraints within each
particular outer-loop step for the inverse kinematics solution. From 1 to 3
inner-loop iterations are required to satisfy the constraints with the
desired accuracy $\varepsilon $. The number of iterations depends on the
rate of change of the tree-joint variable of the loop and the conditioning
of the constraints. If the trajectory is sampled with a larger time step
size of $\Delta t=10^{-2}\,$s, algorithm 1 needs 3 iterations almost all the
time as shown in Fig \ref{figIterAlg1}b). Then the error evolution is
similar to that in Fig. \ref{figIterAlg1} and \ref{figIterAlg2}. 

\clearpage%

\begin{figure}[h]
\centerline{a)\includegraphics[width=10cm]{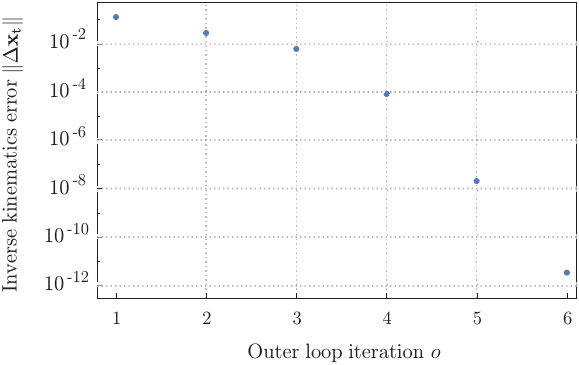} ~~
b)\includegraphics[width=7cm]{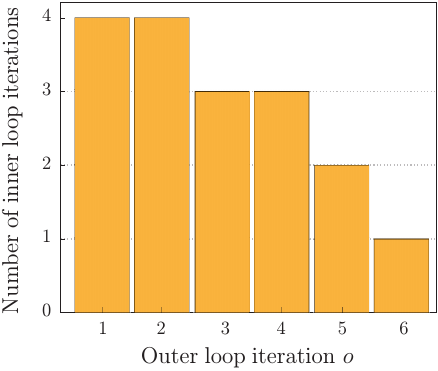}}
\caption{a) Convergence of inverse kinematics error during outer-loop
iterations. b) Number of inner-loop iterations required for solving the loop
constraints of FC $\Lambda _{2\left( l\right) }$ at each iteration step of
the outer-loop solving the inverse kinematics.}
\label{figTestErrorAlg1}
\end{figure}

The Newton-Raphson scheme is known to have a quadratic convergence, and it
is instructive to consider this for the inner- and outer-loop separately. As
an example, the inverse kinematics is solved for a prescribed target
position of the platform $\mathbf{r}_{6,0}+\left( 0.2\,\mathrm{m},0,0.5\,%
\mathrm{m}\right) ^{T}$, with initial configuration $\mathbf{r}_{6,0}$. The
accuracy goal is $\varepsilon =10^{-11}$ for both algorithms. The outer-loop
needs 6 iterations to converge. Fig. \ref{figTestErrorAlg1}a) shows the
convergence of inverse kinematics error. The number of necessary iterations
of the inner-loop differs for each step of the outer-loop. Fig. \ref%
{figTestErrorAlg1}b) shows the number of inner-loop iteration cycles at each
outer-loop step $o$. The corresponding convergence of the constraint error
at each step $o$ of the inverse kinematics outer-loop is shown in Fig. \ref%
{figTestErrorAlg2}. In order to compare the convergence rate, the error $%
\left\Vert g\right\Vert $ is normalized relative to its initial value, i.e.
shown is the error decay not its absolute value, which is always below $%
\varepsilon $ after the last step. Step $i=0$ refers to the initial value,
where $\left\Vert g\right\Vert =1$. 
\begin{figure}[h]
\centerline{\includegraphics[width=13cm]{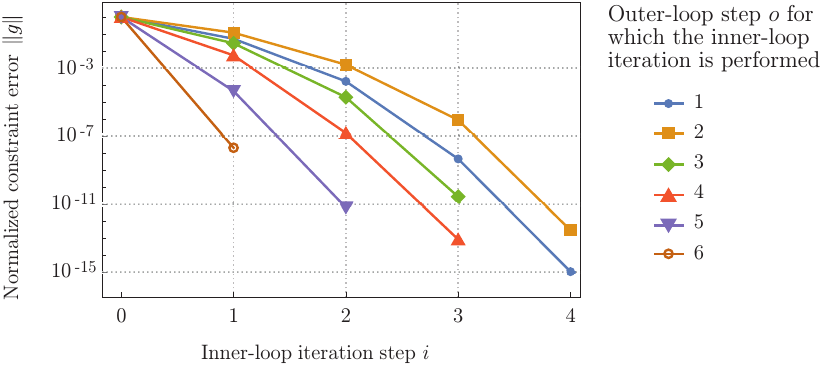}}
\caption{Convergence of constraint error at each step of the outer-loop for
solving the inverse kinematics of limb.}
\label{figTestErrorAlg2}
\end{figure}

\begin{figure}[h]
\centerline{a)~\includegraphics[width=8.3cm]{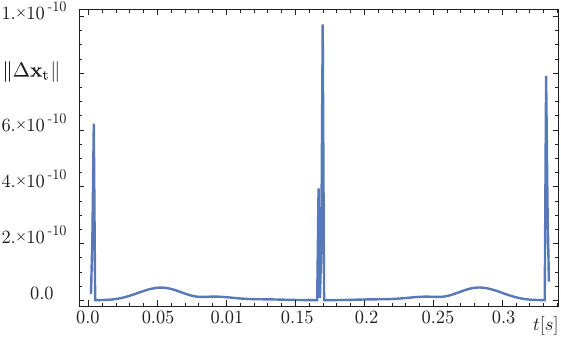}\hfil ~~
b)~\includegraphics[width=8.3cm]{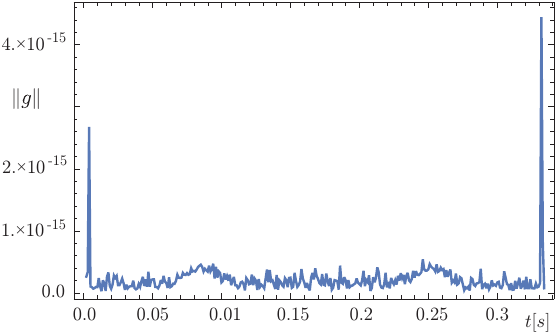}}
\caption{Convergence of a) inverse kinematics and b) constraint error when
using algorithm 3 for accuracy goal $\protect\varepsilon =10^{-10}$.}
\label{figErr1Alg3}
\end{figure}

\textbf{Application of Algorithm 3:} Nesting algorithm 2 (inner-loop) with
algorithm 1 (outer-loop) ensures that, at each outer-loop iteration step for
solving the limb inverse kinematics, the constraints are satisfied with
prescribed accuracy, which corresponds to numerically evaluating (\ref{fl}),
i.e. the inverse of the combined map (\ref{phipl}). This is crucial when the
loop constraints are ill-conditioned. The loop constraints for FC $\Lambda
_{2\left( l\right) }$ of the IRSBot-2 are well-conditioned, however, and
there are no singularities within the used workspace. Separate numerical
solution of the individual constraints is thus not necessary. Instead, both
problems can be solved at once with algorithm 3 using a common iteration
step. Fig. \ref{figErr1Alg3} shows the error evolution, and Fig. \ref%
{figIterAlg3} the necessary number of iterations when algorithm 3 is applied
with accuracy goal $\varepsilon =10^{-10}$. The algorithm needs three
iterations at almost every time step. The total number of iterations is
almost the same as that needed by the nested algorithm, which in average
needs two outer loop iteration steps with one and two inner loop iterations,
respectively. 
\clearpage%
\begin{figure}[h]
\centerline{\includegraphics[width=9cm]{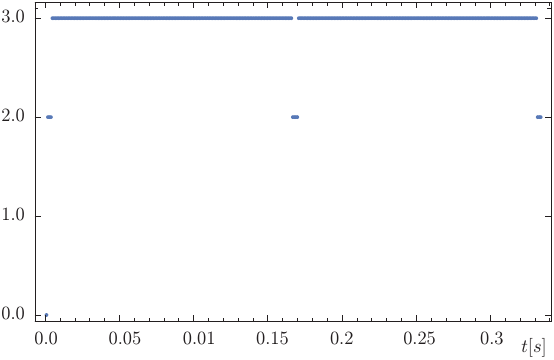}}
\caption{Number of iterations required by algorithm 3 for accuracy goal $%
\protect\varepsilon =10^{-10}$.}
\label{figIterAlg3}
\end{figure}

\subsubsection{Inverse dynamics}

\begin{figure}[h]
\centerline{\includegraphics[width=6.5cm]{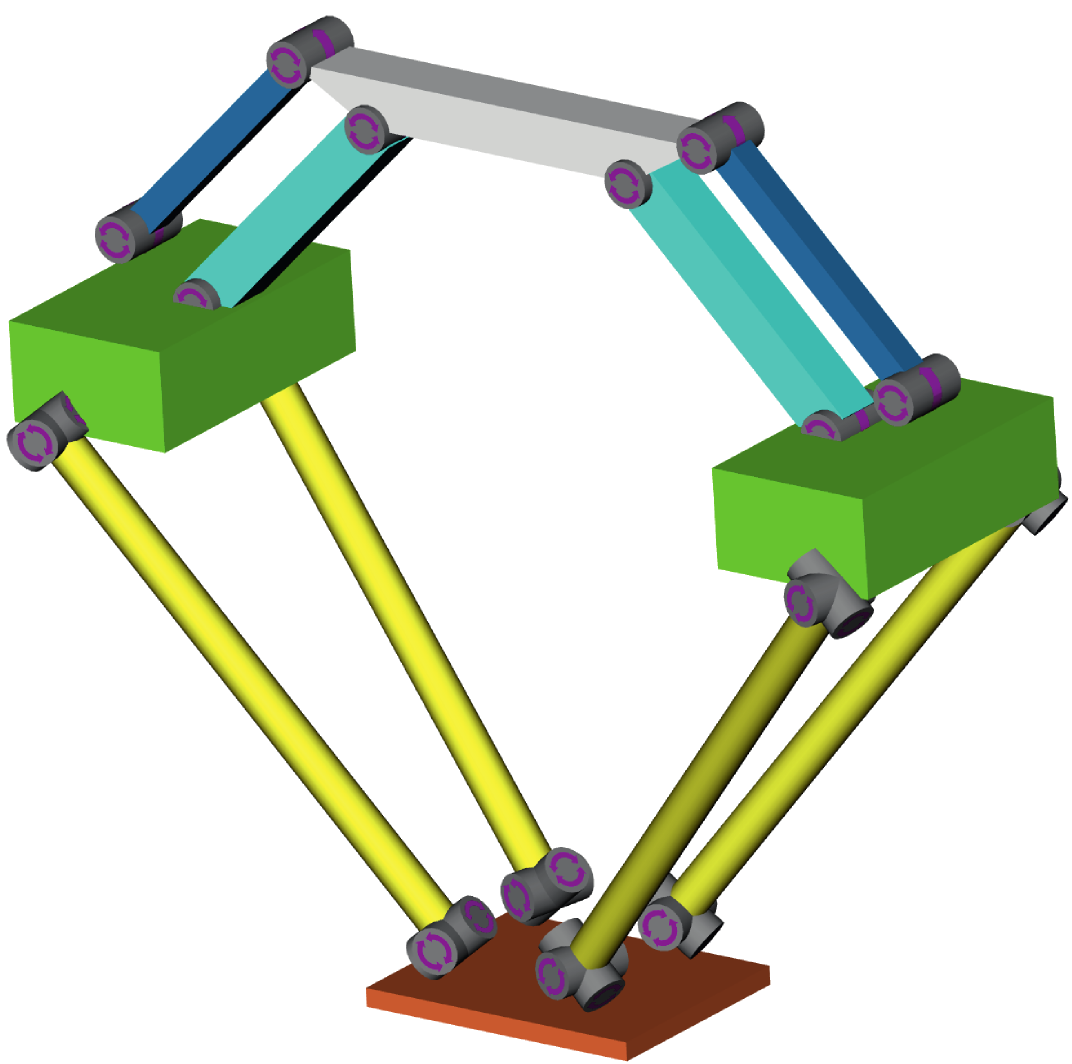}}
\caption{Rendering of the dynamics model using geometric primitives. }
\label{figAlaska}
\end{figure}

\begin{figure}[h]
\centerline{\includegraphics[width=11cm]{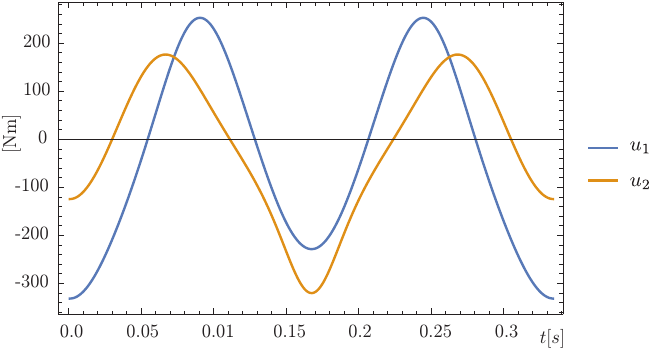}}
\caption{Inverse dynamics solution: drive tourque $u_{1}$ and $u_{2}$ at the
actuated joint 1 of limb $l=1,2$. }
\label{figInvDyn}
\end{figure}

The inverse dynamics solution (\ref{invDyn}) is evaluated for the joint
trajectories $%
\mathbold{\vartheta}%
_{\left( l\right) }\left( t\right) ,l=1,2$ obtained from the inverse
kinematics solution, assuming no additional external forces other than
gravity. The inertia properties of the links are estimated by modeling the
bodies as geometric primitives. All bodies are assumed to be made of
aluminum with homogenous mass distribution. Body 1 is modeled as a rigid
beam with length $L_{1}=1/2\,$m, and square cross section with side length $%
L_{1}/6$. Body 3 is a beam with length $L_{1}$ and side length $L_{1}/10$.
Body 2 is simplified as a rectangular solid plate with side lengths $2\left(
c_{1}+d_{1}\right) $ and $2d_{2}$, and thickness $c_{3}+d_{3}$. Links 4 and
5 are modeled as rigid beams with lengths $L_{2}=3/4\,$m, and circular cross
section with radius $L_{2}/30$. The platform is a solid rectangular plate
with side lengths $4P_{1}$ and $3P_{2}$, and height $P_{2}/4$. A rendering
of the model is shown in fig. \ref{figAlaska}. With the above numerical
data, the mass $m_{i}$ of body $i=1,\ldots ,5$ of each of the two identical
limbs, and the platform $i=6$ are $m_{1}=1.17188\,$kg, $m_{2}=8.11899\,$kg, $%
m_{3}=21.1875\,$kg, $m_{4}=m_{5}=1.1781\,$kg, $m_{6}=1.97754\,$kg. The
inverse dynamics solution obtained by evaluating (\ref{invDyn}) is shown in
Fig. \ref{figInvDyn}, where $u_{1}$ and $u_{2}$ are the drive torques at
joint 1 of limb $l=1$ and $l=2$, respectively. These are the feedforward
control commands to be used model-based non-linear control schemes \cite%
{AngelesBook,BriotKhalilBook2004,Robotics2009,Murray}. 
\newpage%

\subsubsection{Input Singularity}

\begin{figure}[h]
\centerline{a)\includegraphics[width=8.5cm]{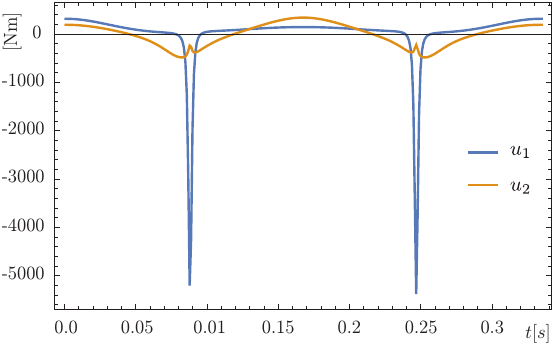}~~~b)%
\includegraphics[width=8cm]{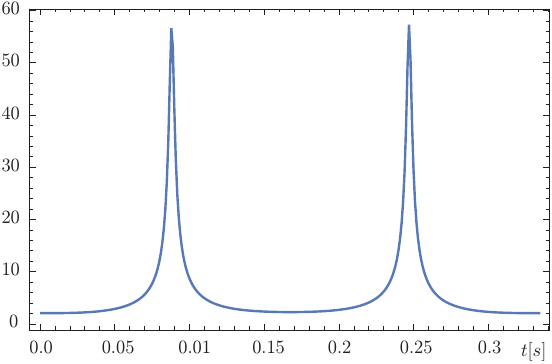}}
\caption{{}a) Inverse dynamics solution (drive torques $u_{1}$ and $u_{2}$)
for the EE-trajectory passing an input singualirity. b) Time evolution of $%
\protect\sqrt{\protect\kappa (\mathbf{F}_{\left( l\right) }^{T}\mathbf{F}%
_{\left( l\right) })}$ along that trajectory.}
\label{figInvDynsing}
\end{figure}

For completeness, the robustness of the proposed numerical solution scheme
is investigated when the PKM passes near an input singularity (also called
serial singularity \cite{GosselinAngeles1990}). To this end, the platform
motion is prescribed as above with $\Delta x=-0.15$\thinspace m, $\Delta
z=-0.6085$\thinspace m, which leads to limb 1 becoming stretched out at the
inflection points of the platform trajectory. As a consequence, the task
space Jacobian $\mathbf{L}_{\mathrm{t}\left( l\right) }$ in (\ref{Vt}), and
thus $\mathbf{F}_{\left( l\right) }$ in (\ref{Ftheta2}), become
ill-conditioned. The inverse kinematics solution is shown in Fig. \ref%
{figIKsing}. The closeness to the singularity causes a fast change in the
motion direction of the limb leading to large accelerations. This is also
visible in the inverse dynamics solution in Fig. \ref{figInvDynsing}a). The
evolution of $\sqrt{\kappa (\mathbf{F}_{\left( l\right) }^{T}\mathbf{F}%
_{\left( l\right) })}$, where $\kappa $ is the condition number, clearly
indicates bad conditioning of the inverse kinematics Jacobian of the limb
near the singularity (Fig. \ref{figInvDynsing}b). The necessary number of
iterations of the inverse kinematics algorithm 1, shown in Fig. \ref%
{figIterSing}, is almost unaffected, however. Also the inner-loop needs at
most two iterations to achieve the accuracy of $\varepsilon =10^{-10}$. 
\begin{figure}[h]
a)\includegraphics[width=11cm]{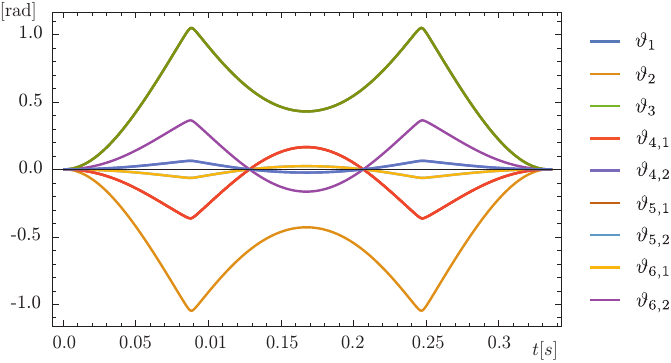}\; \newline
b)\includegraphics[width=11cm]{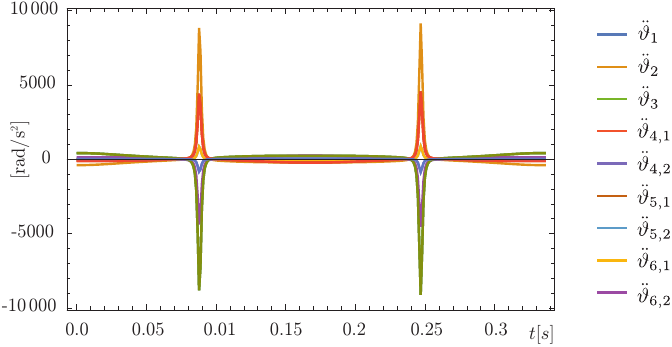}
\caption{a) Tree-joint variables according to the EE-trajectory passing near
an input singularity. b) The corresponding tree-joint accelerations.}
\label{figIKsing}
\end{figure}
\begin{figure}[h]
\centerline{%
\includegraphics[width=8.5cm]{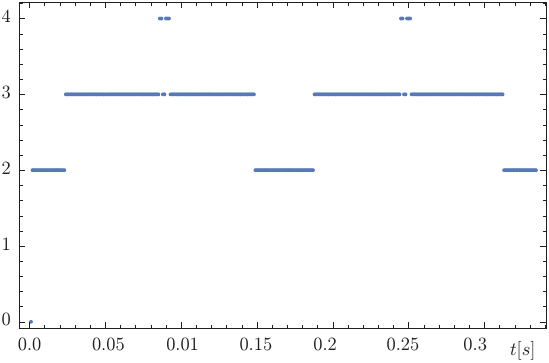}}
\caption{{}Number of itereations needed by the outer-loop algorithm 1 to
solve the limb inverse kinematics with a a precision goal $\protect%
\varepsilon =10^{-10}$.}
\label{figIterSing}
\end{figure}

\section{Conclusion%
\label{secConclusion}%
}

A local constraint embedding approach was presented for the modeling of PKM
with complex loops, and incorporated into the task space formulation of EOM.
The obtained model provides the basis for model-based non-linear control as
well as for time simulation of the PKM dynamics. The constraint embedding
along with the kinematic model of PKM can also be employed to perform
kinematic simulations. For the numerical implementation of the constraint
embedding method, a nested Newton-Raphson algorithm 1 is proposed, which
iteratively solves the inverse kinematics of a limb making use of the
iterative solution of the inner-limb loop constraints (algorithm 2). This
admits dealing with redundant (or general ill-conditioned) inner-loop
constraints. The latter occur for several PKM of practical relevance. A good
example is the 3\underline{R}[2UU] Delta robot, where each limb contains a
single loop modeled as a 4U loop. For this 4U loop, four constraints
introduced, but only three of them are independent. This redundancy can be
dealt with for each of the individual loops in algorithm 2. This is an
advantage of the proposed local constraint embedding strategy, and a clear
advantage over the standard MBS formulation. If the individual constraints
are regular, as in case of the IRSBot-2 example, the loop constraints can be
solved along with the limb inverse kinematics constraints with algorithm 3.
An important aspect of the introduced formulation is that the intra-limb
constraint, the inverse kinematics problem, and the EOM (\ref{EOMLimb}) of
the tree-topology system can be solved and evaluated independently. This
allows for parallel computation in order to increase the computational
efficiency The formulation or algorithm for evaluating the EOM (\ref{EOMLimb}%
) is arbitrary. If appropriate, they can be evaluated with recursive $%
O\left( n\right) $-algorithm so to further improve the efficiency. A method
that is efficient and easy to use at the same time is the Lie group
formulation for EOM in closed form \cite%
{ParkBobrowPloen1995,PloenPark1999,MUBOScrew2} and for the recursive
algorithm \cite{MUBOScrew2,ICRA2017,RAL2020}. Future research will address
adoption of the modular modeling concept \cite{MuellerAMR2020,MMT2022} to
the constraint embedding.%
\appendix%
\newpage%

\section{Cut-Joint Constraints%
\label{secCutConstr}%
}

The cut-joint formulation is well-known in computational multibody system
dynamics \cite{NikraveshBook1988,HaugBook1989,ShabanaBook}. This is
summarized in the following in a form compatible with the presented
formulation. A joint with DOF $\delta $ imposes $m_{\lambda ,l}=6-\delta $
constraints, e.g., a revolute joint imposes $m_{\lambda ,l}=5$, a spherical
(ball-socket) joint imposes $m_{\lambda ,l}=3$, and a universal/hook joint
imposes $m_{\lambda ,l}=4$ constraints. The specific cut-joint constraints
are available for all technical joints. The particular constraints are
obtained by combining \emph{elementary constraints}, i.e. constraints on the
relative translation and rotation of interconnected body.

Notice that in special situations, the system of $m_{\lambda ,l}$
constraints for loop $\Lambda _{\lambda \left( l\right) }$ may by redundant.
Possible redundancy of the overall system of $m_{l}$ intra-limb constraints
for limb $l$, or of the overall system of $m$ constraints that would be
imposed to the PKM model in the standard MBS modeling approach, is not an
issue for the local constraint embedding method.

\subsection{Elementary Cut-Joint Constraints}

The cut-joint of FC $\Lambda _{\lambda \left( l\right) }$ connects body $k$
and $r$ of limb $l$. Denote with $\mathcal{F}_{k\left( l\right) }$ and $%
\mathcal{F}_{r\left( l\right) }$ the body-fixed frame on body $k$ and $r$,
and with $\mathbf{R}_{k\left( l\right) }$ and $\mathbf{R}_{r\left( l\right)
} $ the absolute rotation matrix of $\mathcal{F}_{k\left( l\right) }$ and $%
\mathcal{F}_{r\left( l\right) }$ relative to the IFR, respectively (Fig. \ref%
{figCutJoint}a). They, and the absolute position vectors $\mathbf{r}_{k}$
and $\mathbf{r}_{r}$ are known from the absolute configurations $\mathbf{C}%
_{k}$ and $\mathbf{C}_{r}$. In the following, the index $\left( l\right) $
is omitted as all derivations refer to limb $l$. The rotation matrix of body 
$r$ relative to body $k$ is thus $\mathbf{R}_{k,r}:=\mathbf{R}_{k}^{T}%
\mathbf{R}_{r}$. Let ${^{k}}\mathbf{d}_{k,\lambda }$ and ${^{r}}\mathbf{d}%
_{r,\lambda }$ be the constant position vector of the reference point of the
cut-joint (e.g. rotation center of spherical or universal joints, point at
joint axis of revolute or cylindrical joints) measured in frame $k$ and $r$,
respectively. The velocity and acceleration constraints restrict the
velocities $\mathbf{V}_{r\left( l\right) },\mathbf{V}_{k\left( l\right) }$
and accelerations $\dot{\mathbf{V}}_{r\left( l\right) },\dot{\mathbf{V}}%
_{k\left( l\right) }$ of the two bodies. They are expressed in terms of $%
\dot{%
\mathbold{\vartheta}%
}_{\left( \lambda ,l\right) },\ddot{%
\mathbold{\vartheta}%
}_{\left( \lambda ,l\right) }$ using the geometric Jacobian. In the
following explicit relations for general constraints are presented that are
suited for PKM modeling. The difference to the formulation used in MBS
dynamics is that no dedicated cut-joint frames are introduced. 
\begin{figure}[h]
\centerline{a)\includegraphics[height=6.9cm]{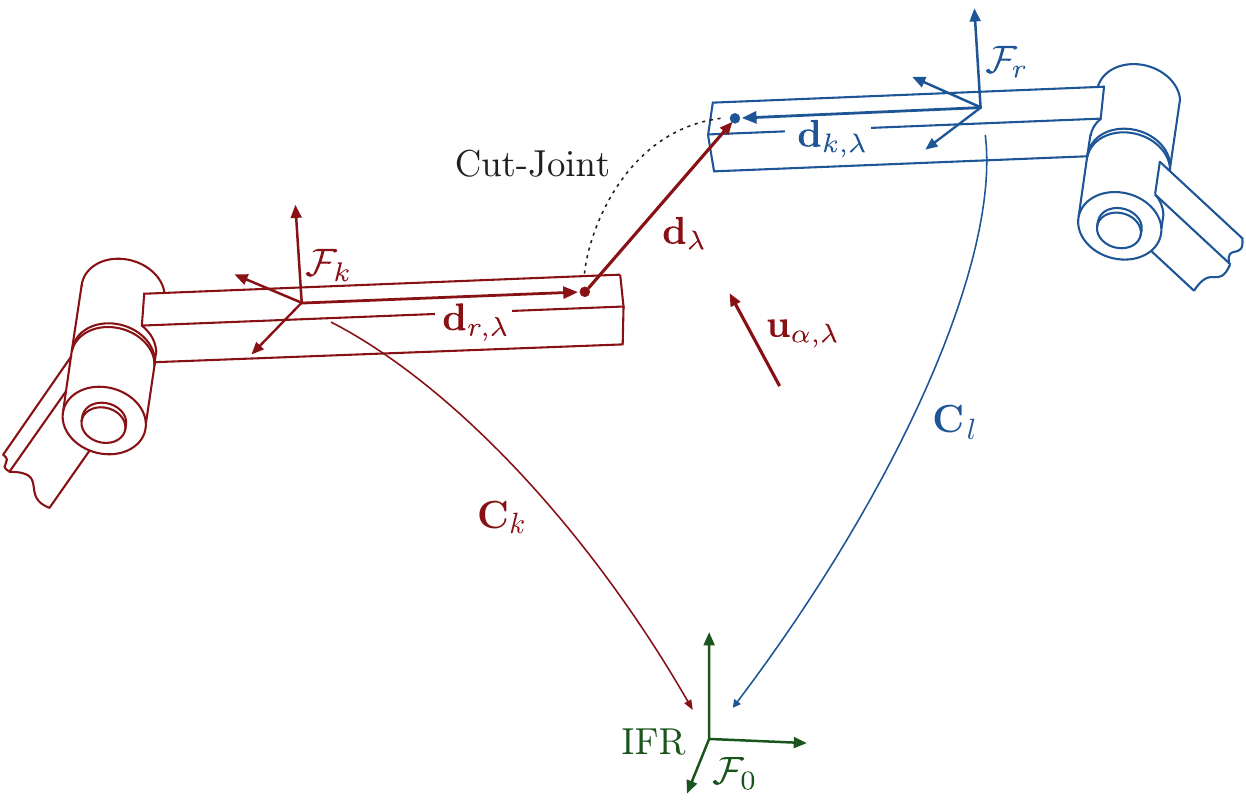}\hfil
b)~~~~\includegraphics[height=3cm]{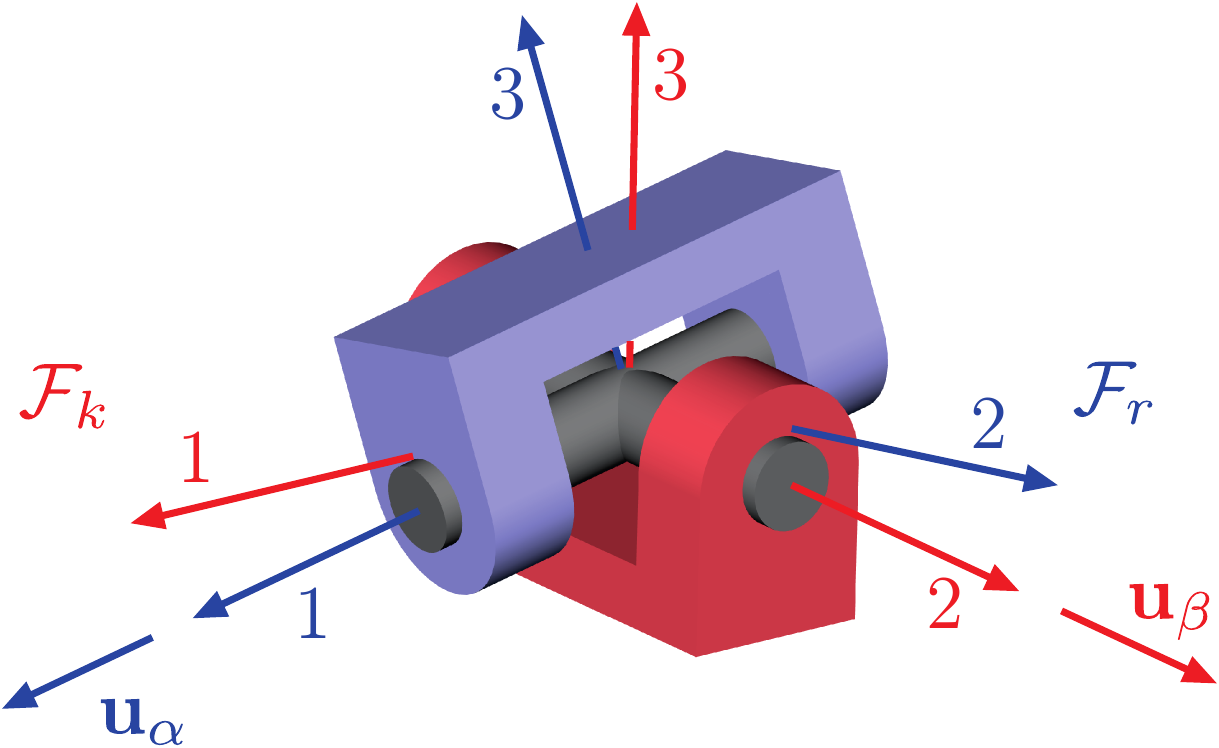}}
\caption{a) Generic definition of cut-joint geometry. b) Definition of two
vectors $\mathbf{u}_{\protect\alpha }$ and $\mathbf{u}_{\protect\beta }$ for
defining the orientation constraint of a U joint. The condition is that they
remain perpendicular.}
\label{figCutJoint}
\end{figure}

\paragraph{Distance constraints:}

The relative translation of the reference point of the cut-joint at body $k$
and $r$, respectively, expressed in $\mathcal{F}_{k}$ is%
\begin{equation}
{^{k}}\mathbf{d}_{\lambda }={^{k}}\mathbf{d}_{r,\lambda }-{^{k}}\mathbf{d}%
_{k,\lambda }+{^{k}}\mathbf{r}_{r}-{^{k}}\mathbf{r}_{k}  \label{PosConst2}
\end{equation}%
with ${^{k}}\mathbf{d}_{r,\lambda }=\mathbf{R}_{k,r}{^{r}}\mathbf{d}%
_{r,\lambda }$ and ${^{k}}\mathbf{r}_{r}=\mathbf{R}_{k}^{T}\mathbf{r}_{r},{%
^{k}}\mathbf{r}_{k}=\mathbf{R}_{k}^{T}\mathbf{r}_{k}$ (Fig. \ref{figCutJoint}%
a). This relative translation is restricted in certain directions, according
to the cut-joint. Denote with ${^{k}}\mathbf{u}_{\alpha ,\lambda \left(
l\right) }$ unit vectors along relative translations that are prohibited by
the cut-joint. The elementary displacement constraint is thus $g_{\alpha
,\lambda }^{\mathrm{pos}}=0$, where%
\begin{equation}
g_{\alpha ,\lambda }^{\mathrm{pos}}:={^{k}}\mathbf{u}_{\alpha ,\lambda }^{T}{%
^{k}}\mathbf{d}_{\lambda }.  \label{PosConst}
\end{equation}%
The relative configuration of the bodies depends on the joint variables of
the FC $\Lambda _{\lambda \left( l\right) }$ only, and thus $g_{\alpha
,\lambda }^{\mathrm{pos}}=g_{\alpha ,\lambda }^{\mathrm{pos}}(%
\mathbold{\vartheta}%
_{\left( \lambda ,l\right) })$.

Using $\dot{\mathbf{R}}_{k,r}=\mathbf{R}_{k,r}{^{r}}\widetilde{\bm{\omega}}%
_{r}-{^{k}}\widetilde{\bm{\omega}}_{k}\mathbf{R}_{k,r}$, the relative
velocity and acceleration of body $k$ and $r$ can be expressed as%
\begin{eqnarray}
{^{k}}\dot{\mathbf{d}}_{\lambda } &=&\mathbf{A}_{\lambda }\dot{%
\mathbold{\vartheta}%
}_{\left( \lambda ,l\right) }=\mathbf{B}_{\lambda }\left( 
\begin{array}{c}
\mathbf{V}_{k} \\ 
\mathbf{V}_{r}%
\end{array}%
\right) \\
{^{k}}\ddot{\mathbf{d}}_{\lambda } &=&\mathbf{A}_{\lambda }\ddot{%
\mathbold{\vartheta}%
}_{\left( \lambda ,l\right) }+\dot{\mathbf{A}}_{\lambda }\dot{%
\mathbold{\vartheta}%
}_{\left( \lambda ,l\right) }=\mathbf{B}_{\lambda }\left( 
\begin{array}{c}
\dot{\mathbf{V}}_{k} \\ 
\dot{\mathbf{V}}_{r}%
\end{array}%
\right) +\dot{\mathbf{B}}_{\lambda }\left( 
\begin{array}{c}
\mathbf{V}_{k} \\ 
\mathbf{V}_{r}%
\end{array}%
\right)
\end{eqnarray}%
with 
\begin{align}
\mathbf{A}_{\lambda }& :=\mathbf{B}_{\lambda }\mathbf{J}_{\lambda } \\
\dot{\mathbf{A}}_{\lambda }& =\mathbf{B}_{\lambda }\dot{\mathbf{J}}_{\lambda
}+\dot{\mathbf{B}}_{\lambda }\mathbf{J}_{\lambda }  \notag
\end{align}%
\begin{align}
\mathbf{B}_{\lambda }& :=%
\Bigg%
(%
\begin{array}{cccc}
{^{k}}\mathbf{d}_{\lambda }+{^{k}\widetilde{\mathbf{d}}}_{r,\lambda } & -%
\mathbf{I} & \ \ \ -{^{k}\widetilde{\mathbf{d}}}_{r,\lambda } & \mathbf{R}%
_{k,r}%
\end{array}%
\Bigg%
)  \notag \\
\dot{\mathbf{B}}_{\lambda }& =%
\Bigg%
(%
\begin{array}{cccc}
{^{k}}\dot{\mathbf{d}}_{\lambda } & \ \ \ \mathbf{0} & \ \ \ \ \ ({^{k}}%
\widetilde{\bm{\omega}}_{r}-{^{k}}\widetilde{\bm{\omega}}_{k}){^{k}%
\widetilde{\mathbf{d}}}_{r,\lambda }\mathbf{R}_{k,r} & \ \ \ ({^{k}}%
\widetilde{\bm{\omega}}_{r}-{^{k}}\widetilde{\bm{\omega}}_{k})\mathbf{R}%
_{k,r}%
\end{array}%
\Bigg%
)  \notag \\
\mathbf{J}_{\lambda }& :=\left( 
\begin{array}{c}
\mathbf{J}_{k,\lambda } \\ 
\mathbf{J}_{r,\lambda }%
\end{array}%
\right)  \label{Jlambda}
\end{align}%
where $\mathbf{J}_{k,\lambda \left( l\right) }$ and $\mathbf{J}_{r,\lambda
\left( l\right) }$ are the submatrices of the Jacobians $\mathbf{J}_{k\left(
l\right) }$ and $\mathbf{J}_{r\left( l\right) }$ of body $k$ and $r$,
respectively, with the $n_{\lambda ,l}$ columns corresponding to the joint
variables of FC $\Lambda _{\left( l\right) }{^{k}}$, and $\bm{\omega}_{r}=%
\mathbf{R}_{k,r}{^{r}}\bm{\omega}_{r}$. The Jacobians and their time
derivatives are given in (\ref{VSys}) and (\ref{JSysDot}). The velocity and
acceleration constraints are thus $\dot{g}_{\alpha ,\lambda }^{\mathrm{pos}%
}=0$ and $\ddot{g}_{\alpha ,\lambda }^{\mathrm{pos}}=0$, with%
\begin{align}
\dot{g}_{\alpha ,\lambda }^{\mathrm{pos}} &:={^{k}}\mathbf{u}_{\alpha
,\lambda }^{T}{^{k}}\dot{\mathbf{d}}_{\lambda }={^{k}}\mathbf{u}_{\alpha
,\lambda }^{T}\mathbf{A}_{\lambda }\dot{%
\mathbold{\vartheta}%
}_{\left( \lambda ,l\right) }  \label{gPosDot} \\
\ddot{g}_{\alpha ,\lambda }^{\mathrm{pos}} &:={^{k}}\mathbf{u}_{\alpha
,\lambda }^{T}{^{k}}\ddot{\mathbf{d}}_{\lambda }={^{k}}\mathbf{u}_{\alpha
,\lambda }^{T}\mathbf{A}_{\lambda }\ddot{%
\mathbold{\vartheta}%
}_{\left( \lambda ,l\right) }+{^{k}}\mathbf{u}_{\alpha ,\lambda }^{T}\dot{%
\mathbf{A}}_{\lambda }\dot{%
\mathbold{\vartheta}%
}_{\left( \lambda ,l\right) }.  \label{gPos2Dot}
\end{align}

\paragraph{Orientation constraints:}

A constraint on the relative orientation of the two bodies is described by
the condition that the angle between a vector ${^{k}}\mathbf{u}_{\alpha }$
fixed on body $k$ and a vector ${^{r}}\mathbf{u}_{\beta }$ fixed at body $r$
remains constant. The particular (and most relevant) situation that these
two vectors are perpendicular is expressed by the constraint $g_{\alpha ,b}^{%
\mathrm{rot}}=0$, where%
\begin{equation}
g_{\alpha ,\beta }^{\mathrm{rot}}:={^{k}}\mathbf{u}_{\alpha }^{T}\mathbf{R}%
_{k,r}{^{r}}\mathbf{u}_{\beta }={^{r}}\mathbf{u}_{\alpha }^{T}{^{k}}\mathbf{u%
}_{\beta }  \label{OriConstr}
\end{equation}%
and ${^{k}}\mathbf{u}_{\beta }=\mathbf{R}_{k,r}{^{r}}\mathbf{u}_{\beta }$.
Fig. \ref{figCutJoint}b) shows this for a U joint. The corresponding
velocity constraints are $\dot{g}_{\alpha ,\beta }^{\mathrm{rot}}(%
\mathbold{\vartheta}%
_{\left( \lambda ,l\right) },\dot{%
\mathbold{\vartheta}%
}_{\left( \lambda ,l\right) })=0$ and $\ddot{g}_{\alpha ,\beta }^{\mathrm{rot%
}}(%
\mathbold{\vartheta}%
_{\left( \lambda ,l\right) },\dot{%
\mathbold{\vartheta}%
}_{\left( \lambda ,l\right) },\ddot{%
\mathbold{\vartheta}%
}_{\left( \lambda ,l\right) })=0$, with%
\begin{align}
\dot{g}_{\alpha ,\beta }^{\mathrm{rot}}& ={^{k}}\mathbf{u}_{\alpha }^{T}%
\mathbf{A}_{\beta }^{\mathrm{rot}}\dot{%
\mathbold{\vartheta}%
}_{\left( \lambda ,l\right) }  \label{gRotDot} \\
\ddot{g}_{\alpha ,\beta }^{\mathrm{rot}}& ={^{k}}\mathbf{u}_{\alpha }^{T}%
\mathbf{A}_{\lambda }^{\mathrm{rot}}\ddot{%
\mathbold{\vartheta}%
}_{\left( \lambda ,l\right) }+{^{k}}\mathbf{u}_{\alpha }^{T}\dot{\mathbf{A}}%
_{\lambda }^{\mathrm{rot}}\dot{%
\mathbold{\vartheta}%
}_{\left( \lambda ,l\right) }  \label{gRot2Dot}
\end{align}%
with $\mathbf{A}_{\beta }^{\mathrm{rot}}:=\mathbf{B}_{\beta }^{\mathrm{rot}}%
\mathbf{J}_{\lambda }$ and $\dot{\mathbf{A}}_{\beta }^{\mathrm{rot}}=\mathbf{%
B}_{\beta }^{\mathrm{rot}}\dot{\mathbf{J}}_{\lambda }+\dot{\mathbf{B}}%
_{\beta }^{\mathrm{rot}}\mathbf{J}_{\lambda }$, where%
\begin{align*}
\mathbf{B}_{\beta }^{\mathrm{rot}}& :=%
\Bigg%
(%
\begin{array}{cccc}
{^{k}\widetilde{\mathbf{u}}}_{\beta } & \ \ \ \mathbf{0} & \ \ \ -\mathbf{R}%
_{k,r}{^{r}\widetilde{\mathbf{u}}}_{\beta } & \ \ \ \mathbf{0}%
\end{array}%
\Bigg%
) \\
\dot{\mathbf{B}}_{\beta }^{\mathrm{rot}}& =-%
\Bigg%
(%
\begin{array}{cccc}
\widetilde{\mathbf{c}} & \ \ \ \mathbf{0} & \ \ \ \ \ ({^{k}}\widetilde{%
\bm{\omega}}_{k}-{^{k}}\widetilde{\bm{\omega}}_{r}){^{k}\widetilde{\mathbf{u}%
}}_{\beta }\mathbf{R}_{k,r} & \ \ \mathbf{0}%
\end{array}%
\Bigg%
)
\end{align*}%
abbreviating $\mathbf{c}:={^{k}\widetilde{\mathbf{u}}}_{\beta }({^{k}}%
\bm{\omega}_{k}-{^{k}}\bm{\omega}_{r})$. In the expression $\dot{\mathbf{B}}%
_{\beta }^{\mathrm{rot}}\mathbf{J}_{\lambda }\dot{%
\mathbold{\vartheta}%
}_{\left( \lambda ,l\right) }$, matrix $\dot{\mathbf{B}}_{\beta }^{\mathrm{%
rot}}$ can be replaced by%
\begin{equation*}
\dot{\overline{\mathbf{B}}}_{\beta }^{\mathrm{rot}}:=-%
\Bigg%
(%
\begin{array}{cccc}
{^{k}}\widetilde{\bm{\omega}}_{k}{^{k}\widetilde{\mathbf{u}}}_{\beta } & \ \
\ \mathbf{0} & \ \ \ \ \ ({^{k}}\widetilde{\bm{\omega}}_{r}-2{^{k}}%
\widetilde{\bm{\omega}}_{k}){^{k}\widetilde{\mathbf{u}}}_{\beta }\mathbf{R}%
_{k,r} & \ \ \mathbf{0}%
\end{array}%
\Bigg%
)
\end{equation*}%
which may be computationally advantageous (notice that $\dot{\overline{%
\mathbf{B}}}_{\beta }^{\mathrm{rot}}$ is not the time derivative of $\mathbf{%
B}_{\beta }^{\mathrm{rot}}$).

\subsection{Constraints for Technical Joints}

Joint specific constraints are introduced by combining the above elementary
constraints. The geometric constraints (\ref{GeomConsLoopCutJoint}) are
formed by the corresponding position and orientation constraints determined
by (\ref{PosConst}) and (\ref{OriConstr}). The rows of the constraint
Jacobian $\mathbf{G}_{\left( \lambda ,l\right) }$ of the velocity
constraints (\ref{VelConsLoopCutJoint}) are ${^{k}}\mathbf{u}_{\alpha
,\lambda }^{T}\mathbf{A}_{\lambda }$ (\ref{gPosDot}), for the position
constraints, and ${^{k}}\mathbf{u}_{\alpha }^{T}\mathbf{A}_{\beta }^{\mathrm{%
rot}}$ in (\ref{gRotDot}) for rotation constraints. The rows of $\dot{%
\mathbf{G}}_{\left( \lambda ,l\right) }$ in (\ref{AccConsLoopCutJoint}) are $%
{^{k}}\mathbf{u}_{\alpha ,\lambda }^{T}\dot{\mathbf{A}}_{\lambda }$ in (\ref%
{gPos2Dot}), and ${^{k}}\mathbf{u}_{\alpha }^{T}\dot{\mathbf{A}}_{\lambda }^{%
\mathrm{rot}}$ in (\ref{gRot2Dot}).

Revolute, spherical, and universal joints, for instance, do not permit
relative displacements, and ${^{k}}\mathbf{d}_{\lambda }$ must vanish. Then (%
\ref{GeomConsLoopCutJoint}) comprises the three translation constraints $%
g_{\alpha }^{\mathrm{pos}}=0,\alpha =1,2,3$ 
\begin{equation}
\mathbf{g}_{\left( \lambda ,l\right) }=\left( g_{1}^{\mathrm{pos}},g_{2}^{%
\mathrm{pos}},g_{3}^{\mathrm{pos}}\right) ^{T}:={^{k}}\mathbf{d}_{\lambda }.
\end{equation}%
Prismatic and cylindrical joints allow for translation along an axis. Denote
with ${^{k}}\mathbf{u}_{\alpha ,\lambda },\alpha =1,2$ independent unit
vectors perpendicular to this axis. The two position constraints $g_{\alpha
}^{\mathrm{pos}}=0,\alpha =1,2$ are then defined by (\ref{PosConst}), and $%
\mathbf{g}_{\left( \lambda ,l\right) }=\left( g_{1}^{\mathrm{pos}},g_{2}^{%
\mathrm{pos}}\right) ^{T}$.

A joint with $\delta ^{\mathrm{rot}}$ rotary DOFs imposes $3-\delta ^{%
\mathrm{rot}}$ such orientation constraints (\ref{OriConstr}).

\section{List of Symbols%
\label{secSymbols}%
}

$%
\begin{tabular}{lll}
$\mathfrak{n}_{l}$ & - & umber of tree-joints in limb $l$ \\ 
$L$ & - & number of (complex) limbs \\ 
$\Gamma _{\left( l\right) }$ & - & topological graph of limb $l=1,\ldots ,L$
\\ 
$\Lambda _{\lambda \left( l\right) }$ & - & FC $\lambda =1,\ldots ,\gamma
_{l}$ of limb $l$ \\ 
$\gamma _{l}$ & - & number of FCs of $\Gamma _{\left( l\right) }$, number of
loops in limb $l$ \\ 
$n$ & - & number of joint variables of the tree-topology system according to 
$\vec{G}$ \\ 
$n_{l}$ & - & number of tree-joint variables of limb $l$ \\ 
$n_{\lambda ,l}$ & - & number of tree-joint variables of limb $l$ when the
platform is removed \\ 
$\bar{n}:=\bar{n}_{1}+\ldots +\bar{n}_{L}$ & - & total number of variables
of the tree-topology system when the platform is removed \\ 
$\underline{k}$ & - & index of the last body in the path from body $k$ to
the ground, i.e. $0=\underline{k}-1$ \\ 
$\lambda $ & - & cut-edge of $\Lambda _{l}$. It serves as start edge when
traversing the FC. \\ 
$\bar{\lambda}$ & - & last edge when traversing the FC starting from $%
\underline{\lambda }$ \\ 
$m_{\lambda ,l}$ & - & number of independent loop constraints of FC $\lambda 
$ of limb $l$ \\ 
$\delta _{0,l}$ & - & number of joint variables of limb $l$ that are not
part of a FC of $\Gamma _{\left( l\right) }$ \\ 
$\delta _{\lambda ,l}=n_{\lambda ,l}-m_{\lambda ,l}$ & - & DOF of FC $%
\lambda $ of limb $l$ when disconnected from the platform \\ 
$\delta _{l}=\delta _{0,l}+\delta _{1,l}+\ldots +\delta _{\gamma _{l},l}$ & -
& DOF of limb $l$ when disconnected from the platform \\ 
$\delta _{\mathrm{P}\left( l\right) }$ & - & platform DOF of separated limb $%
l$ \\ 
$\delta _{\mathrm{P}}$ & - & platform DOF of PKM \\ 
$%
\mathbold{\vartheta}%
_{\left( l\right) }\in {\mathbb{V}}^{n_{l}}$ & - & vector of tree-joint
variables $\vartheta _{1},\ldots ,\vartheta _{n_{l}}$ of limb $l$ (when
connected to platform) \\ 
$\bar{%
\mathbold{\vartheta}%
}_{\left( l\right) }\in {\mathbb{V}}^{\bar{n}_{l}}$ & - & vector of
tree-joint variables $\vartheta _{1},\ldots ,\vartheta _{\bar{n}_{l}}$ of
limb $l$ (when disconnected from platform) \\ 
$%
\mathbold{\vartheta}%
_{\mathrm{act}}$ & - & vector of actuated joint coordinates \\ 
$\mathbf{q}_{\left( \lambda ,l\right) }$ & - & vector of $\delta _{\lambda
,l}$ independent variables used to express the solution the loop constraints
for $\Lambda _{\left( \lambda ,l\right) }$ \\ 
$\mathbf{q}_{\left( l\right) }$ & - & vector of $\delta _{l}$ generalized
coordinates of limb $l$ when separated from PKM \\ 
$\varphi _{\mathrm{p}\left( l\right) }$ & - & forward kinematics map of limb 
$l$, so that $\mathbf{C}_{\mathrm{p}\left( l\right) }(%
\mathbold{\vartheta}%
_{\left( l\right) })=\varphi _{\mathrm{p}\left( l\right) }(\mathbf{q}%
_{\left( l\right) })$ \\ 
$f_{\left( l\right) }$ & - & inverse kinematics map of limb $l$, so that $%
\mathbf{q}_{\left( l\right) }=f_{\left( l\right) }(\mathbf{C}_{\mathrm{p}})$
\\ 
$\mathbf{L}_{\mathrm{t}\left( l\right) }$ & - & task space Jacobian of limb $%
l$, so that $\mathbf{V}_{\mathrm{t}}=\mathbf{L}_{\mathrm{t}\left( l\right) }%
\dot{\mathbf{q}}_{\left( l\right) }$ \\ 
$\mathbf{F}_{\left( l\right) }$ & - & inverse kinematics Jacobian of limb $l$%
, so that $\mathbf{V}_{\mathrm{p}\left( l\right) }=\mathbf{L}_{\mathrm{p}%
\left( l\right) }\dot{\mathbf{q}}_{\left( l\right) }$ \\ 
$\widetilde{\mathbf{x}}=\left( 
\begin{array}{ccc}
0 & -x_{3} & x_{2} \\ 
x_{3} & 0 & -x_{1} \\ 
-x_{2} & x_{1} & 0%
\end{array}%
\right) \in so\left( 3\right) $ & - & $\mathbf{x}=\left( 
\begin{array}{c}
x_{1} \\ 
x_{2} \\ 
x_{3}%
\end{array}%
\right) \in {\mathbb{R}}^{3}$ \\ 
$\widehat{\mathbf{X}}=\left( 
\begin{array}{cc}
\widetilde{\mathbf{x}} & \mathbf{y} \\ 
\mathbf{0} & 0%
\end{array}%
\right) \in se\left( 3\right) $ & - & $\mathbf{X}=\left( 
\begin{array}{c}
\mathbf{x} \\ 
\mathbf{y}%
\end{array}%
\right) \in {\mathbb{R}}^{6},\ \ \mathbf{x},\mathbf{y}\in {\mathbb{R}}^{3}$
\\ 
$\mathbf{Y}_{i}=\left( \mathbf{e}_{i},\mathbf{y}_{i}\times \mathbf{e}%
_{i}+h_{i}\mathbf{e}_{i}\right) ^{T}$ & - & Screw coordinate vector of joint 
$i$, $\mathbf{e}_{i}\in {\mathbb{R}}^{3}$ is a unit vector along the joint
axis, \\ 
&  & $\mathbf{y}_{i}\in {\mathbb{R}}^{3}$ is the position vector to a point
on the axis, $h_{i}\in {\mathbb{R}}^{+}$ is the pitch of the joint. \\ 
$\mathbf{Y}_{i}=\left( \mathbf{e}_{i},\mathbf{y}_{i}\times \mathbf{e}%
_{i}\right) ^{T}$ & - & Screw coordinate vector for a revolute joint \\ 
$\mathbf{Y}_{i}=\left( \mathbf{0},\mathbf{e}_{i}\right) ^{T}$ & - & Screw
coordinate vector for a prismatic joint \\ 
$\mathbf{C}_{i}\in SE\left( 3\right) $ & - & absolute configuration of body $%
i$ w.r.t. to IFR \\ 
$\mathbf{C}_{i,j}:=\mathbf{C}_{i}^{-1}\mathbf{C}_{j}\in SE\left( 3\right) $
& - & relative configuration of body $j$ relative to body $i$ \\ 
$\mathbf{Ad}_{\mathbf{C}}=\left( 
\begin{array}{cc}
\mathbf{R} & \mathbf{0} \\ 
\widetilde{\mathbf{r}}\mathbf{R} & \mathbf{R}%
\end{array}%
\right) $ & - & 
\hspace{-1.5ex}%
\begin{tabular}{l}
Adjoint operator, so that $\mathbf{Y}^{1}=\mathbf{Ad}_{\mathbf{C}_{12}}%
\mathbf{Y}^{2}$ is the transformation of screw coordinates \\ 
represented in frame 1 to those represented in frame 2%
\end{tabular}
\\ 
$\mathbf{ad}_{{\mathbf{X}}}=\left( 
\begin{array}{cc}
\widetilde{%
\mathbold{\xi}%
}\ \  & \mathbf{0} \\ 
\widetilde{%
\mathbold{\eta}%
}\ \  & \widetilde{%
\mathbold{\xi}%
}%
\end{array}%
\right) $ & - & 
\hspace{-1.5ex}%
\begin{tabular}{l}
Matrix form of the adjoint operator, so that $\mathbf{ad}_{{\mathbf{Y}}_{1}}%
\mathbf{Y}_{2}$ is the Lie bracket (screw product) of \\ 
two screw coordinate vectors $\mathbf{Y}_{1}=\left( 
\mathbold{\xi}%
_{1},%
\mathbold{\eta}%
_{1}\right) ^{T}$ and $\mathbf{Y}_{2}=\left( 
\mathbold{\xi}%
_{2},%
\mathbold{\eta}%
_{2}\right) ^{T}$.%
\end{tabular}
\\ 
FC & - & Fundamental Cycle (a topologically independent loop) of the
topological graph \\ 
EE & - & End-Effector \\ 
IFR & - & Inertial Frame%
\end{tabular}%
$

\section*{Acknowledgement}

This work has been supported by the LCM K2 Center for Symbiotic Mechatronics
within the framework of the Austrian COMET-K2 program.

\newpage%
\bibliographystyle{IEEEtran}
\bibliography{RobHandbook}

\end{document}